%% file: main.tex
\documentclass[manuscript]{acmart}
\usepackage{graphicx}
\usepackage{subcaption}
\usepackage{algpseudocode}
\usepackage{algorithm}
\usepackage{tcolorbox}
\usepackage{comment}
\AtBeginDocument{%
  \providecommand\BibTeX{{%
    \normalfont B\kern-0.5em{\scshape i\kern-0.25em b}\kern-0.8em\TeX}}}

\setcopyright{acmcopyright}
\copyrightyear{2023}
\acmYear{2023}

\begin{document}

\title{Reinforcement learning informed evolutionary search for autonomous systems testing}

\author{Dmytro Humeniuk}
\email{dmytro.humeniuk@polymtl.ca}
\orcid{0000-0002-2983-8312}
\affiliation{%
  \institution{Polytechnique Montréal}
  \city{Montreal}
  \state{QC}
  \country{Canada}
}

\author{Foutse Khomh}
\affiliation{%
  \institution{Polytechnique Montréal}
  \city{Montreal}
  \state{QC}
  \country{Canada}
}

\author{Giuliano Antoniol}
\affiliation{%
  \institution{Polytechnique Montréal}
  \city{Montreal}
  \state{QC}
  \country{Canada}
}


\begin{abstract}
Evolutionary search-based techniques are commonly used for testing autonomous robotic systems. However, these approaches often rely on computationally expensive simulator-based models for test scenario evaluation.  To improve the computational efficiency of the search-based testing, we propose augmenting the evolutionary search (ES) with a reinforcement learning (RL) agent trained using surrogate rewards derived from domain knowledge. In our approach, known as  RIGAA (Reinforcement learning Informed Genetic Algorithm for Autonomous systems testing), we first train an RL agent to learn useful constraints of the problem and then use it to produce a certain part of the initial population of the search algorithm. By incorporating an RL agent into the search process, we aim to guide the algorithm towards promising regions of the search space from the start, enabling more efficient exploration of the solution space.
We evaluate RIGAA on two case studies: maze generation for an autonomous `Ant' robot and road topology generation for an autonomous vehicle lane keeping assist system. In both case studies, RIGAA converges faster to fitter solutions and produces a better test suite (in terms of average test scenario fitness and diversity). RIGAA also outperforms the state-of-the-art tools for vehicle lane keeping assist system testing, such as AmbieGen and Frenetic.

\end{abstract}




\maketitle

\section{Introduction}\label{sec:intro}
Thorough testing of autonomous robotic systems, such as self-driving
cars, autonomous robots, and drones is a crucial step in their development cycle.
An important testing stage is software-in-the-loop or model-in-the-loop, where the system model is run in a virtual environment \cite{plummer2006model}. Test scenarios are typically represented by virtual environments, where the system model should accomplish a particular task.
However, the search space of possible parameters defining the test scenarios is huge, and simulating all the parameter combinations is computationally infeasible. It is important to have search-based techniques that generate scenarios with an increased probability of revealing a system failure. These scenarios can then be prioritized for testing the system. 

Recently, a number of search-based techniques were successfully applied to test case generation for autonomous systems. 
In the context of testing, a good test scenario is the one that makes the system falsify some of its requirements, i.e., reveals its failure.
Gambi et al. \cite{gambi2019automatically}, Riccio and Tonella \cite{RiccioTonella_FSE_2020}, Castellano et al. \cite{castellano2021frenetic}, Zohdinasab et al. \cite{zohdinasab2021deephyperion} generate test cases for vehicle lane keeping assist system (LKAS) with evolutionary search algorithms (EA). Abdessalem et al. \cite{abdessalem2018testing}, \cite{abdessalem2018testing2} use a
multi-objective search evolutionary algorithm NSGA-II to obtain test scenarios for autonomous vehicle automated emergency braking (AEB) system. 
One of the main limitations of these approaches is the use of the computationally expensive simulator-based system model to perform the fitness function evaluations.  
A simulator often takes a non-negligible amount of time to evaluate a candidate solution, ranging from tens of seconds to minutes. Moreover, the computational requirements and costs to evaluate each scenario are high. As an example, the minimal requirements for running the BeamNG simulator are 16 GB RAM, Nvidia GeForce GTX 970 videocard and Intel Core i7-6700 3.4Ghz processor or better.
Given that each evaluation is expensive, we are interested in approaches that allow to minimize the use of computational power on the evaluation of non-promising solutions, i.e., those that are very unlikely to lead to a system failure. 
Evolutionary algorithms are effective for exploring large search spaces,  however, their execution involves a considerable number of stochastic decisions, such as parent selection, crossover, and mutation \cite{coello2007evolutionary}. With each evaluation being computationally expensive, they can take a significant amount of time to converge to the promising regions of the search space. This is a disadvantage from the testing perspective, as the testing budget is limited. 

One potential solution is to use some heuristics with prior problem knowledge to guide the search \cite{briffoteaux2022parallel}.
Examples include inserting high-quality solutions into the initial population or using surrogate models to prioritize and select certain solutions \cite{rosales2013hybrid}. In the context of search-based testing, while the combination of surrogate modeling and evolutionary algorithms has been well studied \cite{haq2022efficient}, current techniques for achieving good initialization are limited. These techniques often rely on manually generated test scenarios or test scenarios identified by the search process previously \cite{moghadam2022machine}.

In our approach, we propose to use gradient-based algorithms, specifically reinforcement learning (RL), to automatically and effectively infer the domain knowledge of the problem and then use it to initialize the population of the evolutionary algorithm.
RL algorithms have proven to be effective in generating solutions to problems that can be formulated as a Markov Decision Process (MDP) \cite{sutton2018reinforcement}.
We have the following intuition:  by formulating test scenario generation as an MDP, RL agents can learn the initial constraints of the problem and identify test scenarios with higher fitness compared to randomly generated ones.
However, state-of-the-art RL algorithms are known to be sample inefficient, signifying they require a large number of evaluations before being able to learn a meaningful representation \cite{sutton2018reinforcement}. Consequently, training an RL agent using rewards obtained from executing a simulator-based model of the system becomes computationally prohibitive.

In order to further minimize the computational costs, we suggest training the RL agent using surrogate rewards, based on some domain knowledge of what constitutes a challenging test scenario. We surmise that as long as the trained agent can produce solutions of higher fitness than a random generator and of high diversity, it can be useful for evolutionary algorithm initialization.
We refer to our approach as RIGAA (Reinforcement learning Informed Genetic Algorithm for Autonomous systems testing). 
By leveraging RL for initialization, RIGAA is able to converge to better test scenarios within a smaller time budget, than a randomly initialized search algorithm. The RL agent needs to be trained only once and then can be reused for generating the test scenarios. We choose the reward function to be inexpensive to evaluate so that the RL agent training does not impose a significant computational overhead.

We evaluated our approach for test scenario generation on two different systems: an autonomous RL-controlled ant robot navigating in a maze and an autonomous vehicle lane keeping system (LKAS). These two systems have been previously used in other studies, including the SBST 2022 CPS testing competition  \cite{gambi2022sbst} and RL benchmarking \cite{fu2020d4rl}.

Results show that introducing RL-generated individuals with higher fitness levels into the initial population of a genetic algorithm can lead to faster convergence towards more challenging test scenarios. We also demonstrate that using an RL agent trained with surrogate rewards can be valuable for generating test scenarios for a simulator-based system model. 
RIGAA outperformed NSGA-II by producing 20\% more effective test suites  for the autonomous ant robot in the Mujoco simulator and 12.26\% more effective test suites for the autonomous vehicle in the BeamNG simulator. RIGAA also
discovered 23.6 \% more failures than AmbieGen tool and 83.2 \% more failures than the Frenetic tool, which are state-of-the-art tools for LKAS testing.

In this paper, we make the following contributions: 
\begin{enumerate}
    \item RIGAA, a search-based tool for testing autonomous robotic systems, allowing to insert some problem knowledge into the search via a pre-trained RL agent;
    \item An extensive evaluation of RIGAA, including the assessment of the RL agent's performance and a detailed analysis of the impact of the percentage of RL-produced individuals in the initial population;
    \item  A publicly available replication package, including the implementation of RIGAA (see Section \ref{sec:rep_pack}).
\end{enumerate}

The remainder of this paper is organized as follows.  
Section \ref{sec:background} provides background information on genetic algorithms (GA) and reinforcement learning (RL). In Section \ref{sec:problem} we formalize the problem of scenario generation and describe the concrete test generation problems addressed in this work. Section \ref{sec:related_work} discusses the related works in the domain of search-based and RL based test scenario generation.  We provide the description of RIGAA approach and our methodology in Section \ref{sec:approach}. In Section \ref{sec:evaluation} we formulate the research questions, present the subjects of the study, our evaluation results, and potential threats to validity. Section \ref{sec:discussion} discusses some benefits and the main challenges of using RL agents for EA initialization, as well as the avenues for future work. Section \ref{sec:conclusions} concludes the paper.

\section{Background} \label{sec:background}
In this section, we provide some background information on genetic algorithms and Reinforcement learning, which are essential components of the RIGAA approach.
\subsection{Genetic algorithms}

\begin{figure}[h!]
\includegraphics[scale=0.46]{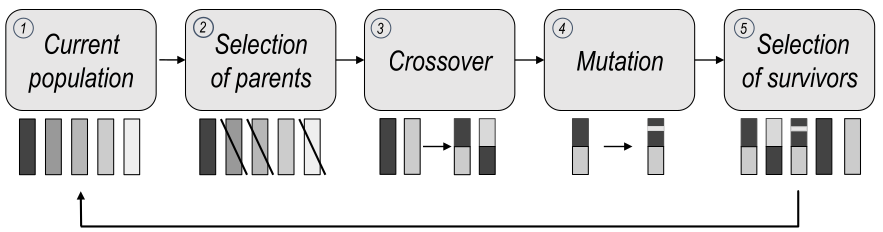}
\centering
\caption{Evolutionary search pipeline}
\label{fig:ga_pipeline}
\end{figure}

Genetic algorithms (GA) are a class of evolutionary algorithms that use a set of principles from genetics, such as selection, crossover, and mutation, to evolve a population of candidate solutions \cite{back1997handbook}. A typical genetic algorithm pipeline is shown in Figure \ref{fig:ga_pipeline}.
The  basic  idea is to start with a set of individuals (candidate solutions) representing the initial population, usually generated randomly from the allowable range of values. Each individual is encoded in a dedicated form, such as a bitstring, and is called a chromosome. A chromosome is composed of genes. Each  individual  is   evaluated and  assigned  a  fitness  value. Some of the individuals are selected for mating. The  search  is  continued  until  a  stopping  criterion  is  satisfied  or  the  number  of  iterations  exceeds   a  specified  limit. 
Three genetic operators are used to evolve the solutions: selection of survivors and parents, crossover, and mutation.
The operators in evolutionary algorithms serve different functions. Mutation and crossover operators promote diversity in the population, allowing for the exploration of the solution space. On the other hand, survivor and parent selection operators promote the quality of the solutions, enabling the exploitation of the search space. It is essential to select the appropriate combination of operators that offers a good balance between exploration and exploitation to achieve optimal results \cite{eiben2015introduction}.

Selection of parents is an operator that gives solutions with higher fitness a higher probability of contributing to one or more children in the succeeding generation. The intuition is to give better individuals more opportunities to produce offsprings.
One of the commonly used selection operators is  
tournament selection \cite{goldberg1991comparative}. The  crossover  operator is used  to  exchange characteristics of candidate solutions among themselves. The mutation operator has been introduced to prevent convergence to local optima; it randomly modifies an individual’s genome (e.g., by flipping some of its bits, if the genome is represented by a bitstring) \cite{antoniol2005search}. In contrast to parent selection, which is typically
stochastic, survivor selection is often deterministic.
For genetic algorithms ($\mu$ + $\lambda$) survivor selection strategy is commonly used. In this strategy, the set of offspring and parents are merged and ranked according to (estimated) fitness, then the top $\mu$ individuals are kept to form the next generation.
\cite{eiben2015introduction}.

Genetic algorithms are effective in solving both single and multi-objective optimization problems \cite{coello2007evolutionary}. However, they are particularly suited to multi-objective problems since they can simultaneously handle a set of possible solutions, allowing the algorithm to find multiple members of the Pareto optimal set in a single run.
Under Pareto optimality, a solution is better than another if it is superior in at least one of the individual fitness functions and not worse in any of the others. This is an alternative to simply aggregating fitness using a weighted sum of the n fitness functions \cite{eiben2015introduction}. 

One is the most popular multi-objective genetic algorithms is the non-dominated sorting algorithm-II (NSGA-II) \cite{deb2002fast}. Within each non-dominated front, NSGA-II sorts the individuals based on their crowding distance, which is  the  average  distance  to  its  two  
neighbouring  individuals in the Pareto front. It uses the crowding distance in its selection operator
to keep a diverse front by making sure each member stays a crowding
distance apart. This keeps the population diverse and helps the algorithm to
explore the fitness landscape \cite{coello2007evolutionary}.
\subsection{Reinforcement learning}
Reinforcement learning (RL) is a subfield of machine learning that focuses on training agents to make sequential decisions in an environment in order to maximize a reward signal. RL agents learn through trial and error, interacting with the environment and adjusting their behavior based on the received rewards \cite{sutton2018reinforcement}.

In the standard mathematical formulation of an RL problem, an RL agent learns to take actions in an environment in order to maximize a reward signal.
This is modeled as a Markov decision process (MDP), which is defined by a 5-tuple:
    $S$: a set of states that the agent can occupy,
    $A$: a set of actions that the agent can take,
    $R: S \times A \rightarrow \mathbb{R}$: a reward function that maps state-action pairs to real-valued rewards,
    $P: S \times A \rightarrow S$: a transition function that defines the probability of transitioning to a new state given the current state and action,
    $\rho_0: S \rightarrow [0, 1]$: an initial state distribution that specifies the probability of starting in each state.
Typically, an RL agent interacts with an environment for one episode, and then its state is reset. An episode is a sequence of interactions between an agent and its environment, starting from an initial state and proceeding through a series of state transitions based on the actions taken by the agent. An episode typically ends when the agent reaches a terminal state or when a specified time limit is reached. The RL agent's goal is to take actions that maximize the expected sum of future rewards obtained during an episode, as shown in Eq. \ref{eq:rewards_sum}:
\begin{align} \label{eq:rewards_sum}
   R = \sum_{t=0}^{T}\gamma^tr_{t} 
\end{align}
where $R$ is the expected reward over the episode, $r_{t}$ is the reward obtained at time $t$, $T$ is the total number of time steps in the given
episode and $\gamma$ is the discount factor that determines the importance of future rewards. 
The agent can maximize the reward by learning a good policy, which is a function that maps states $s$ to actions $a$. The learnt policy $\pi$ is typically evaluated with a Q-function, which provides a way to estimate the expected cumulative reward that an agent can achieve when following a specific policy $\pi$ and starting in a state $s$ with an action $a$:
\begin{align}
Q^\pi(s,a) = \mathbb{E}_\pi \left[ R | s_0 = s, a_0 = a \right]
\end{align}
The objective of the agent is to find the policy $\pi$ with the highest value of Q-function.
The policy can be learned using various RL algorithms, such as Q-learning, policy gradients, or actor-critic algorithms \cite{sutton2018reinforcement}.
One popular state-of-the-art RL algorithm is Proximal Policy Optimization (PPO) \cite{schulman2017proximal}, which is a model-free actor-critic-based algorithm that updates the policy parameters based on the current estimate of the reward function. It has achieved state-of-the-art performance in complex tasks such as robot locomotion training \cite{heess2017emergence} and playing video games \cite{berner2019dota}.

\section{Problem formulation} \label{sec:problem}
In this section, we formulate the problem of test scenario generation taking into account our representation. We describe the two concrete test scenario generation problems we
address in this work. We formulate them as optimization problems, with constraints to produce valid test scenarios and objectives to falsify the requirements of the system under test.

We begin by presenting the definition of the test scenario as well as the representation we use. A test scenario for an autonomous robotic system is a virtual environment in which it operates, as well as the task it needs to accomplish. Similarly to our previous work \cite{humeniuk2022search}, we represent the virtual environment as a combination of discrete elements $E_m$, which we refer to as scenario elements. Each scenario element has a fixed number of attributes $A_n$ that describe its properties. Such representation makes it easy to define and change various parts of a test scenario. The general test scenario  representation is shown in Table \ref{tab:repr}. 
We represent it as $nxm$ matrix, where $m$ defines the number of discrete elements composing the test scenario and $n$ the number of attributes describing each of the elements, i.e. each element $E_m$ is described by attributes $A_{1em}$, $A_{2em}$, $A_{nem}$. For instance, if our scenario is a road topology, each discrete element $E_m$ may correspond to a segment of the road. Each attribute $A_n$ is used to describe a particular aspect of the road, such as its length, curvature, material, etc.  The goal is to find such a combination of scenario elements $E_m$ and the values of attributes $A_n$, that the virtual environment respects the physical constraints $C$ and forces the system to violate the requirements $R$. 
\input{tables/repr.tex}

Below, we provide descriptions of test scenario generation problems we considered in our work.
\subsubsection{Autonomous robot maze}
In our study, a test scenario for an autonomous robot constitutes a closed room with obstacles as well as a start and goal location for the robot.  Similar test scenarios were used in some previous research works \cite{arnold2013testing}, \cite{sotiropoulos2016virtual}. 

\begin{figure}[h!]
\begin{subfigure}{0.45\textwidth}
  \centering
  \includegraphics[scale=0.5]{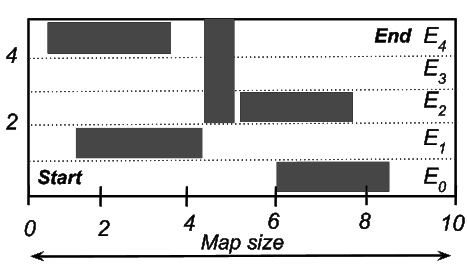}  
  \caption{Autonomous robot scenario represented in Table \ref{tab:rob_scenario}}
  \label{fig:rob_scenario1}
\end{subfigure}
\begin{subfigure}{0.4\textwidth}
  \centering
  \includegraphics[scale=0.54]{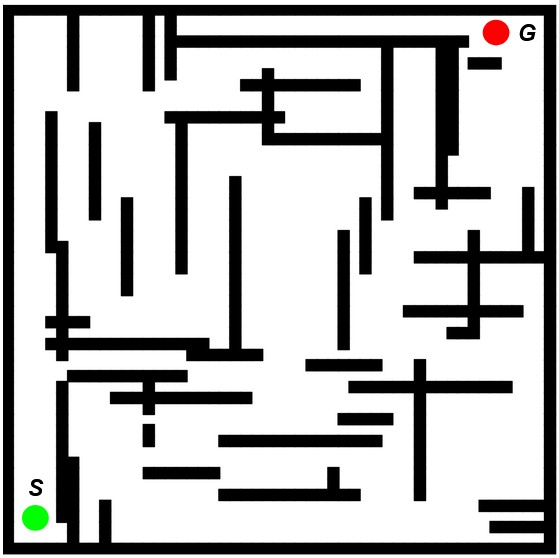}  
  \caption{An example of a test scenario for autonomous robotic system}
  \label{fig:rob_scenario2}
\end{subfigure}
\caption{Autonomous robot system scenario examples}
\label{fig:rob_scenario}
\end{figure}

An example of encoding a simplified scenario is shown in Figure \ref{fig:rob_scenario1} and Table \ref{tab:rob_scenario}. In this example, the map has a size of 5 m x 10 m, and each scenario element $E_m$ corresponds to an area of 1 m x 10 m, for a total of 5 scenario elements. We specify the type, size, and location of obstacles for each scenario element.

An example of a typical test scenario is shown in Figure \ref{fig:rob_scenario2}. It depicts a square room of size 40 m x 40 m with vertical and horizontal walls as obstacles. The robot must navigate from the starting location $S$ (green point) to the goal location $G$ (red point) using only its range sensors and planning algorithm. The starting and goal locations remain constant across all scenarios.
\input{tables/rob_scenario.tex}

In this test scenario, the environment consists of 40 elements $E_m$, each representing a 1 m x 40 m area containing one obstacle (horizontal or vertical wall). Each element has three attributes: $A_1$, the type of the obstacle; $A_2$, the position of the obstacle (ranging from 2 to 38); and $A_3$, the size of the obstacle (ranging from 5 to 15). The goal of the test is to evaluate the ability of the autonomous system to navigate from the start to the goal location without getting stuck or colliding with obstacles. Test scenarios should adhere to the following physical constraints $C$: (1) there must always be at least one path between the start and goal locations; (2) obstacles must not extend beyond the boundaries of the room. The main requirement $R$ for the autonomous system is to reach the goal location safely, without getting stuck or colliding with obstacles.

\subsubsection{Road topology for autonomous vehicle}
In the second problem, the goal is to create a road topology that causes the lane keeping assist system (LKAS) of an autonomous vehicle to fail. 

The road is represented as a sequence of points on a two-dimensional map with a size of 200 m x 200 m. These points are interpolated using cubic splines to form a smooth road with two lanes. The first and last points on the road define the starting and ending locations, respectively.

\begin{figure}[h]
\begin{subfigure}{0.45\textwidth}
  \centering
  \includegraphics[scale=0.37]{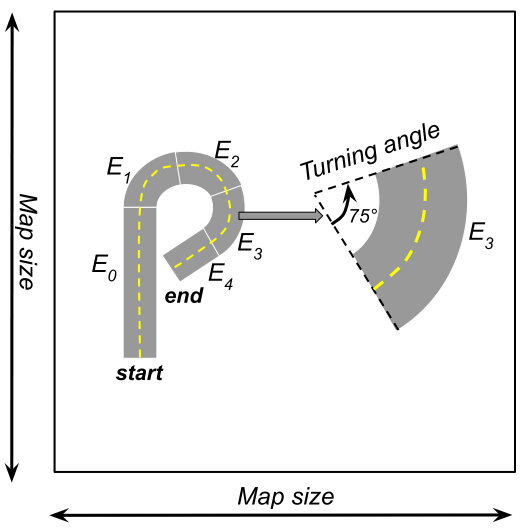}  
  \caption{Autonomous vehicle scenario represented in Table \ref{tab:veh_scenario} }
  \label{fig:veh_scenario2}
\end{subfigure}
\begin{subfigure}{0.4\textwidth}
  \centering
  \includegraphics[scale=0.38]{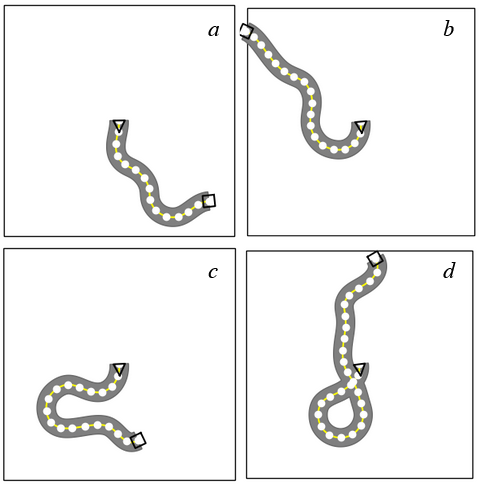}  
  \caption{An example of valid (a) and invalid (b-d) test scenarios for the vehicle LKAS system}
   \label{fig:veh_scenario1}
\end{subfigure}
\caption{Autonomous vehicle system scenario example}
\label{fig:rob_scenario_all}
\end{figure}

An example of encoding a simplified scenario using our representation is shown in Table \ref{tab:veh_scenario} and illustrated in Figure \ref{fig:veh_scenario2}. In the example, the road topology is represented as a combination of five scenario elements, each corresponding to a road segment. We define three attributes to describe each segment: $A_0$, the type of road (straight, right turn, or left turn); $A_1$, the length of the straight road segment (in the range of 5 to 50); and $A_2$, the angle of the turn of the curved segment (in the range of 5 to 85).

\input{tables/veh_scenario.tex}

The constraints $C$ for the test scenario include: (1) the road has to remain within the map borders, (2) the road cannot be too sharp, and (3) the road cannot self-intersect. Examples of valid and invalid scenarios are illustrated in Figure \ref{fig:veh_scenario1}, where scenario (a) is valid, scenario (b) features a road that goes outside the map bounds, scenario (c) shows a road with a too sharp turn, and scenario (d) shows a self-intersecting road.

The main requirement $R$ for the system is to stay within the lane boundaries, with a threshold of $p=0.85$, meaning that the vehicle is considered to be out of the lane if more than 85\% of it lies outside the lane boundaries.

\section{Related work} \label{sec:related_work}
This section is divided into two parts. In the first part, we discuss approaches that improve the efficiency of evolutionary search (ES) in generating test scenarios. In the second part, we review prior works on using RL  for test scenario generation.

\textbf{Search-based test scenario generation for autonomous robotic systems.} 
The AsFault tool \cite{gambi2019automatically} improves the efficiency of scenario generation for vehicle LKAS testing by identifying and filtering out similar tests before executing them. This basic technique can be easily incorporated into ES. In our approach, we remove duplicates at each iteration based on a similarity threshold. The Frenetic tool for LKAS testing \cite{castellano2021frenetic} introduces several heuristics to improve the efficiency of search operators, such as using different mutations depending on whether the test passed or failed, and selecting mutation candidates based on certain fitness thresholds. Improving search operators is orthogonal to our approach and can be combined with RIGAA. We surmise one of the promising research directions is to use RL to automatically learn the best mutation or crossover operators to apply, as previously done in research works in other domains \cite{quevedo2021using}, \cite{chen2020reinforcement}, \cite{sakurai2010method}. 

Moghadam et al. \cite{moghadam2022machine} utilize an initial quality population seed to enhance the search process. The quality population consists of a combination of valid random solutions and solutions that achieved high fitness in earlier runs of the tool. In contrast, our approach leverages a pre-trained RL agent to automatically generate high-quality, diverse solutions and includes them in the initial population. Reusing the solutions generated in previous runs can prevent the search from exploring diverse search space regions. 

One of the earliest works to incorporate machine learning (ML) with ES in the context of test scenario generation is the study by Abdessalem et al. \cite{abdessalem2018testing2}. The authors used ML models, specifically decision trees, to better guide the search process towards promising areas for efficient online testing of vision-based advanced driver systems (ADS). However, one limitation of the approach is the low scalability of decision trees (DTs) to higher dimensions of the test scenarios. For instance, when the algorithm input is high-dimensional, such as in one of our case studies (3x40), DTs would require a significant number of examples and iterations to learn a meaningful representation.

Birchler et al. \cite{birchler2022cost} use ML to
identify simulation-based tests that are unlikely to detect faults in
the autonomous vehicle software under test and skip them before their execution. They use three types of basic ML models: logistic regression, Naive Bayes and Random Forest. The F1 of the reported models is between  47\% and 90\%, indicating that the tool may mistakenly discard some fault-revealing test cases. Moreover, creating the datasets is time-consuming and computationally expensive, requiring a big number of evaluations of a simulator-based model. 

Haq et al. develop SAMOTA (Surrogate-Assisted Many-Objective Testing Approach) \cite{haq2022efficient},
an approach for test suite generation that  utilizes surrogate models to guide the search, minimizing the number of computationally expensive simulator-based evaluations. Surrogate models are designed to  mimic the simulator and to be much less expensive to run. This is a promising approach and can be combined with RIGAA. For example, the initial population can be generated with the RIGAA strategy, and the search can then be guided by the surrogate models from SAMOTA.

\textbf{Reinforcement learning for test scenario generation.} 
Chen et al. \cite{chen2021adversarial} propose an adaptive evaluation framework to efficiently evaluate autonomous  vehicles  in  adversarial  lane-change  scenarios  generated  by deep reinforcement learning (DRL). 
Lee et al. \cite{lee2020adaptive} present adaptive stress testing (AST), a framework for finding the most likely path to a failure event in simulation. They formulate the problem as a Markov decision process and use RL to optimize it, demonstrating the effectiveness of the approach  on an aircraft collision avoidance application.
The main limitation of these approaches is the use of a computationally expensive simulator-based model to train the RL agent. In our work, we reduce the RL agent training costs by utilizing domain knowledge-based surrogate rewards for RL agent training.
\section{RIGAA: Approach}\label{sec:approach}
\begin{figure}[h!]
\includegraphics[scale=0.45]{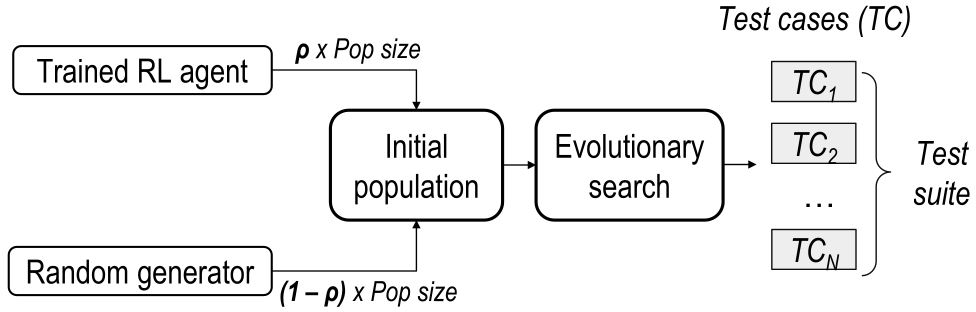}
\centering
\caption{An overview of RIGAA approach for test scenario generation}
\label{fig:approach}
\end{figure}

In this section, we present RIGAA (Reinforcement learning Informed Genetic Algorithm for Autonomous systems testing), a novel evolutionary search-based approach that incorporates a pre-trained reinforcement learning (RL) agent to generate a percentage of the initial population for generating test scenarios for autonomous robotic systems. Figure \ref{fig:approach} shows an overview of the approach.

Unlike a traditional evolutionary algorithm (EA), which generates the initial population randomly or using heuristics such as Latin hypercube sampling to ensure a uniform distribution of input solutions \cite{mckay2000comparison}, we propose using a trained reinforcement learning (RL) agent to generate a certain percentage $\boldsymbol{\rho}$ of the initial population. The RL agent is trained to produce test scenarios with higher fitness than those generated by random generator, while maintaining a high level of diversity. Our hypothesis is that RL-informed initialization will allow the search to find fitter solutions with a smaller evaluation budget, as some previous problem knowledge is already embedded in them. Additionally, it should enable the discovery of new types and patterns of test scenarios.

In evolutionary search, individuals with higher fitness have a greater chance of being selected and propagated in the next iterations \cite{coello2007evolutionary}. Therefore, solutions produced by a trained RL agent have a higher probability of contributing to the algorithm evolution if their initial fitness is higher than that of randomly produced solutions. That is why an important requirement for the RL agent is to produce test scenarios of a higher fitness, than the random generator. Another requirement is to produce diverse test scenarios.

After initial population generation, RIGAA follows the typical steps of EA, such as selection, crossover and mutation. At the end of the search, $N$ individuals from the final generation are selected for inclusion in the final test suite. 
In the following subsections, we provide more details on the implementation of the evolutionary search, the training of the RL agent for test scenario generation, and the overall functioning of RIGAA.

\subsection{Surrogate reward}
The key novelty of the RIGAA approach is in introducing a pre-trained RL agent for generating some part of the initial population of the test scenarios. The agent is trained to produce diverse and challenging test scenarios. When training the agent we are interested in evaluating to what extent the scenario produced by the RL agent falsifies the system under test. However, as mentioned previously, running evaluations of the simulator-based model is computationally expensive. Each evaluation can take from tens of seconds to minutes. In order for the RL agent to be trained, it needs to interact with the environment for a sufficiently large number of steps. Model-free RL algorithms, such as state-of-the-art PPO algorithm, are sample inefficient \cite{sutton2018reinforcement}. They require a lot of samples (sometimes millions of interactions) for the agent to learn a good policy \cite{stable-baselines3}.
In order to reduce the time to perform the evaluations, we suggest using a surrogate reward, based on a simplified approximation of the system performance. The surrogate reward is a proxy of a true reward function evaluated with the simulator-based system model.

\subsubsection{Surrogate reward function for maze generation}
The success of an autonomous robot tasked with navigating to a target location depends on the performance of its planning and control algorithms. Longer paths to the goal location are generally more complex and contain more intricate challenges, requiring the robot's algorithms to adapt and respond to changing conditions. By encouraging the RL agent to optimize the test scenario for longer paths, we can promote the generation of test scenarios that are more likely to expose weaknesses or limitations in the robot's planning and control algorithms. 
We use the Eq. (\ref{eq:reward_robot}) to calculate the surrogate reward for generating maze environments:
\begin{equation}\label{eq:reward_robot}
    R_s = \sum_{i=1}^{n-1} \sqrt{(x_{i+1}-x_i)^2 + (y_{i+1}-y_i)^2}
\end{equation}
where $n$ is the number of trajectory points, $(x_i, y_i)$ are the coordinates of the $i$-th point, and $R_s$ is the surrogate reward, corresponding to the length of the path in units of distance.
\begin{figure}[h!]
\centering
\begin{subfigure}{0.45\textwidth}
  \centering
  \includegraphics[scale=0.2]{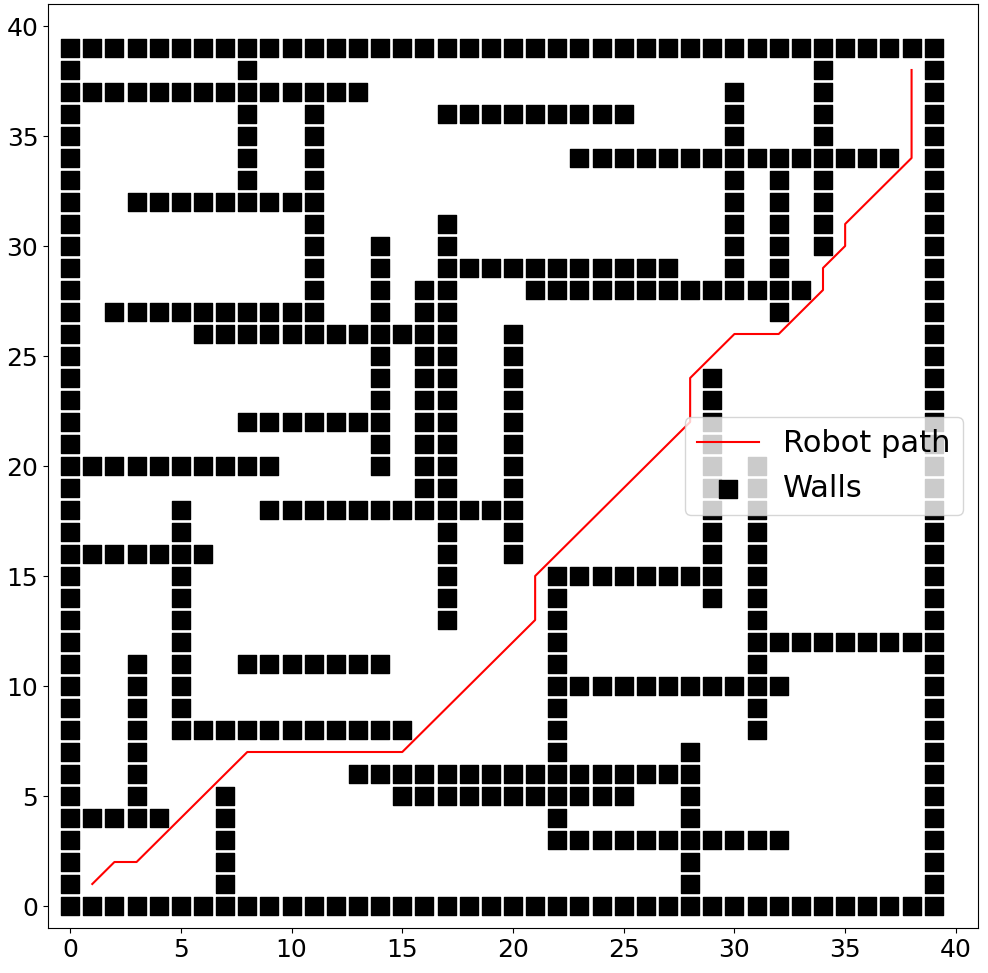}  
  \caption{Example of a test scenario with a lower reward of 58.18}
  \label{fig:low_rew}
\end{subfigure}
\begin{subfigure}{0.45\textwidth}
  \centering
  \includegraphics[scale=0.2]{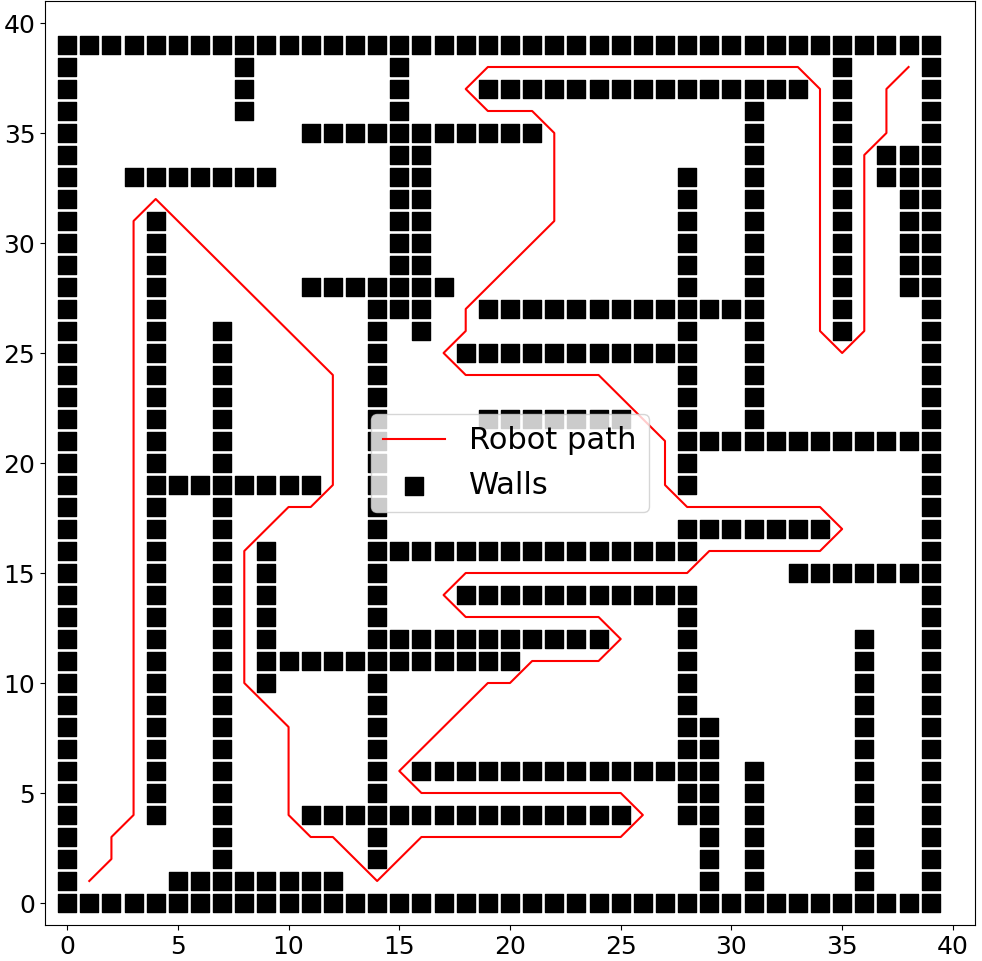}  
  \caption{Example of a test scenario with a higher reward of 220.36}
  \label{fig:high_rew}
\end{subfigure}
\caption{Example of the vehicle kinematic model trajectory (blue points) given the road topology (yellow points)}
\label{fig:s_reward_rob}
\end{figure}
An example of scenarios with low and high rewards is shown in Figure \ref{fig:s_reward_rob}. As you can see, the scenario with the higher reward contains more turns and a longer path to the target, which should present a more challenging environment to navigate.
\subsubsection{Surrogate reward function for road topology generation}
Vehicle kinematics is well studied and known  \cite{rajamani2011vehicle}. We suggest leveraging some knowledge of the vehicle kinematics as a proxy for the evaluation of the vehicle trajectory on a given road topology.
We implement a vehicle kinematic model \cite{campion1996structural} along with a simplified version of the pure pursuit controller \cite{coulter1992implementation} based on the following equations:
\begin{align*}
\delta_{k+1} &= \arctan(\frac{dy}{dx}) - \delta_k \\
\theta_{k+1} &= \theta_k + (\delta_{k+1})\Delta t \\
x_{k+1} &= x_k + v_k\cos(\theta_k)\Delta t \\
y_{k+1} &= y_k + v_k\sin(\theta_k)\Delta t \\
v_{k+1} &= v_k + a_k\Delta t 
\end{align*}
where $x_k$, $y_k$, $\theta_k$, and $v_k$ represent the current position of the center of the vehicle, orientation, and speed of the vehicle, respectively, at time $k$. Parameter $\Delta t$ is the time step between updates, $\delta_k$ is the steering angle at time $k$, $a_0$ is the acceleration of the vehicle. Values $dy$ and $dx$ are the distances of the vehicle from the $y$ and $x$ coordinates of the closest point on the road lane. In our experiments, we set the model parameters as shown in Table \ref{tab:params}. We selected the parameters empirically with the goal to maximize the similarity of the trajectories extracted from the simulator-based model logs and the trajectory predicted with the vehicle kinematic model. We evaluated the trajectory similarity based on the Hausdorf distance \cite{huttenlocher1993comparing}. \input{tables/params}

When running the vehicle model with the controler on a given road topology we record the vehicle location and the corresponding closest point defining the road topology at each timestep $k$. As a result we obtain the set of points defining the vehicle model trajectory $P_{veh}$ and a set of corresponding points from the road topology polyline $P_{rd}$. 

\begin{figure}[h!]
\begin{subfigure}{0.5\textwidth}
  \centering
  \includegraphics[scale=0.25]{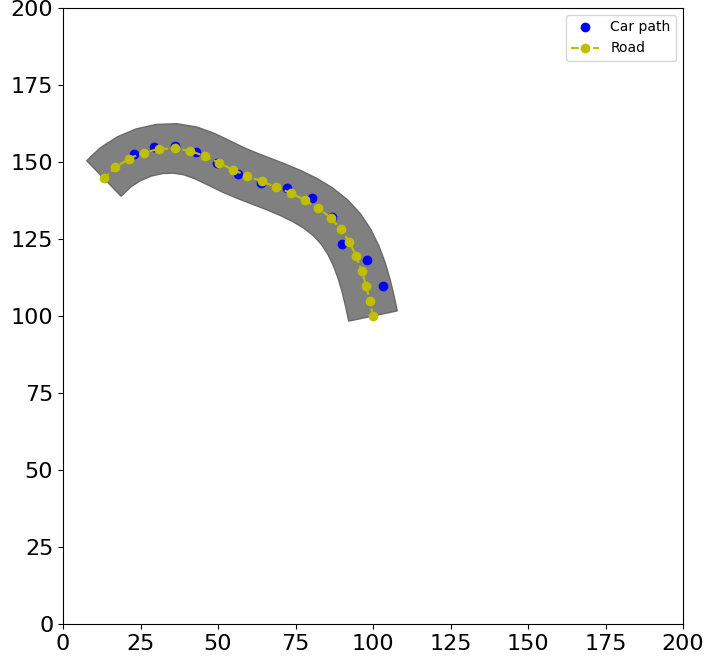}  
  \caption{Example of a test scenario with a lower reward of 2.13}
  \label{fig:low_rew}
\end{subfigure}
\begin{subfigure}{0.45\textwidth}
  \centering
  \includegraphics[scale=0.25]{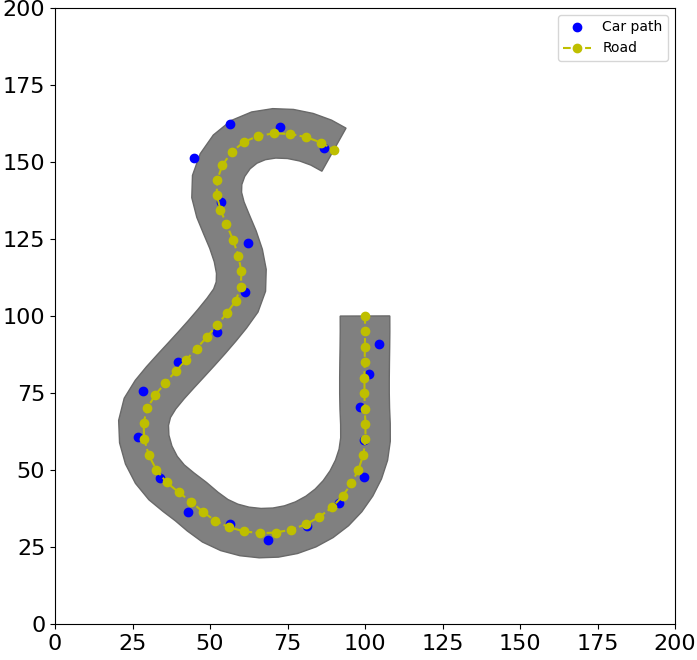}  
  \caption{Example of a test scenario with a higher reward of 9.56}
  \label{fig:high_rew}
\end{subfigure}
\caption{Example of the vehicle kinematic model trajectory (blue points) given the road topology (yellow points)}
\label{fig:s_reward}
\end{figure}

We evaluate the Euclidean distance between the $i$-th point in $P_{rd}$ and the $i$-th point in $P_{veh}$, corresponding to the vehicle deviation from the road as:
\begin{align*}
d_i = \sqrt{(P_{rd_x,i} - P_{veh_x,i})^2 + (P_{rd_y,i} - P_{veh_y,i})^2}
\end{align*}
The surrogate reward is calculated as the maximum deviation of the vehicle from the road points:
\begin{align} \label{eq:reward_vehicle}
    R_{s} = \max_{i=1}^{n} d_i
\end{align}
 where $n$ represents the number of points in the sets $P_{rd}$ and $P_{veh}$.
Examples of test scenarios with low and high rewards are shown in Figure \ref{fig:s_reward}. We can see the road topology with a higher reward contains more turns and thus should be more challenging for the driving agent.

\subsection{RL agent training}
In the RIGAA approach, another important step is training a reinforcement learning agent to generate test scenarios. We represent test scenarios a sequence of discrete scenario elements $E$. The main goal of the RL agent is to learn to generate sequences of the scenario elements $E$, such that valid and challenging test scenarios are produced. 
In the following paragraphs we describe in more detail how we represent the test scenario generation problem as a Markov decision process (MDP), defined by the 5-tuple $< S, A, R, P, \rho_0>$, and how the RL agent is trained.

\textit{State representation}. We represent the state as a 2D integer array, such as described in Table \ref{tab:repr}. Each column corresponds to some discrete part of the test scenario. At initialization, the agent starts from a 2D array, where the first column (first scenario element) contains attribute values randomly sampled from the allowable range of values. The rest of cells are initialized with zero values. In the next steps, the goal of the agent is to learn to select such attribute values $A_{1em}$, $A_{2em}$, $A_{nem}$ for the elements $E_m$ that results in producing a challenging test scenario. 

For an autonomous robot, the state corresponds to a 2D map with obstacles. The state array size is 40 x 3 elements, where 3 corresponds to the number of attributes describing each scenario element (wall type, wall position, wall size).

For an autonomous vehicle LKAS system the state corresponds to a 2D array defining the road topology. The size of the state array size is 30 x 3. Here 30 corresponds to maximum number of scenario elements i.e. road segments. From our experience, the generated test scenarios rarely exceed 30 road segments, that is why we fixed this number to 30. If the scenario contains less, than 30 elements, the rest is padded with zeros. The second dimension corresponds to the number of attributes, which is three (road segment type, road segment length and turning angle).

\textit{Action space}. At each time step $i$, the agent chooses an action that defines the attribute values of scenario element $E_i$. The action is multidimensional, with the number of dimensions equal to the number of attributes. For both test generation agents, the action space has three dimensions.

For the autonomous robot, the first dimension corresponds to the type of obstacle, the second to the obstacle's position, and the third to the obstacle's size. The resulting action space was of size $2x7x37$. For the autonomous vehicle, the first dimension corresponds to the type of road segment, the second to the road segment's length, and the third to the turning angle of the road. The resulting action space was of size $3x25x35$. 

\textit{Reward function.}
During training, the RL agent receives a reward $r_t \in \mathbb{R}$ every time it executes
an action $a_t$. We define the following reward function: 

\begin{equation}
  r_{i} =
    \begin{cases}
      R_s + R_1 + R_2 + R_3 & \text{if test scenario is valid,}\\
      R_4 & \text{if test scenario is invalid,}\\
    \end{cases}       
\end{equation}

where $R_s$ represents the surrogate reward of the agent, $R_1$ denotes the reward improvement score, $R_3$ provides a bonus reward for discovering a good test scenario and $R_2$ provides an additional reward for exploring the search space. $R_4$ represents a negative reward given for producing an invalid test scenario.
The improvement score $R_1$ is a reward given for an action, that leads to a better test scenario, compared to the previous action. We calculate it as $R_1$ = ($R_{s_t}$ - $R_{s_t-1}$), where $R_{s_t}$ is $R_s$ at the current step and $R_{s_t-1}$ is the $R_s$ reward in the previous step. Such reward encourages the RL agent to  seek actions that result in improved test scenarios.
The bonus reward $R_2$ is an additional reward given to the agent if the surrogate reward $R_s$ it received exceeds a certain positive threshold $th$. If the threshold value is not reached, we set it to zero. In our experiments, without such a reward, the agent took more time to discover more challenging scenarios, that have high $R_s$.
Reward $R_3$ for exploring the search space is given when a previously unused scenario element is used e.g., a new road segment configuration, different from the ones used before. This promotes diversity in the created scenarios. 
We give a negative reward $R_4$ when the scenario does not satisfy the problem constraints.

For an autonomous robot, the surrogate reward $R_0$ is defined in Eq. (\ref{eq:reward_robot}) as the length of the path it needs to travel to the goal. Intuitively, the longer the path it takes, the more challenging the scenario and the higher risk of the robot failing its mission. 
We fix the threshold $th$ for obtaining the bonus reward at 110 i.e. the agent should achieve the reward of 110 to get an additional reward $R_2$. We fix $R_2$ value equal to $R_s$.
We also give a reward of $R_3 = 10$ if the action taken specifies an obstacle position or size, not used before. 
We choose the $R_4$ value to be -100.

For an autonomous vehicle LKAS system, the surrogate reward $R_s$ is defined in Eq. (\ref{eq:reward_vehicle}) and represents the vehicle deviation from the road lane center.  The bonus reward of $R_2 = R_s$ is given if the surrogate reward $R_s$ exceeds the value of 5. Additionally, a reward for exploration of $R_3 = 1$ is provided if the action taken specifies a segment length or turning angle that has not been used before. The reward for invalid test scenario $R_4$ is fixed at -50.

\textit{Transition probability function.} The transition function $P$ determines the probability of transitioning to a new state given the current state and action. In our case, this function is deterministic. Every time the action taken defines specific attribute values of the next scenario element. 

\textit{Initial state distribution.} At initialization, the agent
starts from a state i.e. 2D array, where the first column (first scenario element) contains randomly sampled attribute values.

\textit{Episode termination.}
We terminate the episode when the agent provides values for all the scenario elements or after it produces an invalid scenario. For an autonomous robot, we had 40 scenario elements. One of them was initialized at the start, so the agent had to complete 39 steps more to set all the scenario elements and complete the episode. The episode could also be terminated if the agent produced a map not satisfying the constraints.
For an autonomous vehicle LKAS system, we set the maximum number of scenario elements to be 30. The episode was terminated either after 30 steps or when the agent produced an invalid road topology.

\textit{Reinforcement learning algorithm.}
For training, we use the state-of-the-art Proximal Policy Optimization (PPO) algorithm \cite{schulman2017proximal}, implemented in the Stable Baselines framework \cite{stable-baselines}.
PPO is an actor-critic algorithm that consists of two key components: the actor and the critic neural network. The actor network is responsible for selecting actions based on the current state, while the critic network evaluates the chosen actions and provides feedback on their quality. This feedback is then used to guide the actor in adjusting its decision-making process \cite{konda1999actor}. In our study, we use the default multilayer perceptron (MLP) architecture for both the actor and critic, which consists of two layers of 64 neurons. In this case, an observation (which is also an input to the actor network) is a one-dimensional vector that corresponds to the flattened state array. We also experimented with using a convolutional neural network (CNN) for the agent actor-critic architecture, where the observations were images. However, we found that using a multi-layer perceptron (MLP) based architecture resulted in better and more diverse results.

We used the default hyperparameters suggested by the framework, as they showed good stability of training. These hyperparameters include the learning rate of 0.0003, 2048 steps and batch size of 64. One hyperparameter we found was important to fine-tune is the entropy coefficient, which plays an important role in balancing the exploration and exploitation trade-off during training. By default, it is set to 0, resulting in minimal exploration of the action space. In our experiments, with the entropy coefficient set to 0, the agent tended to exploit a particular sequence of actions and the generated scenarios were of low diversity. However, if this value was set too high, the training became unstable, with occasional failures of the agent to find a good policy and outperform the random generator. Empirically, we fixed the entropy coefficient value of 0.005 for both maze and road topology generation agents, which allowed producing diverse, but at the same time challenging test scenarios. 

\subsection{Genetic algorithm implementation}\label{sec:ga}

A key component of RIGAA is the use of evolutionary search, specifically the NSGA-II genetic algorithm. In this subsection, we describe the individual representation, objective functions, crossover, and mutation operators used in our implementation of NSGA-II.

\textit{Individual representation.}
In our representation of the individuals, or chromosomes, we use a two-dimensional array as shown in Table \ref{tab:repr}. Each chromosome corresponds to an array with columns representing the scenario elements and rows containing attributes describing them. 

For autonomous robots, the individual size is fixed at 40 x 3. For autonomous vehicles, the individual size is flexible and is of $N x 3$, subject to the constraint that the resulting test scenario must fit within the bounds of the map (200 m x 200 m).

\textit{Fitness function.} Our framework utilizes a multi-objective search algorithm, NSGA-II \cite{deb2013evolutionary}, to optimize the test suite for both scenario fault revealing power and diversity. The first objective function, $F_1$, measures the extent to which the system falsifies its requirements when executing a given scenario. We can define the requirements for the system in terms of Metric Temporal Logic, which is common in CPS testing literature. MTL is a formalism that enables system engineers to
express complex design requirements in a formal logic \cite{hoxha2014towards}.
Then we are interested in measuring the degree to which this requirement can be satisfied. To this end, in CPS testing, a robustness metric is used, which aims to provide a numerical evaluation
of the degree of the requirement satisfaction \cite{tuncali2018simulation}. We evaluate the objective $F_1$ according to the following equation:
\begin{align}\label{eq:mtl}
F_1 = \rho_{\text{safe}} (\mathcal{G}_{[0, T]} (\phi) ),
\end{align}
 where $\mathcal{G}_{[0,T)}$ denotes that globally i.e. at each time interval $t \in$ $[0,T]$, the condition $\phi$ within the parentheses, which must always hold true. The function $\rho_{\text{safe}}$ is a robustness function that returns a value representing the degree to which the MTL property is satisfied. If the property is not satisfied, it returns a negative real value corresponding to the degree to which the property is violated. If the property is satisfied, it returns a positive value.

The $\phi$ function is problem-specific and can be tailored to the specific system being tested.
For example, in the case of an autonomous robot, we define the following function $\phi$:
\begin{align}
    \phi = \lnot\pi_{coll} \land \lnot\pi_{fall}, 
\end{align}
where $\pi_{coll}$ is an event of colliding with an object and $\pi_{fall}$ is an event of falling down. The $\phi$ function states that the robot should not collide with objects and should not fall. Then the robustness function $\rho_{safe}$ evaluates the proportion of waypoints the robot completed, out of all the waypoints that define the path to the target location.
It returns the negative value corresponding to the proportion of completed waypoints. If all the waypoints were completed, it returns a positive value.
Intuitively, the fewer waypoints are completed by the robot, the less successful it is in its mission of reaching the target.

For an autonomous vehicle LKAS system, $\phi$  is defined as follows: 
\begin{align}
   \phi = |dist(t)| \leq \delta,
\end{align}
where $dist$ is an Euclidian distance from the nearest road center point and $\delta$ is the maximum allowable deviation from the road center. 
The $\phi$ function states that the vehicle should stay within certain road lane boundaries. The robustness function $\rho_{safe}$ returns the percentage of the vehicle area, that went out of the road lane taken with a negative sign.

To promote diversity in the test suite, we also evaluate a second objective function, $F_2$, which measures the novelty of an individual scenario. It is defined as the average Jaccard distance of the scenario to $N$ individuals with the highest value of the first fitness function. In our experiments we used the value of $N=5$.  Jaccard distance is a well known metric for evaluating the dissimilarity between two sets \cite{chung2019jaccard}.  The value of $F_2$ objective ranges from 0 to 1.  This function is inspired by novelty search algorithms \cite{lehman2011evolving}, where the solution is compared to the archive of the best solutions. Below we provide more details about calculating the Jaccard distance and the $F_2$ objective function.

The Jaccard distance, or diversity $D$, between two test cases is calculated as follows:
\begin{equation}\label{jaccard}
    D = 1 - \frac{|TS_1 \cap TS_{2}|}{|TS_1 \cup TS_{2}|} 
\end{equation}
Here $|TS_1 \cap TS_{2}|$ represents the number of scenario elements that are shared between two test scenarios, $TS_1$ and $TS_2$ i.e. belong to the intersection set between two test scenarios. Scenario elements are added to the intersection set if their attributes are sufficiently similar, as defined by thresholds $th_0$, $th_1$, $th_n$ for attributes $A_0$, $A_1$, $A_n$, respectively. The thresholds should be selected based on the domain knowledge of the problem. Two scenario elements are considered similar if all of their corresponding attributes are similar to one another.
We calculate the number of scenario elements in the union of two sets, $TS_1$ and $TS_2$, using the Eq. (\ref{eq:union}):

\begin{equation} \label{eq:union}
| TS_1 \cup TS_{2} | = |TS_1| + |TS_2| - |TS_1 \cap TS_2|
\end{equation}

where $|TS_1|$ is the number of elements in $TS_1$, $|TS_2|$ is the number of elements in $TS_2$, and $|TS_1 \cap TS_2|$ is the number of scenario elements that are common to both $TS_1$ and $TS_2$.

Finally, to evaluate the $F_2$ objective of the individual, as mentioned above, we calculate the average diversity between the individual and 5 individuals with the highest $F_1$ fitness function in the population, as shown in Eq. (\ref{eq:diversity}):
\begin{align} \label{eq:diversity}
F_{2} = \frac{1}{N}\sum_{i=1}^{N} D(tc_0,tc_i),
\end{align}
where $tc_0$ represents the test scenario, for which we evaluate the fitness, and $tc_i$ represents one of the $N$ best scenarios in terms of the $F_1$ function discovered by the algorithm.


In our implementation, we aim to minimize $F_1$ and maximize $F_2$. The Pymoo framework \cite{pymoo} that we use for genetic algorithm implementation, only supports function minimization. As a result, we minimize the negative value $F_2$ instead.

\begin{figure}[h]
\includegraphics[scale=0.6]{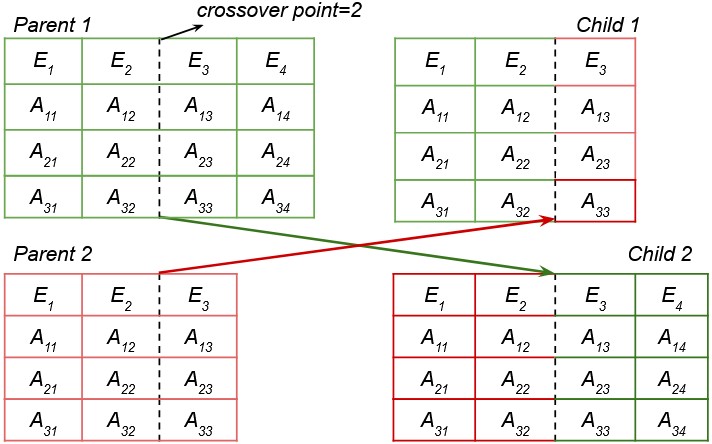}
\caption{An example of the crossover operator functioning, crossover point is selected equal to 2}
\label{fig:cross}
\end{figure}

\textit{Crossover operator.} We are using a one point crossover operator, which is one of the commonly used operators for variable-length solution representation. An example of applying this operator is shown in Figure \ref{fig:cross}. This operator creates two new test cases by exchanging scenario elements between two existing test scenarios. 

\begin{figure}[h!]
\includegraphics[scale=0.6]{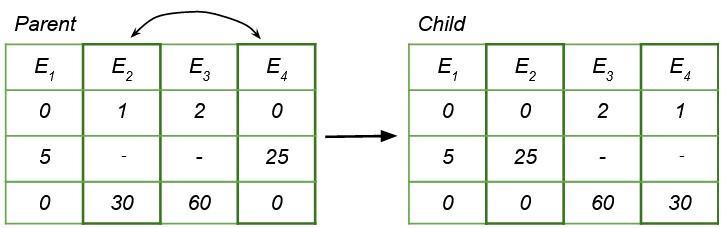}
\caption{An example of the exchange mutation operator, where segments 2 and 4 are exchanged}
\label{fig:mut1}
\end{figure}

\begin{figure}[h!]
\includegraphics[scale=0.6]{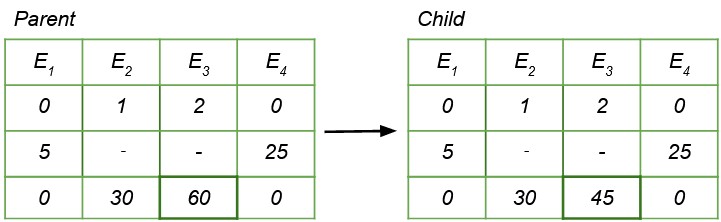}
\caption{An example of the change of variable mutation operator, where the value of the third attribute of the third segment is changed}
\label{fig:mut2}
\end{figure}

\textit{Mutation operators.} We define two mutation operators: exchange and change of variable operator. For exchange operator, the attributes of the randomly selected scenario elements $E_m$ are exchanged the positions. In change of variable operator, scenario element $E_m$ in a chromosome is randomly selected. Then the value of one of its attributes $A_{nem}$  is changed according to its type and maximum as well as minimum value. Examples of applying these operators are shown the Figure \ref{fig:mut1} and Figure \ref{fig:mut2} respectively.

\textit{Duplicate removal.} At each iteration, we remove the individuals which have small diversity $D$ (see Eq. \ref{jaccard}) between them. We fix the $D$ threshold to be 0.2. If the diversity between the two individuals is less than 0.2, one of them is removed. 

\subsection{RIGAA algorithm implementation}
The main steps of RIGAA are outlined in Algorithm \ref{alg:rigaa}, and the value ranges of the main hyperparameters are shown in Table \ref{tab:hyper_param}. 
\input{tables/param_rigaa.tex}

As every evolutionary algorithm, RIGAA starts by creating an initial population of the size $N_{pop}$. Differently from existing initial population sampling techniques, in RIGAA some percentage $\rho$ of individuals is generated by a pre-trained RL agent and the other by a random generator. 
After initial population generation, RIGAA is following the steps of the NSGA-II algorithm with objectives that we formulated in the sub-section \ref{sec:ga}. At each iteration, tournament selection is used to choose the individuals for crossover and mutation which are then performed on individuals with probabilities $CROS$ and $MUT$ respectively. 
The search is continued until a termination criterion is met. Finally, $N_{TC}$ best individuals are selected from the final population to be added to the test suite. In our case, we used the number of evaluations or the running time as the termination criteria.

In the following section, we assess the performance of RIGAA by comparing its convergence, fitness, and diversity of solutions with other algorithms. Specifically, we compare it with our previous implementation of NSGA-II-based search, also known as the AmbieGen approach \cite{humeniuk2022search}, as well as random search.

\input{algorithms/rigaa.tex}


\section{Evaluation} \label{sec:evaluation}
In this section, we formulate and answer the research questions related to the performance of RIGAA. Our primary objective is to assess the extent to which the RL-based initialization enhances the performance of a multi-objective evolutionary algorithm (MOEA) in comparison to the random initialization. Additionally, we aim to explore which RIGAA configuration yields the optimal performance.

When comparing the algorithms, we assess the following metrics: the average fitness and diversity of the solutions in the test suite produced by the algorithm, as well as the highest fitness value $F_{1max}$ of a solution  found within a fixed evaluation budget of $N_{eval}$ evaluations or a fixed time budget.  The fitness value is directly proportional to the extent of safety requirement falsification by the test scenario. The test suite is composed of the $N=30$ best solutions (based on non-dominance rank) obtained by the algorithm.

We evaluate the average diversity between all the pairs of test scenarios in a test suite according to the following equation:
\begin{equation}
D_{av} = \frac{2}{N(N-1)}\sum_{i=1}^{N-1}\sum_{j=i+1}^{N} D(tc_i,tc_j),
\end{equation}
where $\frac{N(N-1)}{2}$ is the total number of pairs in the test suite, $N$  is the test suite size, $D$ is a diversity metric evaluated between two test scenarios $tc_i$ and $tc_j$ according to Eq. (\ref{jaccard}).

We evaluate the average fitness as:
\begin{equation}\label{eq:fit_full}
    F_{av} = \frac{1}{N}\cdot\sum_{i=1}^{N}|F_{1i}|,
\end{equation}
where $N$ is the test suite size, $F_{1i}$ is the value of $F_1$ fitness function of a test scenario $i$.

For some experiments, where we select the optimal RIGAA configuration, we used a surrogate reward function to evaluate the fitness of the scenarios to reduce the computational time. In this case, the average surrogate fitness $F_{avs}$ is evaluated as: 
\begin{equation}\label{eq:fit_simple}
    F_{avs} = \frac{1}{N}\cdot\sum_{i=1}^{N}|R_{si}|,
\end{equation}
where $R_s$ is the value of a surrogate reward function of a test scenario and $N$ is a test suite size.

We repeat all evaluations a statistically significant number of times, i.e. at least 10 when using the simulator and at least 30 otherwise. For all results, we report the non-parametric Mann-Whitney U test with a significance level of $\alpha = 0.05$, as well as the effect size measure in terms of Cliff's delta \cite{mann1947test,macbeth2011cliff}.

We run our experiments on a PC with a 16-core AMD Ryzen 7 4800HS CPU @ 2.90 GHz, 16 GB of Memory, and an NVidia GeForce GTX 1660 GPU @ 6GB.

\subsection{Research questions}


In the first part of our experiments, we evaluate the performance of the trained RL agents by measuring the average diversity ($D_{av}$) and average fitness ($F_{av}$) of the generated test scenarios for an autonomous robot and an autonomous vehicle. We compare these metrics to those obtained using a random generator. Our expectation is that the RL agent will produce scenarios of higher fitness than a random generator while maintaining a similar level of diversity in the produced scenarios. 
In the second part, we aim to identify the optimal configuration of RIGAA and compare it with the baseline MOEA (MOEA with random initialization). We test RIGAA with different values of the initial percentage of RL-produced solutions, as well as with two different MOEAs. To reduce evaluation time and complexity, we measure the quality of RIGAA-produced test scenarios based on a surrogate reward function.
In the third part of our experiments, we evaluate the performance of RIGAA in terms of its ability to falsify simulator-based autonomous robotic systems. Specifically, we consider such systems as an autonomous vehicle and an ant robot.
  We formulate the following research questions:
\begin{enumerate}
     \item \textbf{RQ1} (Comparing the performance of the RL-based test generator and random test generator).
\textit{How do the performance metrics of the RL agent and random generator compare in terms of the average fitness and diversity of the test scenarios produced, as well as the generation time?}

In $RQ1$ we aim to assess whether the trained RL agent produces scenarios of higher fitness, than the random parameter initialization. We evaluate the fitness of the solutions based on both the surrogate reward and the fitness of the scenarios when applied to the simulator-based models of the systems under test. Moreover, we aim to evaluate if the RL agent can generate solutions with high diversity. If it is the case, we can argue that the RL agent can indeed provide a better initialization for the evolutionary algorithm, than a random generator. 

    \item \textbf{RQ2} (Selecting the $\rho$ hyperparameter of the RIGAA algorithm).
   \textit{What is the effect of different values of $\rho$ on the final fitness and diversity of the test suite?}

The aim of RQ2 is to investigate the impact of the percentage $\rho$ of RL-produced solutions in the initial population on the quality and diversity of the resulting test scenarios in the test suite. We evaluate RIGAA using five different values of the $\rho$ hyperparameter, namely $\rho \in [0.2, 0.4, 0.6, 0.8, 1]$, and select the optimal value based on the resulting fitness and diversity of the test scenarios.

    \item \textbf{RQ3} (Comparing  RIGAA and randomly initialized MOEA).
 \textit{How do different configurations of RIGAA compare to each other, baseline MOEA and the random search?}
 
In $RQ3$, our objective is to evaluate the effectiveness of the RL-based initialization in guiding an evolutionary algorithm and improving its performance. For this purpose, we compare different RIGAA configurations based on two MOEAs, namely NSGA-II \cite{deb2002fast} and another multiobjective optimization algorithm SMS-EMOA\cite{beume2007sms} to the corresponding MOEA configurations with the random initialization and random search. We selected NSGA-II as it is one of the most popular MOEA to be used for search-based software engineering problems \cite{harman2001search}. We select SMS-EMOA as another type of MOEA, having a different selection algorithm. It differs from NSGA-II in promoting the dominated hypervolume  rather than the crowding distance of solutions. SMS-EMOA allows finding solutions with a good compromise of the search objectives and a well-distributed Pareto front, which is promising for test scenario generation as we aim to maximize the diversity of solutions i.e. the test scenarios. To reduce evaluation time and
complexity, we utilize a surrogate reward function to measure the fitness of scenarios produced by the algorithms we analyze.

 \item \textbf{RQ4} (Effectiveness of RIGAA for simulator-based testing of autonomous robotic systems).
 \textit{To what extent can RIGAA improve the identification of higher fitness and more diverse test scenarios for autonomous robotic systems compared to baseline approaches?}
 
In $RQ4$ we compare the average test suite fitness and diversity as well as the best test scenarios found for the two simulator-based models of the systems under test. We also compare the RIGAA performance in terms of number of failures found and their diversity with state-of-the-art tools for vehicle LKAS system testing as AmbieGen \cite{humeniuk2022search} and Frenetic \cite{castellano2021frenetic}.

 \end{enumerate}

\subsection{Subjects of the study}
In our study, we consider two simulator-based autonomous robotic systems: an autonomous ant robot in the Mujoco simulator \cite{todorov2012mujoco} as well as an autonomous vehicle in BeamNG simulator \cite{beamng}.
\begin{figure}[h!]
\begin{subfigure}{0.45\textwidth}
  \centering
  \includegraphics[scale=0.21]{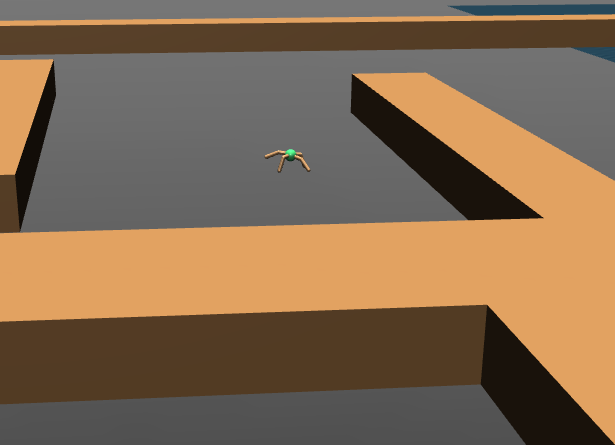}  
  \caption{Autonomous ant robot in Mujoco simulator}
  \label{fig:rob_agent_rl}
\end{subfigure}
\begin{subfigure}{0.45\textwidth}
  \centering
  \includegraphics[scale=0.35]{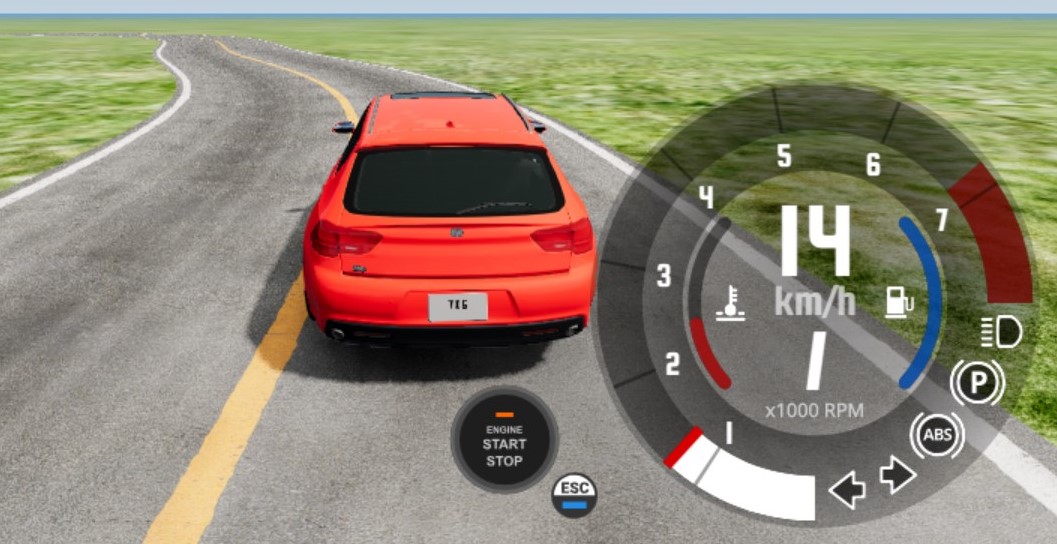}  
  \caption{Autonomous vehicle in BeamNG simulator}
  \label{fig:veh_beam}
\end{subfigure}
\caption{Autonomous ant and autonomous vehicle systems used in our study}
\label{fig:agents}
\end{figure}

The autonomous ant robot, shown in Figure \ref{fig:rob_agent_rl} is an 8-DoF (degrees-of-freedom) “Ant” quadruped robot designed for research on RL-based agent by Schulman et al. \cite{schulman2015high}. This robot consists of one torso (free rotational body) with four legs attached to it with each leg having two links. The goal of the RL algorithm is to coordinate the four legs to move in the forward (right) direction by applying torques on the eight hinges connecting the two links of each leg and the torso. We utilize the Ant robot pre-trained with an RL policy from the D4RL benchmark \cite{fu2020d4rl} for locomotion. Additionally, we employ a maze environment provided by the authors of D4RL benchmark, which requires the autonomous ant to navigate towards goal locations. In our experiments, we modified the environment to create larger maze maps and evaluate the behavior of the agent.

As the second test subject, we used an autonomous driving agent BeamNG.AI, built-in in the BeamNG.tech \cite{beamng} simulator, shown in Figure \ref{fig:veh_beam}. The goal of the driving agent is to navigate a given road, without going out of its bounds.
It has knowledge of the geometry of the entire road and uses a complex optimization process to plan trajectories that keep the ego-car as close as possible to the speed limit while staying inside the lane as much as possible. 

\subsection{Results}
In this subsection, we present our experimental results for each RQ formulated above.
\subsubsection{RQ1 – Comparing the performance of the RL-based test generator and random test
generator}
We trained two types of RL agents: one to generate virtual mazes for autonomous robot testing, and the other to generate road topologies for vehicle LKAS system testing. To demonstrate the stability of our training approach, we trained five different RL agents of each type. 
\begin{figure}[h]
\begin{subfigure}{0.48\textwidth}
  \centering
  \includegraphics[scale=0.4]{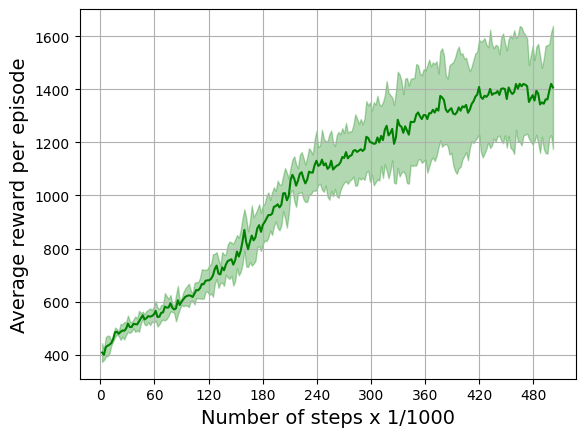}  
  \caption{Autonomous robot}
  \label{fig:rob_rl_conv}
\end{subfigure}
\begin{subfigure}{0.48\textwidth}
  \centering
  \includegraphics[scale=0.4]{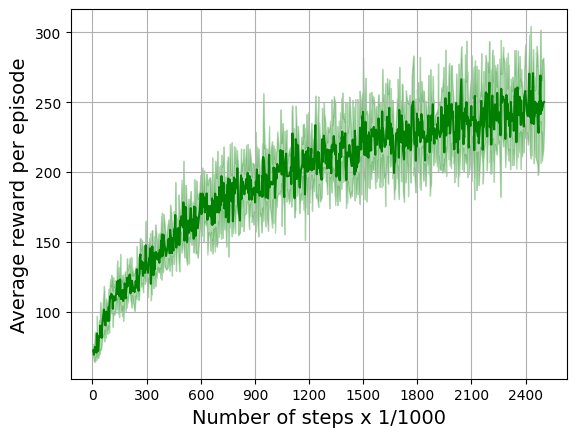}  
  \caption{Autonomous vehicle}
  \label{fig:veh_rl_conv}
\end{subfigure}
\caption{Average reward per episode during RL agents training}
\label{fig:rl_agent_conv}
\end{figure}
Figure \ref{fig:rl_agent_conv} illustrates the mean reward per episode during training, averaged across the five agents. We used a batch size of 64, a learning rate of 0.003 and an entropy coefficient of 0.005 with all other hyperparameters set to their default values as specified in the Stable Baselines 3 implementation \cite{stable-baselines3}. We trained the agents for autonomous robot testing for 500,000 steps and the agents for autonomous vehicle testing for 2,500,000 steps. After training the RL agents, we selected one of them to compare its performance to that of a random generator. All the five trained agents achieved similar levels of test scenario fitness and diversity.

We generated 30 different test suites containing 30 test scenarios with both the RL agent and the random generator. 
First, we evaluated the test scenarios in terms of the surrogate reward function from Eq. (\ref{eq:fit_simple}) as well as the average diversity of the scenarios in each test suite as in Eq. (\ref{eq:fit_full}).

\begin{figure}[h]
\centering
\begin{subfigure}{0.4\textwidth}
  \includegraphics[scale=0.35]{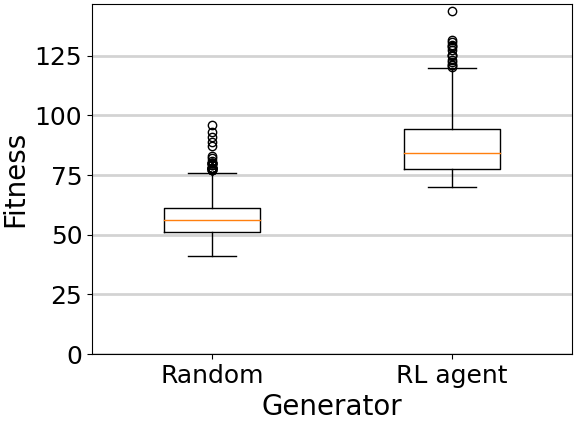}  
  \caption{Average fitness }
  \label{fig:fit_rob_gen}
\end{subfigure}
\begin{subfigure}{0.4\textwidth}
  \includegraphics[scale=0.35]{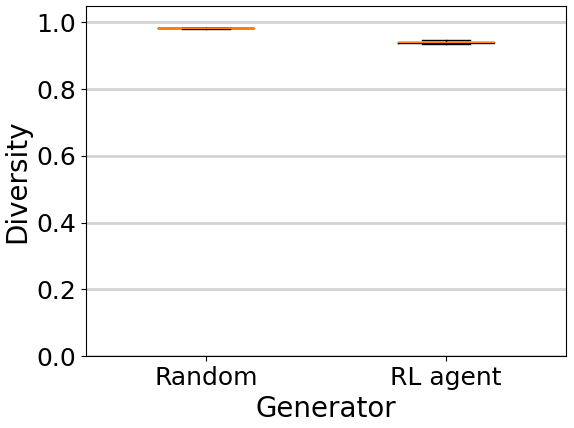}  
  \caption{Average diversity}
  \label{fig:div_rob_gen}
\end{subfigure}
\caption{Average fitness and diversity of the test scenarios for autonomous robot produced randomly (left box) and by the RL agent (right box)}
\label{fig:rob_gen}
\end{figure}

\begin{figure}[h]
\centering
\begin{subfigure}{0.45\textwidth}
  \includegraphics[scale=0.35]{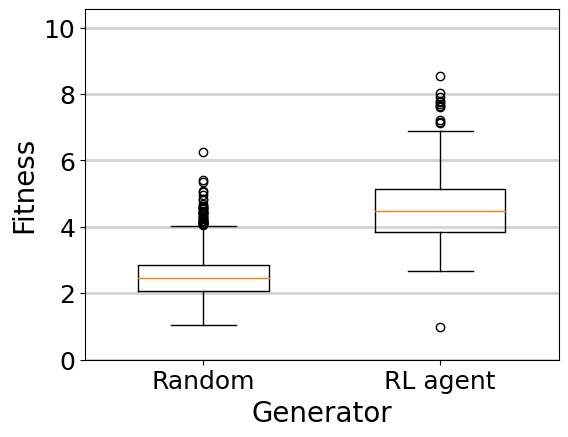}  
  \caption{Average fitness }
  \label{fig:fit_veh_gen}
\end{subfigure}
\begin{subfigure}{0.4\textwidth}
  \includegraphics[scale=0.35]{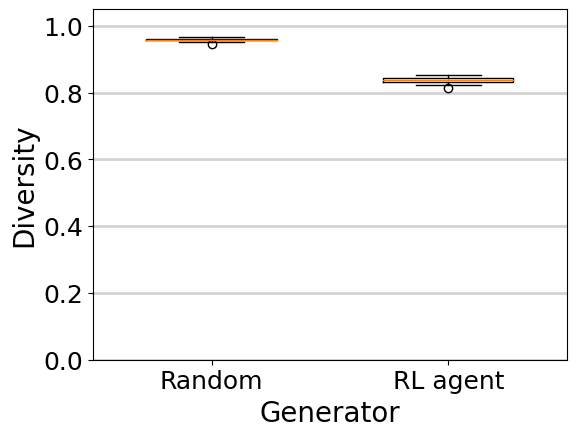}  
  \caption{Average diversity  }
  \label{fig:div_veh_gen}
\end{subfigure}
\caption{Average fitness and diversity of the test scenarios for autonomous vehicle LKAS system produced randomly (left box) and by the RL agent (right box)}
\label{fig:veh_gen}
\end{figure}

The box plots of the average fitness and diversity of the test scenarios for an autonomous robot are shown in Figure \ref{fig:fit_rob_gen} and Figure \ref{fig:div_rob_gen}. 
The box plots of the average fitness and diversity of the road topology test scenarios for an autonomous vehicle are shown in Figure \ref{fig:fit_veh_gen} and Figure \ref{fig:div_veh_gen}. 

Table \ref{tab:ts_gen} summarizes the obtained values for the average test suite fitness and diversity for the considered problems and generators. As we can see, all the results are statistically significant and with a large effect size. The RL agent outperforms the random generator in terms of the surrogate reward function of the scenarios produced by 52.4\% for maze generation and 79.8\% for road topology generation. The random agent produces 4.46 \% more diverse mazes and 14.3 \% more diverse road topologies. Although the random generator creates more diverse road topologies, we find the diversity value of 0.839 of the RL generator is sufficiently high, indicating that, on average, each test scenario within a test suite differs by approximately 83.9\% from one another.

\input{tables/rq1_mean.tex}

Having obtained positive results for the performance of RL agents in terms of scenario diversity and fitness using surrogate reward functions, we then evaluate generators' ability to falsify simulator-based agents.

\begin{figure}[h!]
\centering
\begin{subfigure}{0.45\textwidth}
  \includegraphics[scale=0.35]{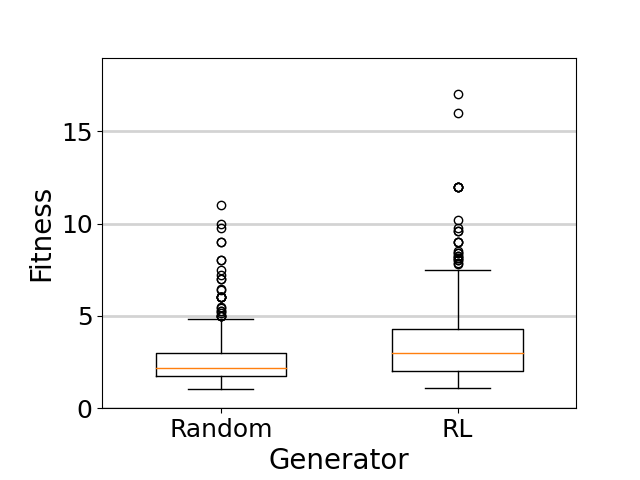}  
  \caption{Average scenario fitness for ant robot in simulator }
  \label{fig:fit_rob_sim}
\end{subfigure}
\begin{subfigure}{0.4\textwidth}
  \includegraphics[scale=0.35]{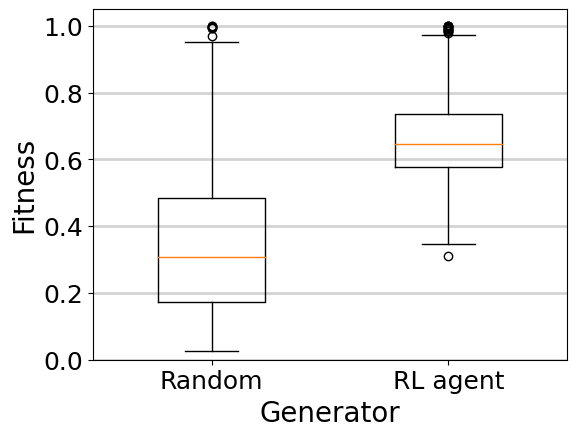}
  \caption{Average scenario fitness for an autonomous vehicle in simulator  }
  \label{fig:fit_rob_sim}
\end{subfigure}
\caption{Average fitness test scenarios for autonomous robot and vehicle LKAS system in the simulators}
\label{fig:gen_sim}
\end{figure}

We evaluate the test generators with the simulator-based models of the system under test, using the average fitness metric defined in Eq. (\ref{eq:fit_full}). We generated 10 test suites with 30 test scenarios each. For an autonomous robot, we report the inverse value of the $F_{1}$ function (defined in Eq. (\ref{eq:mtl})) for the better visualization purposes, where a higher value corresponds to a better fitness. Original fitness function evaluates the relative number of waypoints completed (lower is better). We average this metric over 3 executions of the agent in the same scenario, due to the stochasticity of the agent's behavior.
For an autonomous vehicle, the $F_1$ is evaluated as the maximum percentage of vehicle area that went out of bounds over the scenario execution.
The corresponding box plots for the average fitness value obtained are shown in Figure \ref{fig:gen_sim}.

The results are reported in Table \ref{tab:rq1_mean_fitness}. We can see that the differences in the fitness values are statistically significant for both systems under test.
\input{tables/rq1_mean2}
The RL agent produces 34.5 \% fitter scenarios for an autonomous ant robot and 88.5 \% fitter scenarios for an autonomous vehicle LKAS system.

\begin{figure}[h!]
\centering
\begin{subfigure}{0.45\textwidth}
  \includegraphics[scale=0.35]{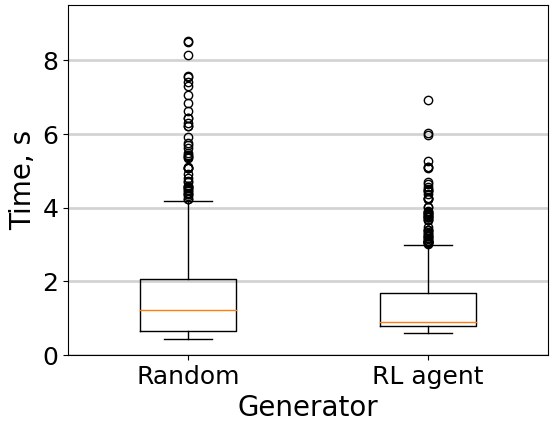}  
  \caption{Robot test scenario generation }
  \label{fig:time_rob_gen}
\end{subfigure}
\begin{subfigure}{0.4\textwidth}
  \includegraphics[scale=0.35]{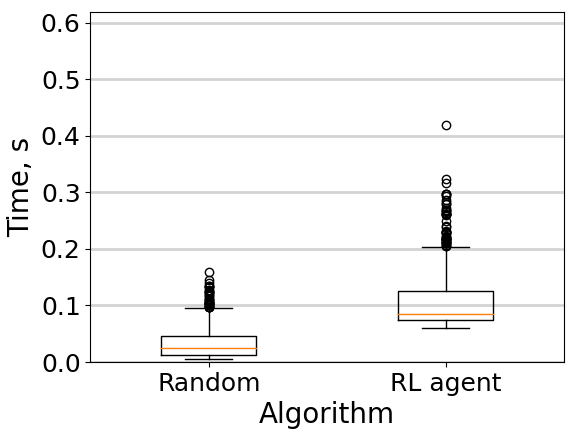}  
  \caption{Vehicle test scenario generation  }
  \label{fig:time_veh_gen}
\end{subfigure}
\caption{Average time to generate one test scenario by different generators }
\label{fig:time_gen}
\end{figure}

To evaluate the generators in terms of generation efficiency, we measure the average time taken to produce one solution while creating the test suites that we previously evaluated. The box plots with the generation time of the two RL agents and random generator are shown in Figure \ref{fig:time_gen}. The corresponding results are reported in Table \ref{tab:ts_gen}.

\input{tables/gen_time}
As we can see, there is a negligible difference between the random and RL-based test generators for the autonomous robot. The RL-based test generator for the autonomous vehicle takes on average 72 ms more to produce a test scenario, which is also an insignificant overhead, given that the initial population size would rarely exceed 150 individuals.

\begin{tcolorbox}
\textbf{Summary of RQ1:} The RL agents for both the autonomous ant robot and autonomous vehicle LKAS system outperformed the random generator in terms of the fitness of the solutions produced. The RL agent generated test scenarios that were 52.4\% and 79.8\% fitter for the ant robot and autonomous vehicle, respectively. The random generator produced test scenarios with higher diversity, by 4.46\% and 14.3\% for the robot and vehicle, respectively. The RL test generator for the autonomous vehicle still achieved a high level of diversity (0.839), meaning that the test scenarios were at least 83.9\% different from each other. The overhead of the RL generator in terms of generation time is insignificant.
\end{tcolorbox}

\subsubsection{RQ2 - Selecting the $\rho$ hyperparameter of the RIGAA algorithm}
In RIGAA the percentage of the initial population generated by the RL agent is defined by the $\rho$ hyperparameter. In this RQ, we investigate the impact of different $\rho$ values on the final fitness and diversity of the test suite, as well as on the algorithm's convergence. To reduce the computational time, we evaluate the fitness of the solutions using the surrogate reward function defined in Eq. (\ref{eq:fit_simple}). We use the surrogate rewards for finding the optimal configuration of our algorithm, and we evaluate its performance with the simulation-based models in the RQ4. 
  For the experiments, we selected the NSGA-II evolutionary algorithm, which is commonly used for multi-objective search in search-based software engineering \cite{harman2001search}. We inserted solutions produced by the trained RL agent, as a percentage $\rho$ of the population size, where $\rho$ ranges from 0.2 to 1 in increments of 0.2, i.e. $\rho$ $\in $ [0.2, 0.4, 0.6, 0.8, 1]. In total, we evaluated 5 configurations of RIGAA. 

We performed 30 runs of the algorithm for each configuration, resulting in 30 test suites, each containing 30 test scenarios. To generate test cases for an autonomous robot, we used the genetic algorithm configuration with a population size of 150, 100 offspring per generation, and 8000 evaluations in total. The average runtime to complete 7000 evaluations was approximately 1 hour.
To generate test scenarios for an autonomous vehicle, we had the population size of 150, 150 offspring per generation, and 65000 evaluations in total. The average runtime to complete 65000 evaluations was approximately 45 minutes.

\begin{figure}[h!]
\centering
\begin{subfigure}{0.45\textwidth}
  \includegraphics[scale=0.4]{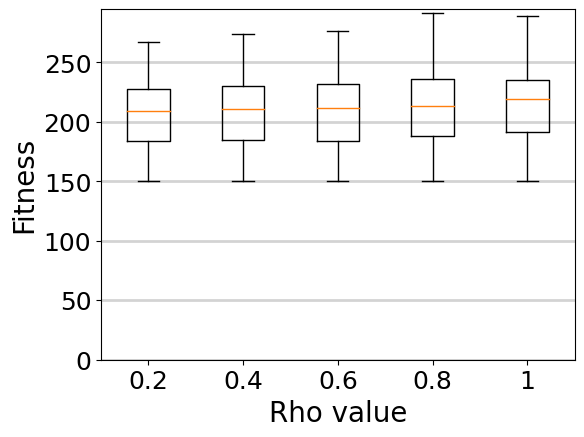} 
  \caption{Average test suite fitness for an autonomous robot for different values of $\rho$}
  \label{fig:Fitness_sweep_rob}
\end{subfigure}
\begin{subfigure}{0.45\textwidth}
  \includegraphics[scale=0.4]{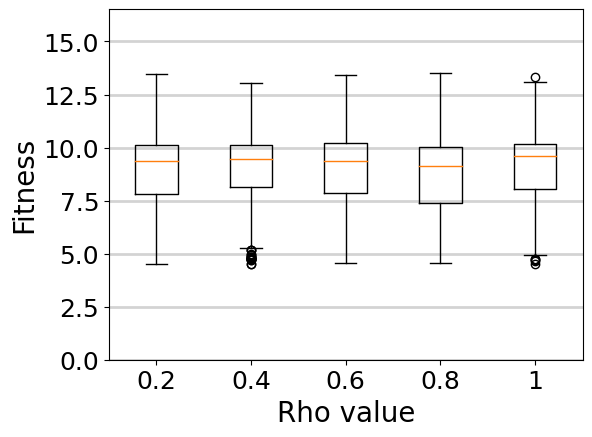} 
  \caption{Average test suite fitness for an autonomous vehicle for different values of $\rho$ }
  \label{fig:Fitness_sweep_veh}
\end{subfigure}
\caption{Average test suite fitness for different values of $\rho$}
\label{fig:fitness_sweep}
\end{figure}

\begin{figure}[h!]
\centering
\begin{subfigure}{0.45\textwidth}
  \includegraphics[scale=0.4]{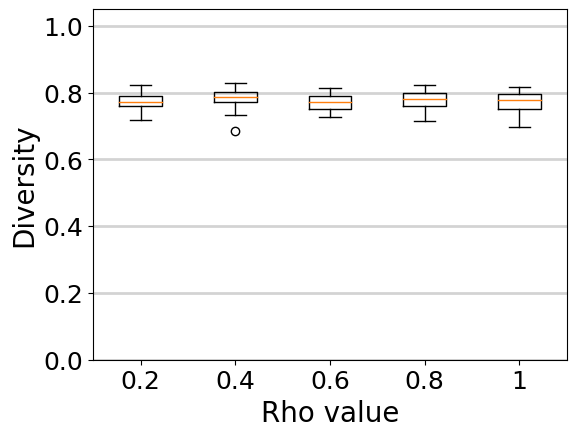}
  \caption{Average test suite diversity for an autonomous robot for different values of $\rho$ }
  \label{fig:Diversity_sweep_rob}
\end{subfigure}
\begin{subfigure}{0.45\textwidth}
  \includegraphics[scale=0.4]{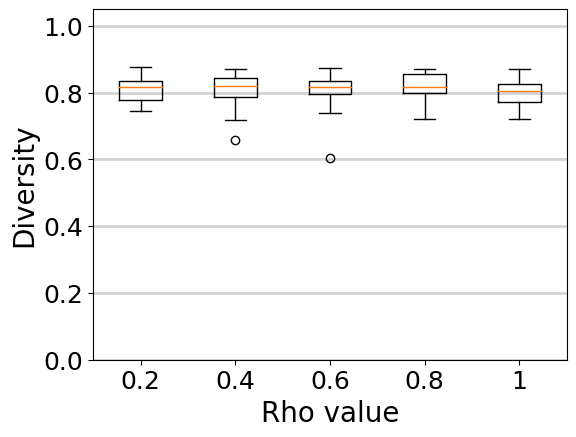} 
  \caption{Average test suite diversity for an autonomous vehicle for different values of $\rho$}
  \label{fig:Diversity_sweep_veh}
\end{subfigure}
\caption{Autonomous vehicle system scenario example}
\label{fig:diversity_sweep}
\end{figure}

\begin{figure}[h!]
\centering
\begin{subfigure}{0.45\textwidth}
  \includegraphics[scale=0.42]{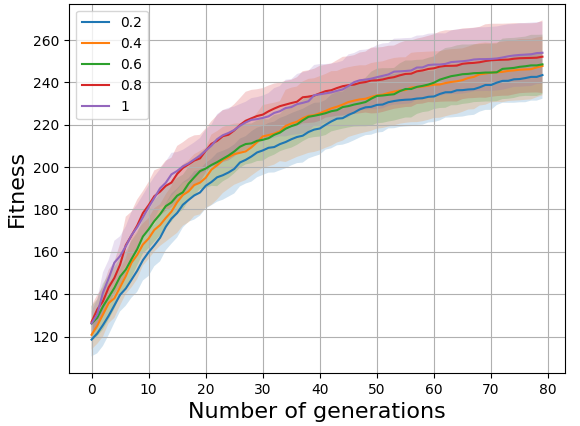}
  \caption{Convergence of RIGAA in autonomous robot problem for different values of $\rho$ }
\label{fig:convergence_sweep_rob}
\end{subfigure}
\begin{subfigure}{0.45\textwidth}
  \includegraphics[scale=0.42]{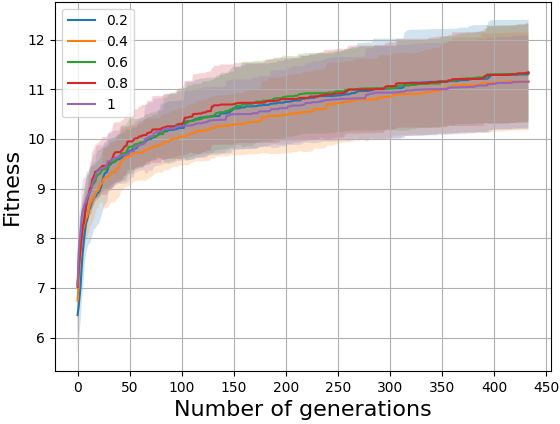}  
  \caption{Convergence of RIGAA in autonomous vehicle problem for different values of $\rho$}
\label{fig:convergence_sweep_veh}
\end{subfigure}
\caption{RIGAA convergence for different epsilon values}
\label{fig:rigaa_sweep}
\end{figure}

The box plots depicting the average fitness $F_{avs}$ (Eq. (\ref{eq:fit_simple})) of the solutions obtained for each configuration for both the autonomous robot maze and autonomous vehicle road topology generation are presented in Figure \ref{fig:fitness_sweep}, while the plots for the average diversity $D_{av}$ (Eq. (\ref{eq:diversity})) are displayed in Figure \ref{fig:diversity_sweep}. The mean values are summarized in Table \ref{tab:sweep_values}. Tables \ref{tab:ts_eff_rob} and \ref{tab:ts_eff_veh} report the results of the statistical tests and effect size measurements.
The convergence plots are shown in Figure \ref{fig:rigaa_sweep}, with Table \ref{tab:rq2_best_val_mean} showing the average best fitness found and Table \ref{tab:rq2_best_val_p} showing the corresponding results of the statistical tests.
\input{tables/rq2_mean}

\input{tables/rq2_effect_rob}

\input{tables/rq2_effect_size_veh}

The statistical analysis of the obtained results shows that there are no significant differences between the different configurations of the RIGAA algorithm regarding the fitness and diversity of the generated test scenarios. For the autonomous robot, the $\rho=1$ configuration generates statistically fitter solutions, than the $\rho$ $\in$ [0.2, 0.4, 0.6] configurations by 3.9 \%, 2.8 \% and 2.5 \% respectively. However, all these differences have a negligible effect size.
The diversity of test scenarios generated by each configuration showed negligible differences. This could be attributed to the high diversity of scenarios produced by both RL agents, which allowed the evolutionary algorithm to explore the search space without being constrained to a particular set of solutions. Moreover, in every configuration the algorithm uses the same duplicate removal threshold, preserving the diversity in the population.

\input{tables/rq2_best_value_mean}
\input{tables/rq2_best_val_p}
The convergence plots do not show any significant deviation in terms of the best solutions found. We can only observe a statistically significant difference between $\rho=0.2$ and $\rho=1$ configurations for the robot problem, where the latter finds 4.3 \% fitter solutions given the same evaluation budget.

As a result of our analysis, we have not uncovered any substantial evidence of a significant impact of $\rho$ on the algorithm's performance. We can conclude that RIGAA is robust to the choice of the $\rho$ hyperparameter. Therefore, based on the recommendations of Eiben A.E. and Smith J.E. \cite{eiben2015introduction},  we fix the value of $\rho$ at 0.4 for both the autonomous robot and autonomous vehicle RIGAA implementations. 

\begin{tcolorbox}
\textbf{Summary of RQ2:} We tested different values of the $\rho$ hyperparameter and 
concluded that RIGAA is robust to the choice of the $\rho$ hyperparameter. Therefore, based on the recommendations from the literature, we fixed the value of $\rho$ at 0.4 for both autonomous robot and autonomous vehicle RIGAA implementations to produce a more balanced initial population.
\end{tcolorbox}

\subsubsection{RQ3 – Comparing RIGAA and randomly initialized MOEA}
In this research question we compare two configurations of RIGAA, namely RIGAA based on NSGA-II (referred to as "RIGAA") and RIGAA based on SMS-EMOA (referred to as "SRIGAA"), with the corresponding algorithms NSGA-II, SMS-EMOA (referred to as `SEMOA' in the plots) having a random initialization, and random search. In the baseline MOEA (i.e. NSGA-II and SMS-EMOA), the initial population is generated randomly, whereas in RIGAA and SRIGAA, 40\% of the population is produced using a pre-trained RL agent. 

We ran each of the algorithms 30 times, resulting in 30 test suites with 30 test scenarios in each. We used the algorithm configuration selected in RQ2.
As in the previous RQ, to reduce the computational time, we evaluated the fitness of the solutions using the surrogate reward function $F_{avs}$ (Eq. \ref{eq:fit_simple}).

\begin{figure}[h!]
\begin{subfigure}{0.45\textwidth}
  \centering
  \includegraphics[scale=0.41]{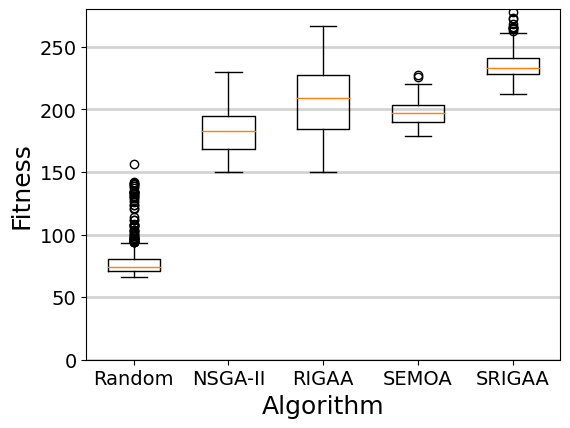}  
  \caption{Average fitness of the autonomous robot test suite}
  \label{fig:rob_fitness}
\end{subfigure}
\begin{subfigure}{0.45\textwidth}
  \centering
  \includegraphics[scale=0.41]{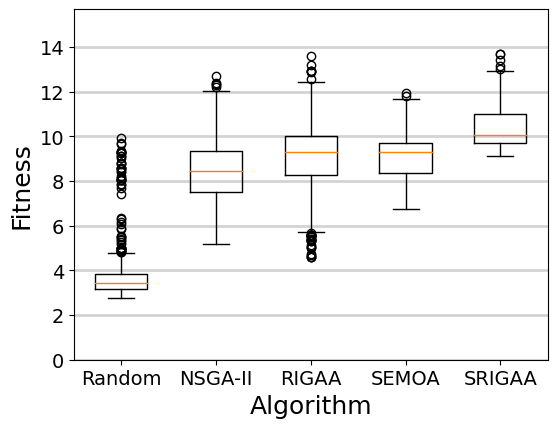}  
  \caption{Average fitness of the autonomous vehicle test suite}
  \label{fig:fitness_rq3}
\end{subfigure}
\caption{The average fitness of test suites obtained with different algorithms}
\label{fig:quality_rq3}
\end{figure}

\begin{figure}[h!]
\begin{subfigure}{0.45\textwidth}
  \centering
  \includegraphics[scale=0.41]{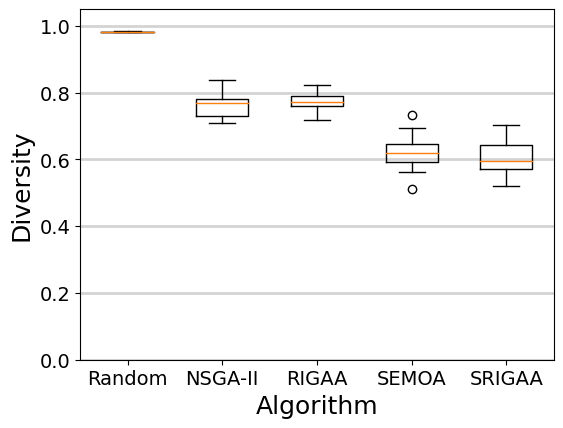}  
  \caption{Average diversity of the autonomous robot test suite}
  \label{fig:rob_div}
\end{subfigure}
\begin{subfigure}{0.45\textwidth}
  \centering
  \includegraphics[scale=0.41]{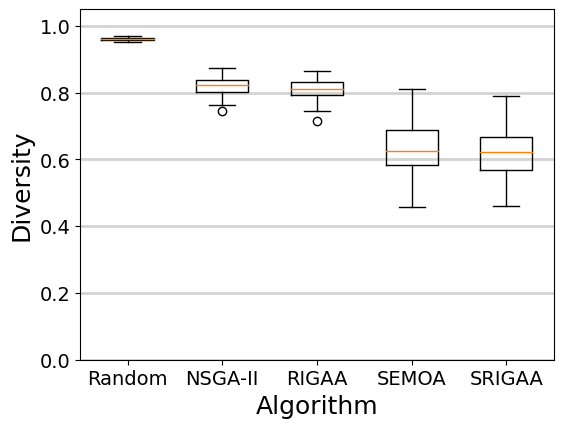}  
  \caption{Average diversity of the autonomous vehicle test suite }
  \label{fig:div_rq3}
\end{subfigure}
\caption{The average diversity of test suites obtained with different algorithms}
\label{fig:diversity_rq3}
\end{figure}

The box plots for the average fitness of the test suites for an autonomous robot and an autonomous vehicle are shown in Figure \ref{fig:quality_rq3}. The boxplots with the average diversity of the test suites for the autonomous robot and vehicle are shown in Figure \ref{fig:diversity_rq3}.

\begin{figure}[h!]
\centering
\begin{subfigure}{0.45\textwidth}
  \includegraphics[scale=0.44]{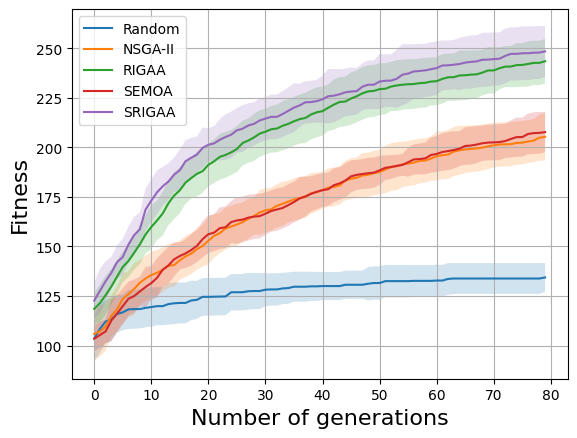}
  \caption{Convergence of algorithms in autonomous robot problem  }
\label{fig:convergence_sweep_rob}
\end{subfigure}
\begin{subfigure}{0.45\textwidth}
  \includegraphics[scale=0.44]{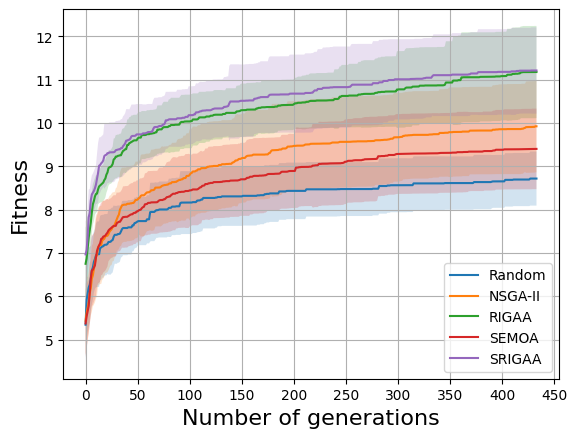}  
  \caption{Convergence of algorithms in autonomous vehicle problem}
\label{fig:convergence_sweep_veh}
\end{subfigure}
\caption{Convergence of algorithms for autonomous robot and autonomous vehicle problems}
\label{fig:rigaa_rq3_conv}
\end{figure}

The corresponding mean values are given in Table \ref{rq3_mean}. The results of statistical tests as well as the effect size evaluation for autonomous robots test suites are shown in Table \ref{rq3_rob_eff} and autonomous vehicle test suites in Table \ref{rq3_veh_eff}. The convergence plots for the algorithms are shown in Figure \ref{fig:rigaa_rq3_conv}. Table \ref{tab:rq3_best_fit_mean} and Table \ref{tab:rq3_best_values_p_val} show the mean best values found and the corresponding statistical tests results.
\input{tables/rq3_rob_mean}
\input{tables/rq3_rob_eff}
\input{tables/rq3_veh_eff}

Overall, we can see that the algorithms RIGAA and SRIGAA, where some part of the initial population is generated with an RL agent, are able to find fitter solutions with large effect size, than the corresponding algorithms NSGA-II and SMS-EMOA with random initialization. Moreover, the average diversity of the produced test suites is not statistically different. RIGAA and SRIGAA also converge to bigger fitness values faster, than NSGA-II and SMS-EMOA. 

Comparing SRIGAA and RIGAA, for an autonomous robot problem SRIGAA finds 14 \% fitter solutions on average, and for an autonomous vehicle – 13.7 \% fitter solutions. However, from the convergence plots and Tables \ref{tab:rq3_best_fit_mean}-\ref{tab:rq3_best_values_p_val}, we can see that the difference in the fitness of the best solutions found by the algorithms is not statistically significant. Moreover, the average diversity of the solutions  found by SRIGAA is 27.6 \% and 31.8 \% lower for the autonomous robot and vehicle problems respectively.
In software testing, the diversity of test scenarios is of a big importance \cite{ammann_offutt_2016}. 
We choose the RIGAA configuration based on NSGA-II for further experiments, as it produces test scenarios of a higher diversity.

\input{tables/rq3_best_value_mean}
\input{tables/rq3_best_values_p_val}
Comparing RIGAA and NSGA-II, RIGAA is able to find 13 \% fitter solutions for the autonomous robot problem and 8.9 \% fitter solutions for the autonomous vehicle problem. The best solution found by RIGAA is of a higher fitness, than the solutions found by NSGA-II with large effect size for both problems. 
Both algorithms produce solutions of high diversity of around 0.76, without significant differences in the average diversity value.

\begin{tcolorbox}
\textbf{Summary of RQ3:} SRIGAA configuration based on SMS-EMOA can find around 14 \% fitter solutions, than RIGAA configuration. However, SRIGAA test suites contain from 27.6 \% to 31.8 \% less diverse solutions, compared to RIGAA. As test diversity is critical for software testing, we choose the RIGAA configuration based on NSGA-II. RIGAA finds from 8.9 \% to 13 \% fitter solutions, than NSGA-II and converges to fitter solutions faster. There is no statistical difference between the diversity of the test suites produced by RIGAA and NSGA-II.
\end{tcolorbox}

 \subsubsection{RQ4 – Usefulness of RIGAA for simulator-based testing of autonomous robotic systems.}

In this research question, our aim is to assess the ability of RIGAA to generate diverse and challenging test scenarios for an autonomous ant robot in the Mujoco simulator and an autonomous lane keeping assist system in the BeamNG simulator. We compare RIGAA to baseline NSGA-II and random search in terms of the average fitness $F_{av}$ and diversity $D_{av}$ of the produced test suites. The average fitness metric is proportional to the degree the safety requirements are falsified by the test suite. The average diversity shows how test scenarios in the test suites are different from each other, with the value of 1 corresponding to test scenarios with no common elements.

We allocate a two-hour time budget for each algorithm to create a test suite, and we run each algorithm at least 10 times. For the autonomous robot, we utilized a genetic algorithm with a population size of 40 individuals and 20 offspring. The fitness function $F_1$ (defined in Eq. \ref{eq:mtl}) is evaluated as the percentage of the waypoints the robot completed out of all waypoints specified in the target trajectory. For better visualization, where the higher value corresponds to better fitness, we report the inverted value of the relative number of waypoints completed over the 3 runs of the robot to account for the stochasticity of its behavior. For instance, a fitness value of 5 indicates that the robot has completed only 1/5 = 0.2 or 20\% of all the waypoints in the target path. The goal of our search algorithm is to find a test scenario minimizing the number of waypoints completed. Intuitively, the fewer waypoints the robot completed, the less successful it was in achieving its mission of arriving at the specified location. Some examples of the RIGAA detected autonomous ant robot failures can be found in \href{https://youtu.be/eS25h7iLA1Q}{the following video}. 

For the autonomous vehicle, we used a genetic algorithm with a population size of 50 and 10 offspring. The fitness function $F_1$ corresponds to the percentage of the vehicle going out of road bounds. The objective of the algorithm is to minimize this fitness function taken with a negative sign. Some examples of the failures of the driving agent that were revealed with RIGAA can be found in the following \href{https://youtu.be/zaTSuFrVrhw}{demo video}.

\begin{figure}[h!]
\begin{subfigure}{0.45\textwidth}
  \centering
\includegraphics[scale=0.37]{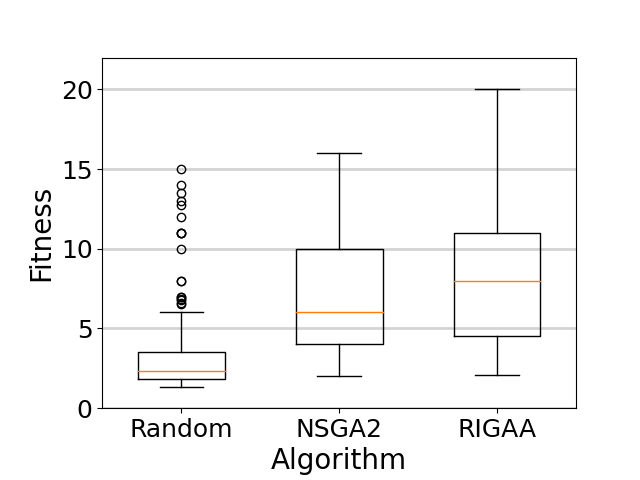} 
  \caption{The average test suite fitness for autonomous ant robot model in the simulator}
  \label{fig:rq4_rob_fit}
\end{subfigure}
\begin{subfigure}{0.45\textwidth}
  \centering
  \includegraphics[scale=0.37]{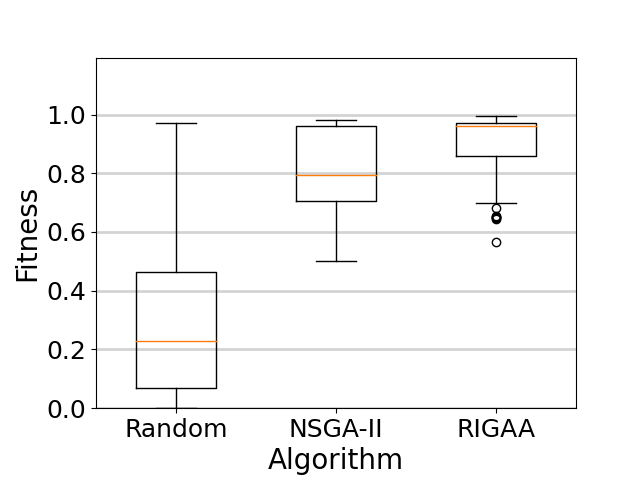}  
  \caption{The average test suite fitness for autonomous vehicle model in the simulator}
  \label{fig:rq4_veh_fit}
\end{subfigure}
\caption{The average test suite fitness produced by different algorithms for the simulator-based autonomous robotic systems}
\label{fig:rq4_sim_fitness}
\end{figure}

\begin{figure}[h!]
\begin{subfigure}{0.45\textwidth}
  \centering
\includegraphics[scale=0.37]{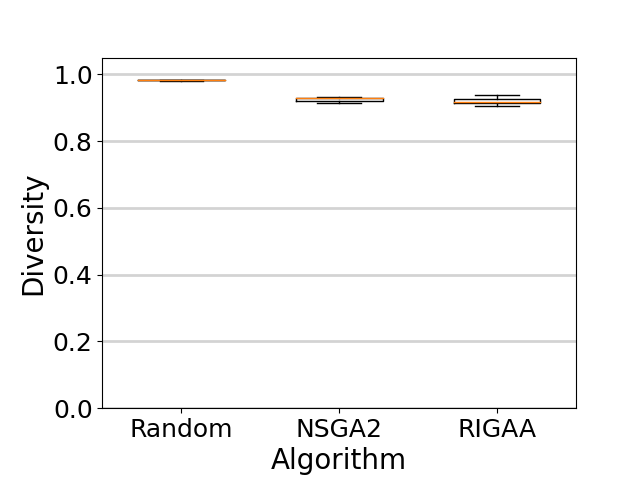} 
  \caption{The average test suite diversity for autonomous ant robot model in the simulator}
  \label{fig:rq4_rob_div}
\end{subfigure}
\begin{subfigure}{0.45\textwidth}
  \centering
  \includegraphics[scale=0.37]{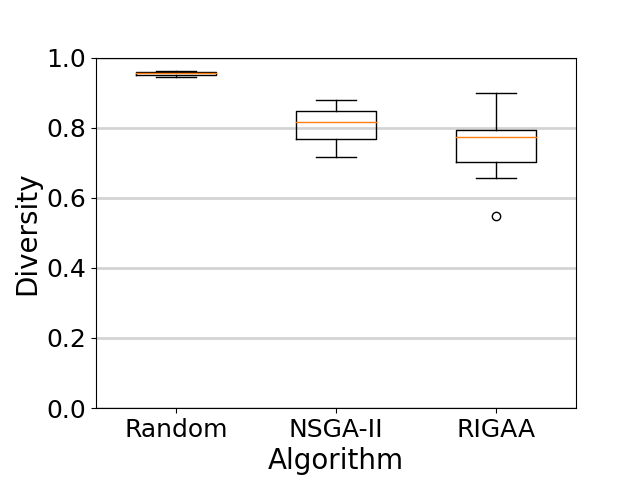}  
  \caption{The average test suite diversity for autonomous vehicle model in the simulator}
  \label{fig:rq4_veh_div}
\end{subfigure}
\caption{The average test suite diversity produced by different algorithms for the simulator-based autonomous robotic systems}
\label{fig:rq4_sim_diversity}
\end{figure}

\begin{figure}[h!]
\begin{subfigure}{0.45\textwidth}
  \centering
\includegraphics[scale=0.42]{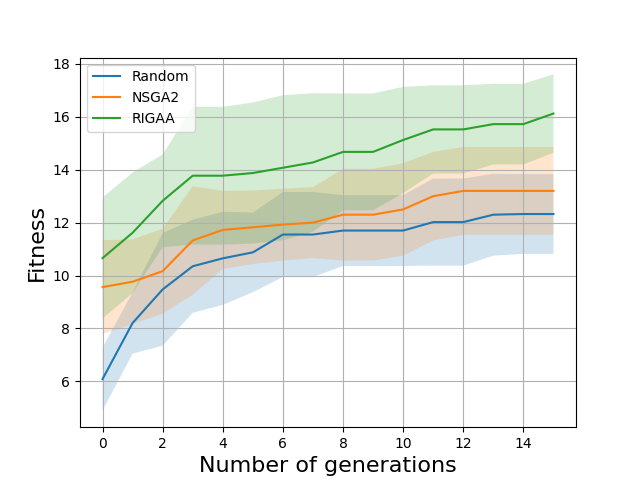} 
  \caption{Convergence of algorithms for autonomous robot testing}
  \label{fig:rq4_rob_conv}
\end{subfigure}
\begin{subfigure}{0.45\textwidth}
  \centering
  \includegraphics[scale=0.42]{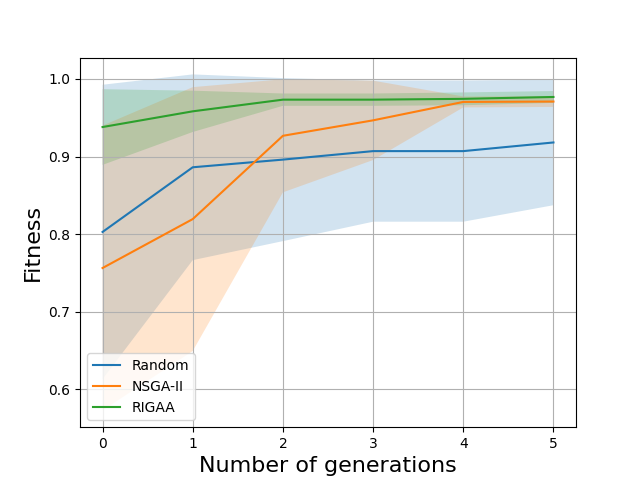}  
  \caption{Convergence of algorithms for autonomous vehicle testing}
  \label{fig:rq4_veh_conv}
\end{subfigure}
\caption{Convergence of algorithms for autonomous robot and autonomous vehicle simulator based models testing}
\label{fig:rq4_sim_conv}
\end{figure}

\input{tables/rq4_mean}
The box plots for the average fitness of the test suites for the autonomous robot and vehicle are presented in Figure \ref{fig:rq4_sim_fitness} and the average diversity box plots are shown in Figure \ref{fig:rq4_sim_diversity}. The corresponding mean values are provided in Table \ref{rq4_mean}. The results of statistical tests and the obtained effect size values for the autonomous robot and vehicle are given in Table \ref{rq4_rob_eff} and Table \ref{tab:rq4_veh_eff}, respectively. The convergence plots can be found in Figure \ref{fig:rq4_sim_conv} with the Tables \ref{tab:rq4_best_val_full_mean} - \ref{tab:rq4_best_val_full_p} showing the mean best value found and the corresponding statistical testing results respectively.

We can observe that RIGAA is able to generate test suites of higher fitness for every system under test compared to NSGA-II and random search. For the autonomous robot, RIGAA finds solutions that are on average 20\% fitter than those found by NSGA-II and 153.2\% fitter than those found by random search. There is no significant difference in diversity between NSGA-II and RIGAA test suites. Random search produces only 6.5\% more diverse test suites. From Figure \ref{fig:rq4_rob_conv} we can see that RIGAA converges to the fitter solutions faster and the best solutions are on average 22.1 \% fitter solutions than those found by NSGA-II, given the same evaluation budget.

For the autonomous vehicle, RIGAA produces test suites with solutions that are 12.26\% fitter than NSGA-II and 192.2\% fitter than random search. The difference in diversity between NSGA-II and RIGAA-generated test suites is negligible. Random search produces test suites that are 18\% more diverse than those generated by RIGAA and NSGA-II. Nonetheless, the average diversity for RIGAA is 0.749 and for NSGA-II is 0.80, indicating high levels of diversity. The best solutions found by RIGAA and NSGA-II are 6.5 \% fitter than the random search found solutions. There is no significant difference between the fitness of the best solutions found by NSGA-II and RIGAA. At the same time, from Figure  \ref{fig:rq4_sim_conv}, we can observe that RIGAA converges faster to test scenarios that cause the vehicle to go out of the road bounds.
\input{tables/rq4_rob_eff}
\input{tables/rq4_veh_eff}

\input{tables/rq4_best_val_full_mean}
\input{tables/rq4_best_val_full_p}
We also compare RIGAA with other state-of-the-art tools for testing autonomous vehicle LKAS system. Unfortunately, to the best of our effort, we have not found reproducible tools for autonomous robot testing with customizable simulation environments. We compare RIGAA with Frenetic tool \cite{castellano2021frenetic}, the winner of the SBST 2021 competition \cite{panichella2021sbst} and AmbieGen \cite{humeniuk2022search}, the winner of the SBST 2022 tool competition \cite{gambi2022sbst}. We use the configuration of AmbieGen, where the simulator-based model is used to guide the search, rather, than the simplified model. In this case, RIGAA and AmbieGen run the same algorithm, with the difference only in the initialization. AmbieGen is initialized randomly, while in RIGAA  40 \% of the initial population is generated with a trained RL agent.
\begin{figure}[h]
\begin{subfigure}{0.45\textwidth}
  \centering
\includegraphics[scale=0.4]{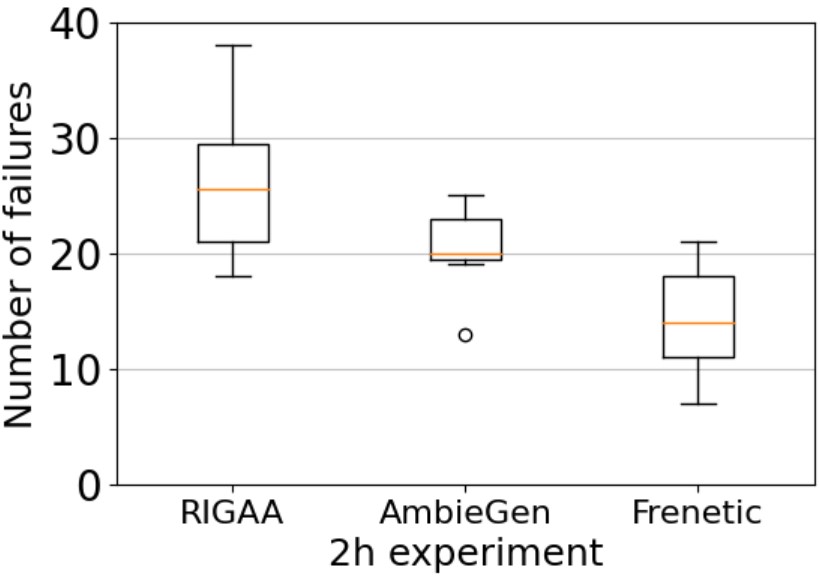} 
  \caption{Number of failures detected by the tools}
  \label{fig:rq4_rob_div}
\end{subfigure}
\begin{subfigure}{0.45\textwidth}
  \centering
  \includegraphics[scale=0.4]{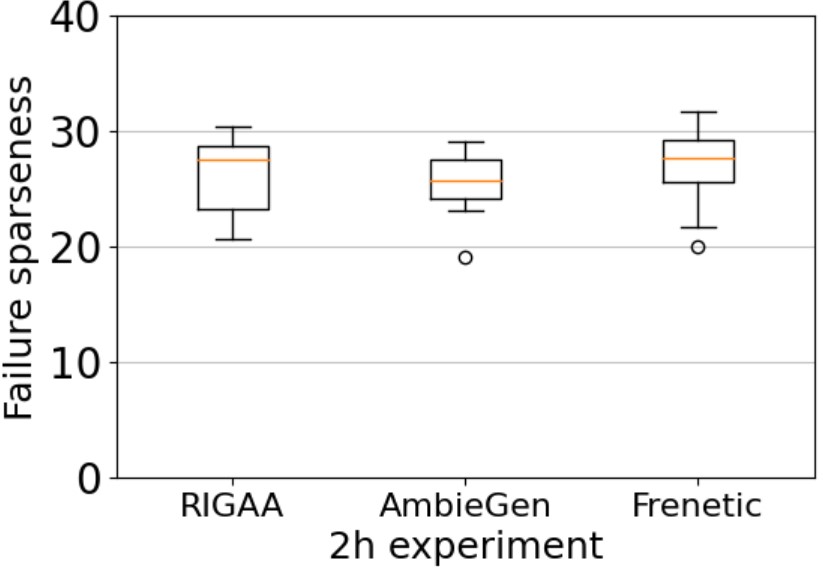}  
  \caption{The diversity of the failures detected}
  \label{fig:rq4_veh_div}
\end{subfigure}
\caption{Comparing state-of-the-art tools for autonomous vehicle lane keeping assist system testing}
\label{fig:rq4_sim_tools}
\end{figure}
\input{tables/rq5_lkas_mean}
\input{tables/rq4_lkas_p}

For the tool comparison, we followed the guidelines of the competition by allocating a fixed time budget of 2 hours to all the tools. We then calculated the total number of failures detected as well as the sparseness of the detected failures, corresponding to their diversity, as defined by Riccio et al. \cite{riccio2020model}. A failure was considered to have occurred when the vehicle went out of the lane bounds by more than 85\% of its area. We run each tool at least 10 times.

The box plots showing the number of failures revealed, and their sparseness, are presented in Figure \ref{fig:rq4_sim_tools}. Mean values for these metrics can be found in Table \ref{rq4_mean_lkas}. Results of statistical tests and corresponding effect sizes are summarized in Table \ref{rq4_lkas_p}.
We can see that RIGAA outperforms the other tools with a large effect size in terms of the number of failures revealed. RIGAA discovers 24.6\% more failures than AmbieGen and 83.2\% more failures than Frenetic within a two-hour execution budget. All three tools generate failures with high diversity, and the difference in diversity i.e., sparseness among the tools is not significant.

\begin{tcolorbox}
\textbf{Summary of RQ4:} RIGAA outperforms NSGA-II by generating test suites with 20\% higher average fitness for an autonomous ant in the Mujoco simulator, and with 12.26\% higher average fitness for an autonomous vehicle in the BeamNG simulator. RIGAA is also able to converge to fitter solutions faster. The difference in diversity between RIGAA and NSGA-II produced solutions is negligible. RIGAA also outperforms the state-of-the-art tools in terms of the number of failures revealed for an autonomous vehicle lane keeping assist system. It discovers 23.6 \% failures more, than AmbieGen tool and 83.2 \%, than the Frenetic tool given a two-hour evaluation budget.
\end{tcolorbox}

\subsection{Threats to validity}

\textbf{Internal validity.} 
To minimize the threats to internal validity, relating to experimental errors and biases, whenever available, we used standardized frameworks for development and evaluation. We implemented the evolutionary search algorithms based on a popular Python-based Pymoo framework \cite{pymoo}. We used the PPO RL algorithm implementation provided by the Stable-baselines framework \cite{stable-baselines3}. For the autonomous robot problem, we used an open-source and standard RL Ant-maze environment based on Gymnasium library \cite{schulman2015high}, \cite{ant-gym-env}. To evaluate the scenarios for the LKAS case study, we utilized a standardized test pipeline from the SBST 2022 workshop tool competition \cite{gambi2022sbst}. 

\textbf{Conclusion validity.}
 Conclusion validity is related to random variations and inappropriate
use of statistics. To mitigate it, we followed the guidelines in \cite{arcuri2014hitchhiker} for search-based algorithm evaluation. We repeat all evaluations a statistically significant number of times, i.e. at least 10 when using the simulator and at least 30 otherwise. For all results, we report the non-parametric Mann-Whitney U test with a significance level of $\alpha = 0.05$, as well as the effect size measure in terms of Cliff's delta. 

\textbf{Construct validity.} Construct validity is related to the degree to which an evaluation measures what it claims. In our experiments, we used two evaluation metrics: the test scenario fitness as well as test scenario diversity. We evaluated the fitness proportionally to the safety requirement violations. For the autonomous ant robot, we calculated the percentage of the target trajectory completed and for the autonomous vehicle, the maximum deviation from the road lane. For the evaluation of diversity, we utilized the Jaccard distance, which compares test scenarios' similarity based on their input representation. This metric provides normalized diversity values, which is useful for comparison on different testing problems. However, it only evaluates the differences in the inputs given to the system, and not the differences in the output behavior of the system. In the future, we plan to also include metrics for evaluating the diversity in the output behavior coverage by the test scenarios.

\textbf{External validity.} External validity relates to the generalizability of our results. We demonstrated how our framework can be applied to generate environments for two different autonomous robotic systems. However, we only considered a limited number of test subjects and limited levels of environment complexity. Therefore, additional experiments should be conducted with different robotic systems and environments with higher complexity for better generalizability evaluation of RIGAA. 

We have also experimented with two different multi-objective evolutionary algorithms (MOEA) to be combined with a pre-trained RL agent. Both of them have shown positive results, however, we need to conduct experiments with a bigger variety of search techniques, such as evolution strategies and particle swarm optimization \cite{harman2008search}, to confirm that RL-based initialization is beneficial for different search algorithms.
\subsection{Data availability} \label{sec:rep_pack}
The replication package of our experiments, including the implementation of the search algorithms and the instructions for installation of the simulators we used, is available at Zenodo repository \cite{dmytro_humeniuk_2023_8242223}.
\section{Discussion}\label{sec:discussion}
A non-random initial population can direct the search into particular regions of the search space that contain good solutions \cite{eiben2015introduction}. This is of special interest to search-based testing of autonomous robotic systems, where the cost of solution evaluation is high. By reducing the time spent on evaluating non-promising solutions, more failures can be potentially revealed in less amount of time. Our search-based approach, RIGAA,  leverages a pre-trained RL agent to insert some proportion of high-quality test scenarios in the initial population. Preliminary results have shown that RL informed initialization allows the search algorithm to find better test scenarios within the same computational budget, in contrast to a configuration based on random initialization.
In this section we discuss in a more detail the advantages and disadvantages of RL-based initialization as well as the promising directions for expanding our work.
\subsection{Discovering more challenging test scenarios with RL-based initialization}


In RIGAA, instead of manually designing domain knowledge-based initialization heuristics or reusing pre-existing scenarios, we rely on the RL agent to automatically infer the domain knowledge rules via continuous interactions with the provided environment. The domain knowledge can be obtained by both interactions with the actual simulator environment or the environment with surrogate rewards. We opted for the latter in order to reduce the computational cost of training the agent. Our experiments show that as long as the RL agent produces test scenarios with higher fitness, than the random generator, it can be useful for initialization of the evolutionary search.
We surmise that incorporating individuals with initial problem knowledge into the evolutionary search process can lead to the discovery of new types of challenging test scenarios that a purely stochastic search process may fail to discover within a reasonable computational budget. 

\begin{figure}[h]
\centering
\begin{subfigure}{0.45\textwidth}
\includegraphics[scale=0.15]{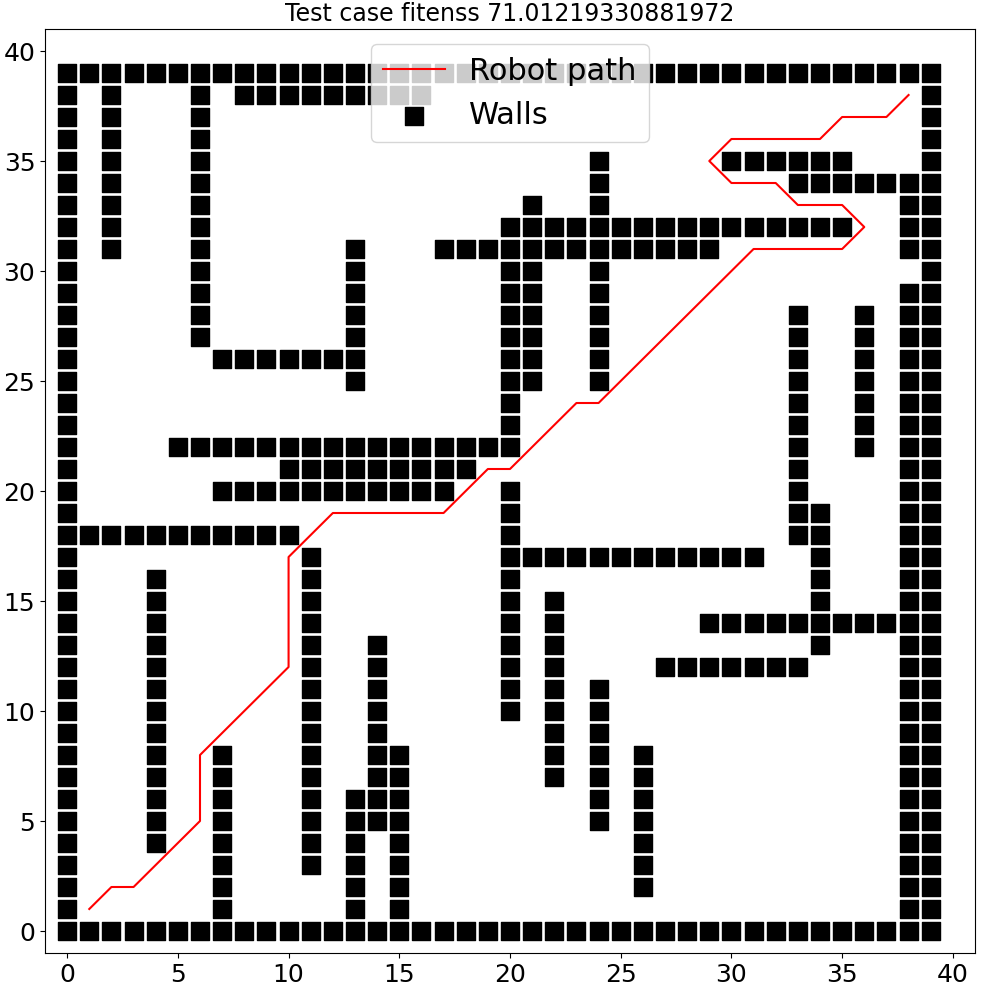} 
  \caption{Randomly produced scenario}
  \label{fig:rob_ran}
\end{subfigure}
\begin{subfigure}{0.45\textwidth}
  \includegraphics[scale=0.15]{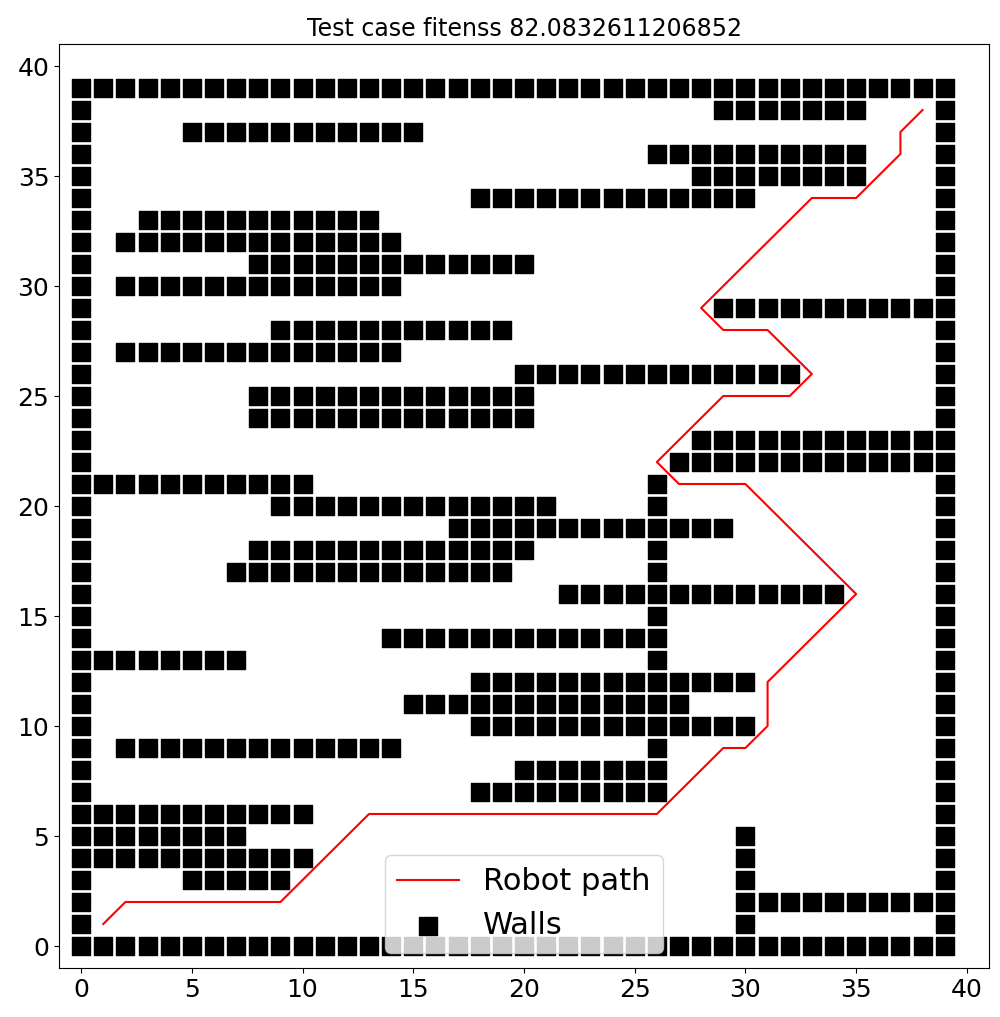}  
  \caption{RL produced scenario}
  \label{fig:rob_rl}
\end{subfigure}
\begin{subfigure}{0.45\textwidth}
  \includegraphics[scale=0.15]{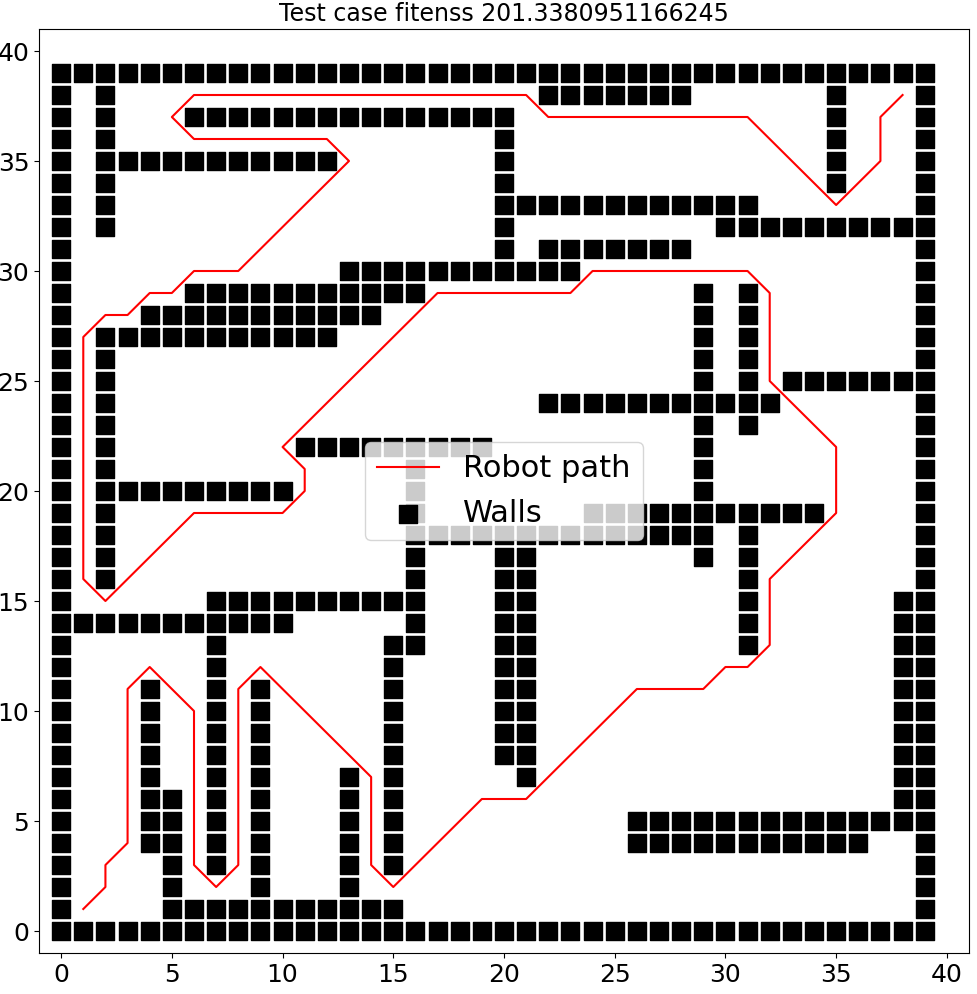}  
  \caption{NSGA-II produced scenario}
  \label{fig:rob_nsga2}
\end{subfigure}
\begin{subfigure}{0.45\textwidth}
  \includegraphics[scale=0.15]{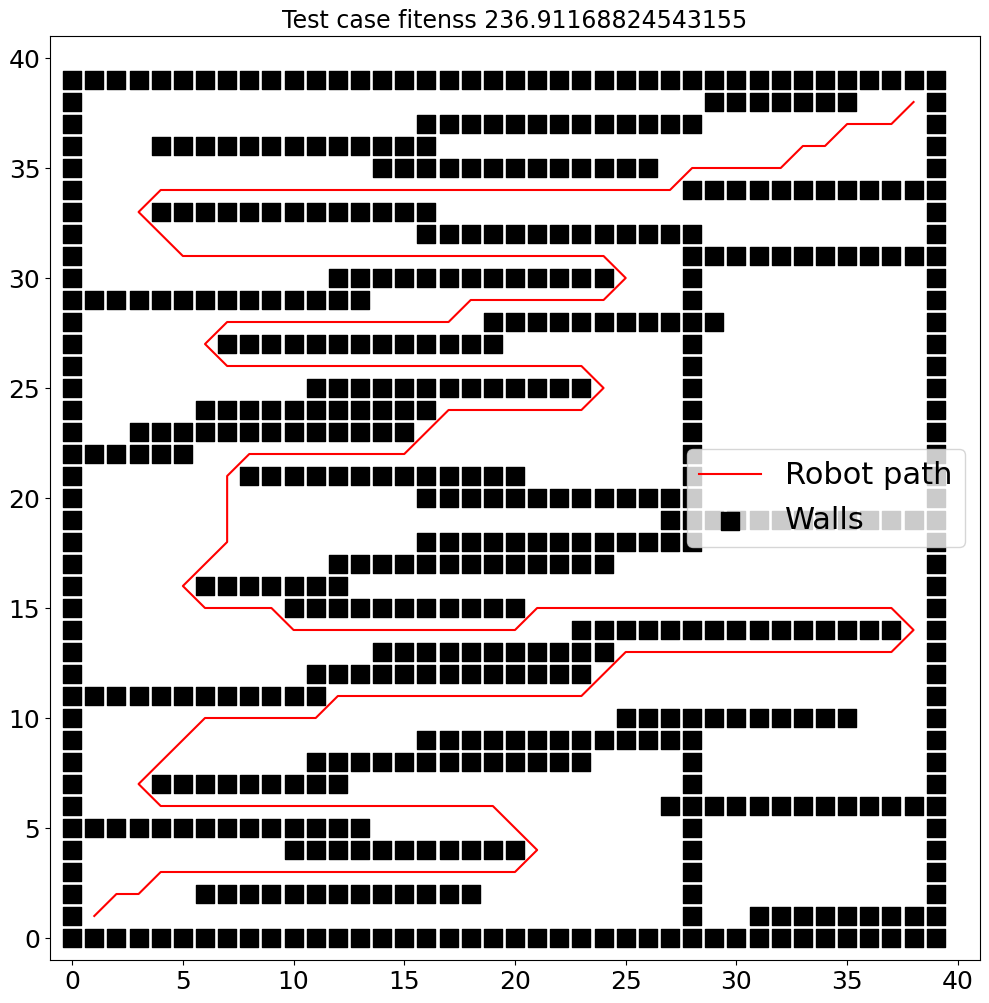}  
  \caption{RIGAA produced scenario }
  \label{fig:rob_rigaa}
\end{subfigure}
\caption{Comparing test scenarios for an autonomous robot}
\label{fig:rob_scenarios}
\end{figure}

\begin{figure}[h!]
\centering
\begin{subfigure}{0.45\textwidth}
\includegraphics[scale=0.2]{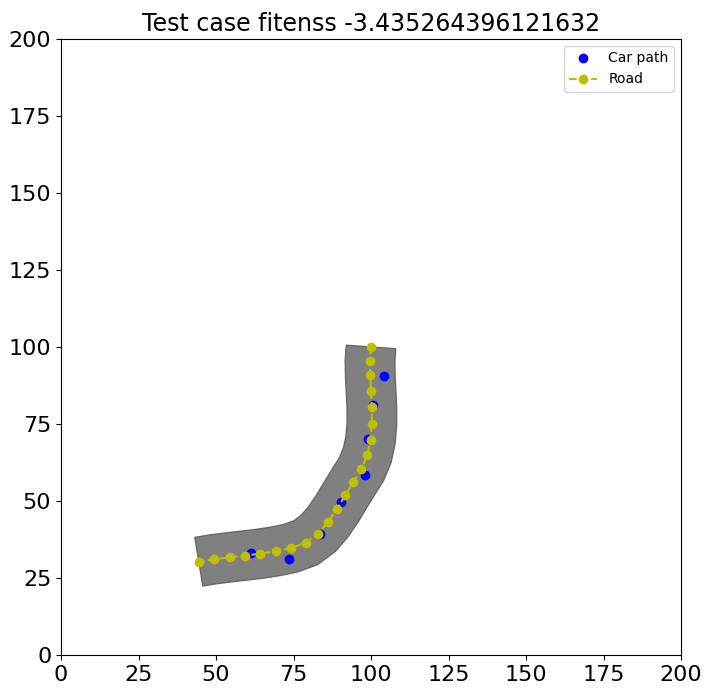} 
  \caption{Randomly produced scenario}
  \label{fig:veh_ran_ex}
\end{subfigure}
\begin{subfigure}{0.45\textwidth}
  \includegraphics[scale=0.2]{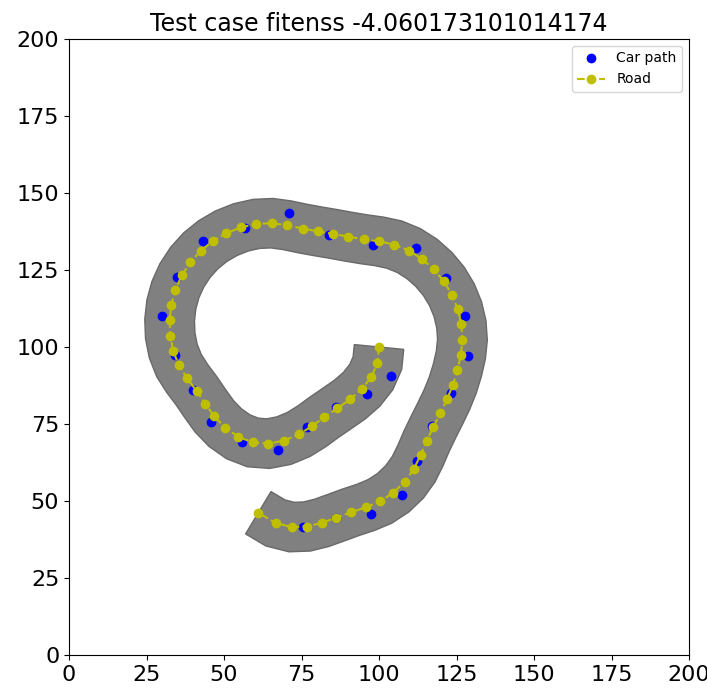}  
  \caption{RL produced scenario}
  \label{fig:veh_rl_ex}
\end{subfigure}
\begin{subfigure}{0.45\textwidth}
  \includegraphics[scale=0.2]{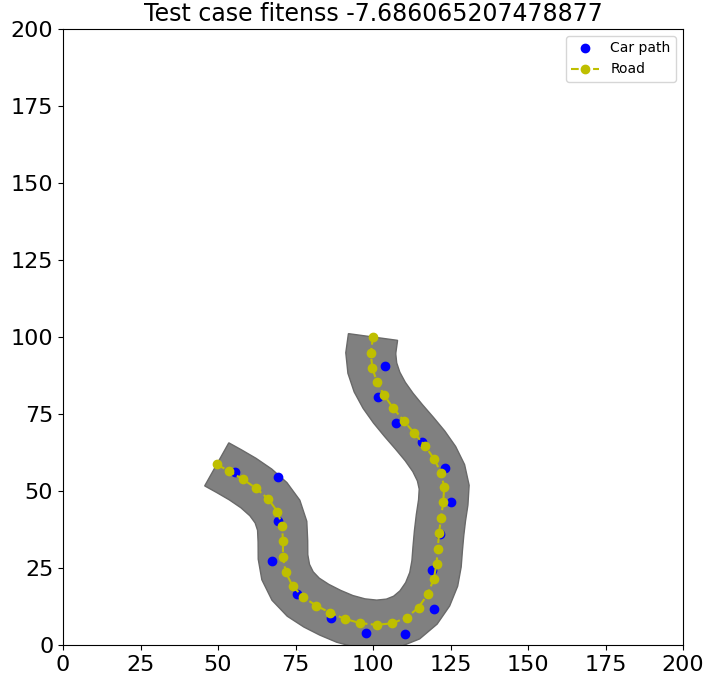}  
  \caption{NSGA-II produced scenario}
  \label{fig:veh_nsga2_ex}
\end{subfigure}
\begin{subfigure}{0.45\textwidth}
  \includegraphics[scale=0.2]{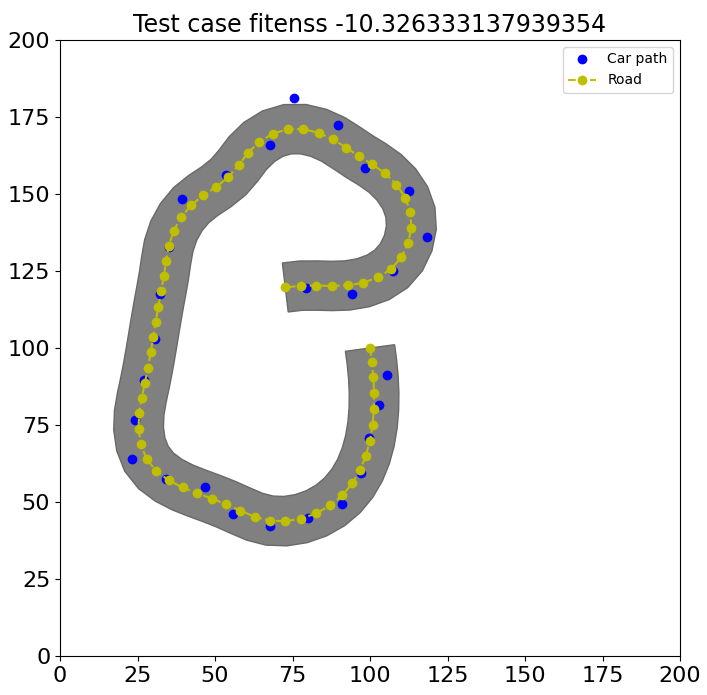}  
  \caption{RIGAA produced scenario }
  \label{fig:veh_rl_ex}
\end{subfigure}
\caption{Comparing test scenarios for an autonomous vehicle}
\label{fig:veh_scenarios_ex}
\end{figure}


As an example, in Figure \ref{fig:rob_scenarios} we show examples of the generated mazes for an autonomous robot produced randomly, with an RL agent, with NSGA-II and with RIGAA.
 In Figure \ref{fig:rob_rl} and Figure \ref{fig:rob_rigaa} we can observe a maze pattern with mostly horizontal walls, produced with an RL agent and RIGAA respectively. Firstly, this pattern was introduced into the initial population by a trained RL agent, then it was evolved with an evolutionary search, producing a maze with the highest fitness observed in our experiments. It is different from a typical maze pattern that was discovered by random search and NSGA-II, shown in Figure \ref{fig:rob_ran} and Figure \ref{fig:rob_nsga2} that has arbitrary choices of vertical and horizontal walls. 

In Figure \ref{fig:veh_scenarios_ex} we show the corresponding examples of road topologies produced for testing autonomous vehicles. 
In Figure \ref{fig:veh_rl_ex} we can see a curved road topology pattern, produced by an RL agent. After being evolved by evolutionary search, more challenging road topology was produced (Figure \ref{fig:veh_rl_ex}), forcing the vehicle to go further from the road lane center, than in NSGA-II discovered test scenarios.


The examples presented above illustrate that RIGAA can be useful for discovering new types and patterns of test scenarios, potentially revealing previously unseen system failures.
\subsection{Challenges of using RL agents for test scenario generation}
In RIGAA we use a pre-trained RL agent to generate some percentage of the initial population. Below, we outline some of the challenges related to training an RL agent for test scenario generation.

\textbf{Sample inefficiency}. RL agents typically require a large number of interactions with the environment to learn effective policies. The process of training the RL agent may be computationally expensive and time-consuming, especially if the RL agent needs to explore a complex and high-dimensional action space. For an autonomous robot, the action space to explore was multidimensional and of the size $2x7x37$. The average time to compute the reward value was 0.1 s. We trained the agent for a total of 500 000 steps, taking around 14 hours on average. For an autonomous vehicle, the action space was of the size $3x25x35$, with the average time to compute the reward of 0.01 s. We trained the agent for a total of 2 500 000 steps, taking around 7 hours on average. When the problem action space is high-dimensional, we recommend the use of an inexpensive reward function, as the computation of the rewards based on the system behavior in the simulator becomes prohibitively expensive. In our case studies, test scenario evaluation in Mujoco simulator (autonomous robot) could take from 5 to 10 s, and in the BeamNg simulator (autonomous vehicle) from 10 to 40 s. 

\textbf{Reward shaping and hyperparameter fine-tuning}. RL algorithms typically have several hyperparameters as well as a reward function that require careful tuning to achieve good performance. In the context of testing, we are interested in producing not only challenging but also diverse test scenarios. One of the important steps is to choose the reward function encouraging challenging scenarios as well as the hyperparameters that provide a good exploration-exploitation trade-off. Otherwise, the trained RL agent might learn to generate only a specific type of test scenario, lacking diversity. In our experiments, we gave additional rewards to the agent for selecting novel elements for the test scenarios. Additionally, we fine-tuned the entropy coefficient hyperparameter, that balances exploration and exploitation of the agent. 

\textbf{Limitations in application domains}. To be able to use RL, the test scenario problem should be first formulated as an MDP. If it is not possible in a given problem, RIGAA may not be applicable. We hypothesize that the majority of test scenario generation problems for autonomous robotic systems can be formulated as an MDP.

 \textbf{Surrogate reward design}. in some test generation problems, it can be challenging to formulate the domain knowledge rules and apply them to design the surrogate rewards, that serve as a proxy for the rewards obtained from executing the simulator-based model of the system. One solution is to directly utilize the feedback from the simulator-based model as a reward for training the RL agent. In this case, the action space should be minimized, to be easier for the agent to explore. In addition, for better efficiency it can be beneficial to only use feedback from the simulator in the form of sparse rewards i.e., positive reward that is only given when the task is achieved e.g., the vehicle goes out of the lane. 

Overall, training an RL agent requires additional effort for the creation of the training environment, as well as the computational resources to train the agent. However, once the agent is trained to produce challenging and diverse scenarios, it can be re-used as a generator of test scenarios, with minimal costs of generation. In our experiments, the efficiency of the RL generator was comparable with that of the random generator. Moreover, it can help uncover new types of failures, not previously found with randomly initialized algorithms. In this case, the performance on system falsification may take
priority over the algorithm cost and complexity. 

\subsection{Future work and promising applications of RL for SBST} 

Evolutionary algorithms are of particular appeal to test scenario generation, due to their innate ability to explore vast search spaces. Given the high-evaluation cost nature of autonomous robotic system testing problems, we surmise that maximizing the effectiveness of these algorithms within a limited time budget is of high importance.
A number of research works have been done in this direction, such as using a diversity metric \cite{gambi2019automatically} or machine learning \cite{abdessalem2018testing} to prioritize test scenario evaluation, based on previous failure-inducing test scenarios \cite{moghadam2022machine}. Scenarios extracted from the real world \cite{gambi2022generating} were used to improve the initialization. RIGAA approach, where the RL agent is used for EA initialization, is orthogonal to these approaches and can be used separately or in combination with them. We surmise one of the strong sides of using RL for test scenario design is its utilization of gradient-based methods to optimize the objective function. In our experience, it effectively complements gradient-free methods like evolutionary search. The use of RL independently of EA for scenario generation is possible, however, in our experiments, combining RL with EA always yielded better results.
Below, we outline some promising research directions for using reinforcement learning in search-based testing of autonomous robotic systems.

\textbf{Using surrogate rewards to train the RL agent.} 
Current state-of-the-art RL algorithms, such as PPO \cite{schulman2017proximal}, DQN \cite{mnih2013playing}, A2C \cite{mnih2016asynchronous} are model-free reinforcement learning algorithms, where the agent learns its policy directly from interaction with the environment without explicitly building a model of the environment dynamics \cite{sutton2018reinforcement}. Despite being applicable to a wide range of problems, one limitation of these algorithms is their sample inefficiency, meaning they require a large number of interactions with the environment to converge to an optimal policy. As evaluating the rewards with the simulator is expensive, we surmise it is important to build the surrogate models of the system, with such techniques as Bayesian Optimization \cite{snoek2012practical} or Genetic Programming \cite{koza1994genetic}, that can be used as accurate, but less computationally expensive approximations of the system behavior, defining the reward. The feedback from the real simulator can be used only in the form of sparse, i.e., infrequent rewards, to reduce the RL agent training computational cost.

\textbf{Exploratory RL algorithms for test scenario design}.
In our work, we have mainly relied on the PPO algorithm for the agent training, as it has demonstrated a state-of-art performance on a number of tasks \cite{schulman2017proximal}. However, we surmise that RL algorithms that encourage exploration are very promising for test scenario design. For example, in A3C algorithm \cite{mnih2016asynchronous}, instead of having one agent, multiple agents are trained at the same time, having their own set of parameters. This approach leverages diverse learning experiences from the individual agents to enhance overall training and exploration. Another example is the random network distillation approach \cite{burda2018exploration}, which encourages exploration by providing additional rewards for discovering novel observations.

\textbf{Using RL in the other parts of GA pipeline.}
Apart from initialization, reinforcement learning can be applied to the other stages of the evolutionary algorithm, i.e. in order to make better choices of mutation or crossover operators \cite{pettinger2002controlling}, mutation or crossover rate \cite{quevedo2021using} as well as the parents for mating \cite{sakurai2010method}. We surmise the use of RL is promising for improving the efficiency and effectiveness of search-based algorithms for autonomous robotic systems testing, where evaluations are resource-intensive.


\textbf{Using RL to design dynamic scenarios.}
Recently, RL has been applied to the design of dynamic test scenarios \cite{chen2021adversarial}, \cite{haq2022many}, \cite{lee2020adaptive}
, i.e., test scenarios that evolve based on the behavior of the systems under test. The scenario parameters that are typically static, such as the weather, luminosity level, and moving objects' speed (e.g., vehicles or pedestrians), can be changed by a trained RL agent during the scenario execution to produce even more challenging scenarios, covering a higher percentage of safety objectives \cite{haq2022manyobjective}. 


\section{Conclusions}\label{sec:conclusions}


In our study, we proposed RIGAA, a search-based approach that improves the effectiveness of search-based closed-loop testing of autonomous robotic systems. By utilizing a pre-trained RL agent, RIGAA incorporates prior knowledge into the test generation process, enabling effective exploration of safety requirement violating scenarios. At the same time, RIGAA reduces the evaluation time spent on irrelevant tests. We evaluated RIGAA on two systems: an RL-controlled Ant robot and a vehicle lane keeping assist system. Our results demonstrate that RIGAA outperforms the search algorithms with random initialization. Notably, RIGAA outperformed NSGA-II by producing 20\% more effective test suites  for the autonomous ant robot in the Mujoco simulator and 12.26\% more effective test suites for the autonomous vehicle in the BeamNG simulator. RIGAA also outperformed some of the state-of-the-art tools for LKAS testing.  It
discovered 23.6 \% more failures, than AmbieGen tool and 83.2 \% more, than the Frenetic tool given
a two-hour evaluation budget.

As part of future work, we plan to explore the possibility of using surrogate models of the system to train the RL agent, as well as apply the RL agents to different parts of the evolutionary search. Moreover, we surmise RL algorithms can be useful for designing dynamic test scenarios, where the parameters change depending on the actions of the other scenario participants. RIGAA could be used to provide a good initialization for such dynamic scenarios.
\begin{acks}
This work is funded by the Natural Sciences and Engineering Research Council of Canada (NSERC) [Grant No: RGPIN-2019-06956] and the Canadian Institute for Advanced Research (CIFAR). 
\end{acks}

\bibliographystyle{ACM-Reference-Format}
\bibliography{sample-base}

\end{document}

%% file: tables/repr.tex
\begin{table}[h!]
\caption{Test case representation}
\label{tab:repr}
\centering
\begin{tabular}{ccccc}
\hline
&    $E_1$    & $E_2$     & $...$ & $E_m$ \\
\hline
$A_1$ & $A_{1e1}$ &$A_{1e2}$ & $...$ &$A_{1em}$\\
$A_2$ & $A_{2e1}$ &$A_{2e2}$ & $...$ & $A_{2em}$ \\
$...$ & $...$     &$...$     & $...$ & $...$ \\
$A_n$ & $A_{ne1}$ &$A_{ne2}$ & $...$ & $A_{nem}$ \\    
\hline
\end{tabular}
\end{table}

%% file: tables/rob_scenario.tex
\begin{table}[H]
\caption{Example of test scenario representation for an autonomous robot}
\label{tab:rob_scenario}
\centering
\begin{tabular}{cccccc}
\hline
                      & $E_0$      & $E_1$    & $E_2$       & $E_3$    &  $E_4$ \\
\hline
$A_0,wall\;type$     & horizontal & horizontal & horizontal & vertical  &  horizontal \\
$A_1,wall\;position$ & 7         & 3          & 6.5 & 4.5 &   2.5 \\
$A_2,wall\;size$    & 2          &3           & 2 & 3 &  3 \\    
\hline
\end{tabular}
\end{table}

%% file: tables/veh_scenario.tex
\begin{table}[h!]
\caption{Example of test scenario representation for an autonomous vehicle}
\label{tab:veh_scenario}
\centering
\begin{tabular}{cccccc}
\hline
                      & $E_0$      & $E_1$    & $E_2$       & $E_3$    &  $E_4$ \\
\hline
$A_0,road\;type$     & straight & right & right & right &  straight \\
$A_1,length$         & 15        & -    & -    & - &   5 \\
$A_2,turning\;angle$ & -        &60     & 60      & 75  &  - \\    
\hline
\end{tabular}
\end{table}

%% file: tables/params.tex
\begin{table}[h!]
\caption{Parameters for the simplified vehicle kinematic model}
\label{tab:params}
\centering
\begin{tabular}{ccccc}
\hline
$\delta_0$ & $\theta_0$ & $\nu_0$ & $\alpha_0$ &  $\Delta t$ \\
 0         &   0       & 15         & 0.1          &   0.7 \\
\hline
\end{tabular}
\end{table}

%% file: tables/param_rigaa.tex
\begin{table}[h]
\caption{RIGAA algrithm hyperparameters}
\label{tab:hyper_param}
\centering
\begin{tabular}{cccccc}
\hline
Parameter     & Description & Value    \\
\hline
$N_{pop}$     & Population size & 150 \\
$CROS$        & Crossover rate      & 0.9    \\
$MUT$         & Mutation rate       & 0.4     \\   
$N_{TC}$      & Test suite size     &   30       \\
$\rho$ & Percentage of RL agent generated individuals & 0.4\\
\hline
\end{tabular}
\end{table}

%% file: algorithms/rigaa.tex
\begin{algorithm}
\caption{RIGAA: Reinforcement learning Informed Genetic
Algorithm for Autonomous systems testing}\label{alg:rigaa}
\begin{algorithmic}[1]
\State Set hyperparameters $N_{pop}$, $CROS$, $MUT$, $N_{TC}$, $\rho$, 
\State Load a pre-trained RL model
\State \textit{Initialize population $P$:}
\While {$N \neq N_{pop}$}
 \State Generate an individual with the pre-trained RL model with probability $\rho$
\State Generate an individual randomly with probability $1 - \rho$
\State $N = N+1$
\EndWhile

\State \textit{Run the genetic algorithm:}
\State Evaluate objectives $F_1$, $F_2$ over population $P$
\While{not ($T$) }
\quad{\# $T$ - a termination criterion}

\State Select individuals for mating from $P \rightarrow P'$ based on tournament selection
\State Apply crossover with probability of $CROS$ $P' \rightarrow P''$
\State Apply mutation with probability  $MUT$ $P''\rightarrow P'''$ 
\State Evaluate fitness objectives of the offsprings $P'''$
\State Construct new population$(P \cup P''') \rightarrow P^*$ from parents and offsprings
\State Select the fittest individuals based on crowding distance from $P^*$ of the size of $N_{pop}$, $P^* \rightarrow P$ 

\EndWhile\label{euclidendwhile}

\State Add $N_{TC}$ best individuals from the final generation to the test suite.

\end{algorithmic}
\end{algorithm}

%% file: tables/rq1_mean.tex
\begin{table}[h!]
\caption{Results for comparing the test generators in terms of a surrogate reward function and diversity}
\label{tab:ts_gen}
\centering
    \begin{tabular}{cccccc}
    \hline
        Problem & Metric & Random & RL agent & p-value & Effect size \\ \hline
        robot & $F_{avs}$ & 57.047 & 86.961 & <0.001 & 0.967 L \\ 
        robot & $D_{av}$ & 0.982 & 0.94 & <0.001 & 1.0 L \\ 
        vehicle & $F_{avs}$ & 2.531 & 4.553 & <0.001 & 0.927 L \\ 
        vehicle & $D_{av}$ & 0.959 & 0.839 & <0.001 & 1.0 L \\ \hline
    \end{tabular}
\end{table}

%% file: tables/rq1_mean2.tex
\begin{table}[h!]
\caption{Results for comparing the test generators in terms of a simulator-based fitness function}
\label{tab:rq1_mean_fitness}
\centering
    \begin{tabular}{cccccc}
    \hline
        Problem & Metric & Random & RL agent & p-value & Effect size \\ \hline
        robot & $F_{av}$ & 2.729 &3.671 & <0.001 & 0.243 S \\ 
        vehicle & $F_{av}$ & 0.35 & 0.66 & <0.001 & 0.747 L \\ \hline
    \end{tabular}
\end{table}

%% file: tables/gen_time.tex
\begin{table}[h!]
\caption{Results for comparing test case generation time}
\label{tab:ts_gen}
\centering
    \begin{tabular}{cccccc}
    \hline
        Problem & Metric & Random & RL agent & p-value & Effect size \\ \hline
        robot & mean generation time, s & 1.627 & 1.421 & 0.24 & 0.032, N \\ 
        vehicle & mean generation time, s & 0.034 & 0.106 & <0.001 & 0.886, L \\ \hline
    \end{tabular}
\end{table}

%% file: tables/rq2_mean.tex
\begin{table}[h!]
\caption{Mean values of test suite fitness and diversity for different values of $\rho$}
\label{tab:sweep_values}
\centering
    \begin{tabular}{ccccccc}
    \hline
        Problem & Metric & 0.2 & 0.4 & 0.6 & 0.8 & 1 \\ \hline
        robot & $F_{avs}$ & 205.712 & 207.856 & 208.304 & 211.414 & 213.68 \\ 
        robot & $D_{av}$ & 0.775 & 0.782 & 0.773 & 0.78 & 0.771 \\ 
        vehicle & $F_{avs}$ & 8.976 & 9.075 & 9.045 & 8.842 & 9.119 \\ 
        vehicle & $D_{av}$ & 0.812 & 0.813 & 0.809 & 0.816 & 0.801 \\ \hline
    \end{tabular}
\end{table}

%% file: tables/rq2_effect_rob.tex
\begin{table}[h!]
\caption{Non-parametric test results and effect sizes for autonomous robot test suite parameters for different values of $\rho$}
\label{tab:ts_eff_rob}
\centering

    \begin{tabular}{cccccccccc}
    \hline
        Metric & A & B & p-value & Effect size & Metric & A & B & p-value & Effect size \\ \hline
        fitness & 0.2 & 0.4 & 0.124 & -0.043, N & diversity & 0.2 & 0.4 & 0.118 & -0.24, S \\ 
        ~ & 0.2 & 0.6 & 0.045 & -0.055, N & ~ & 0.2 & 0.6 & 0.834 & 0.033, N \\ 
        ~ & 0.2 & 0.8 & <0.001 & -0.104, N & ~ & 0.2 & 0.8 & 0.396 & -0.131, N \\ 
        ~ & 0.2 & 1 & <0.001 & -0.156, S & ~ & 0.2 & 1 & 0.994 & 0.002, N \\ 
        ~ & 0.4 & 0.6 & 0.644 & -0.013, N & ~ & 0.4 & 0.6 & 0.142 & 0.23, S \\ 
        ~ & 0.4 & 0.8 & 0.026 & -0.063, N & ~ & 0.4 & 0.8 & 0.561 & 0.092, N \\ 
        ~ & 0.4 & 1 & <0.001 & -0.112, N & ~ & 0.4 & 1 & 0.198 & 0.202, S \\ 
        ~ & 0.6 & 0.8 & 0.07 & -0.051, N & ~ & 0.6 & 0.8 & 0.283 & -0.168, S \\ 
        ~ & 0.6 & 1 & 0.001 & -0.098, N & ~ & 0.6 & 1 & 0.954 & -0.01, N \\ 
        ~ & 0.8 & 1 & 0.126 & -0.043, N & ~ & 0.8 & 1 & 0.372 & 0.14, N \\ \hline
    \end{tabular}

\end{table}

%% file: tables/rq2_effect_size_veh.tex
\begin{table}[h!]
\caption{Non-parametric test results and effect sizes for autonomous vehicle test suite parameters for different values of $\rho$}
\label{tab:ts_eff_veh}
\centering
    \begin{tabular}{cccccccccc}
    \hline
        Metric & A & B & p-value & Effect size & Metric & A & B & p-value & Effect size \\ \hline
        Fitness & 0.2 & 0.4 & 0.211 & -0.034, N & Diversity & 0.2 & 0.4 & 0.348 & -0.142, N \\ 
        ~ & 0.2 & 0.6 & 0.253 & -0.031, N & ~ & 0.2 & 0.6 & 0.819 & -0.036, N \\ 
        ~ & 0.2 & 0.8 & 0.142 & 0.04, N & ~ & 0.2 & 0.8 & 0.42 & -0.122, N \\ 
        ~ & 0.2 & 1 & 0.029 & -0.06, N & ~ & 0.2 & 1 & 0.223 & 0.184, S \\ 
        ~ & 0.4 & 0.6 & 0.933 & -0.002, N & ~ & 0.4 & 0.6 & 0.549 & 0.091, N \\ 
        ~ & 0.4 & 0.8 & 0.006 & 0.075, N & ~ & 0.4 & 0.8 & 0.83 & -0.033, N \\ 
        ~ & 0.4 & 1 & 0.248 & -0.031, N & ~ & 0.4 & 1 & 0.112 & 0.24, S \\ 
        ~ & 0.6 & 0.8 & 0.007 & 0.073, N & ~ & 0.6 & 0.8 & 0.464 & -0.111, N \\ 
        ~ & 0.6 & 1 & 0.382 & -0.024, N & ~ & 0.6 & 1 & 0.201 & 0.193, S \\ 
        ~ & 0.8 & 1 & <0.001 & -0.099, N & ~ & 0.8 & 1 & 0.158 & 0.213, S \\ \hline
    \end{tabular}

\end{table}

%% file: tables/rq2_best_value_mean.tex
\begin{table}[!ht]
    \centering
    \caption{Best value found over iterations by different configurations of RIGAA}
    \begin{tabular}{ccccccc}
    \hline
        Problem & Metric & 0.2 & 0.4 & 0.6 & 0.8 & 1 \\ \hline
        robot & Mean best value found & 243.458 & 247.744 & 248.52 & 252.134 & 253.979 \\ 
        vehicle & Mean best value found & 11.31 & 11.159 & 11.333 & 11.339 & 11.148 \\ \hline
    \end{tabular}
    \label{tab:rq2_best_val_mean}
\end{table}

%% file: tables/rq2_best_val_p.tex
\begin{table}[!ht]
    \centering
    \caption{Non-parametric test results and effect sizes for best values found over generations}
    \scalebox{0.95}{
    \begin{tabular}{cccccccccc}
    \hline
        Problem & A & B & p-value & Effect size & Problem & A & B & p-value & Effect size \\ \hline
        robot & 0.2 & 0.4 & 0.24 & -0.181, S & vehicle & 0.2 & 0.4 & 0.684 & 0.062, N \\ 
        ~ & 0.2 & 0.6 & 0.122 & -0.238, S & ~ & 0.2 & 0.6 & 0.9 & -0.02, N \\ 
        ~ & 0.2 & 0.8 & 0.087 & -0.263, S & ~ & 0.2 & 0.8 & 0.853 & -0.029, N \\ 
        ~ & 0.2 & 1 & 0.01 & -0.398, M & ~ & 0.2 & 1 & 0.652 & 0.069, N \\ 
        ~ & 0.4 & 0.6 & 0.756 & -0.05, N & ~ & 0.4 & 0.6 & 0.54 & -0.093, N \\ 
        ~ & 0.4 & 0.8 & 0.517 & -0.102, N & ~ & 0.4 & 0.8 & 0.446 & -0.116, N \\ 
        ~ & 0.4 & 1 & 0.187 & -0.207, S & ~ & 0.4 & 1 & 0.935 & -0.013, N \\ 
        ~ & 0.6 & 0.8 & 0.623 & -0.078, N & ~ & 0.6 & 0.8 & 0.971 & -0.007, N \\ 
        ~ & 0.6 & 1 & 0.354 & -0.145, N & ~ & 0.6 & 1 & 0.455 & 0.113, N \\ 
        ~ & 0.8 & 1 & 0.743 & -0.052, N & ~ & 0.8 & 1 & 0.379 & 0.133, N \\ \hline
    \end{tabular}}
    \label{tab:rq2_best_val_p}
\end{table}

%% file: tables/rq3_rob_mean.tex
\begin{table}[!ht]
    \centering
    \caption{Test suite average fitness and diversity }
    \label{rq3_mean}
    \begin{tabular}{ccccccc}
    \hline
        Problem & Metric & Random & NSGA-II & RIGAA & SEMOA & SRIGAA \\ \hline
        robot & $F_{avs}$ & 78.529 & 182.027 & 205.712 & 197.617 & 234.522 \\ 
        robot & $D_{av}$ & 0.982 & 0.764 & 0.775 & 0.62 & 0.607 \\ 
        vehicle & $F_{avs}$ & 3.719 & 8.376 & 9.125 & 8.988 & 10.382 \\ 
        vehicle & $D_{av}$ & 0.961 & 0.819 & 0.807 & 0.628 & 0.621 \\ \hline
    \end{tabular}
\end{table}

%% file: tables/rq3_rob_eff.tex
\begin{table}[!ht]
    \centering
    \caption{Non-parametric test results and effect sizes for autonomous robot test suite fitness and diversity}
    \label{rq3_rob_eff}
    \scalebox{0.8}{
    \begin{tabular}{cccccccccc}
    \hline
        Metric & A & B & p-value & Effect size & Metric & A & B & p-value & Effect size \\ \hline
        Fitness, & Random & NSGA-II & <0.001 & -1.0, L & Diversity, & Random & NSGA-II & <0.001 & 1.0, L \\ 
        $F_{avs}$ & Random & RIGAA & <0.001 & -1.0, L & $D_{av}$ & Random & RIGAA & <0.001 & 1.0, L \\ 
        ~ & Random & SEMOA & <0.001 & -1.0, L & ~ & Random & SEMOA & <0.001 & 1.0, L \\ 
        ~ & Random & SRIGAA & <0.001 & -1.0, L & ~ & Random & SRIGAA & <0.001 & 1.0, L \\ 
        ~ & NSGA-II & RIGAA & <0.001 & -0.507, L & ~ & NSGA-II & RIGAA & 0.196 & -0.196, S \\ 
        ~ & NSGA-II & SEMOA & <0.001 & -0.551, L & ~ & NSGA-II & SEMOA & <0.001 & 0.972, L \\ 
        ~ & NSGA-II & SRIGAA & <0.001 & -0.994, L & ~ & NSGA-II & SRIGAA & <0.001 & 1.0, L \\ 
        ~ & RIGAA & SEMOA & <0.001 & 0.234, S & ~ & RIGAA & SEMOA & <0.001 & 0.996, L \\ 
        ~ & RIGAA & SRIGAA & <0.001 & -0.631, L & ~ & RIGAA & SRIGAA & <0.001 & 1.0, L \\ 
        ~ & SEMOA & SRIGAA & <0.001 & -0.994, L & ~ & SEMOA & SRIGAA & 0.447 & 0.152, S \\ \hline
    \end{tabular}}
\end{table}

%% file: tables/rq3_veh_eff.tex
\begin{table}[!ht]
    \centering
    \caption{Non-parametric test results and effect sizes for autonomous vehicle test suite fitness and diversity}
        \label{rq3_veh_eff}
    \scalebox{0.8}{
    \begin{tabular}{cccccccccc}
    \hline
        Metric & A & B & p-value & Effect size & Metric & A & B & p-value & Effect size \\ \hline
        Fitness & Random & NSGA-II & <0.001 & -0.959, L & Diversity & Random & NSGA-II & <0.001 & 1.0, L \\ 
        $F_{avs}$ & Random & RIGAA & <0.001 & -0.974, L & $D_{av}$ & Random & RIGAA & <0.001 & 1.0, L \\ 
        ~ & Random & SEMOA & <0.001 & -0.974, L & ~ & Random & SEMOA & <0.001 & 1.0, L \\ 
        ~ & Random & SRIGAA & <0.001 & -0.998, L & ~ & Random & SRIGAA & <0.001 & 1.0, L \\ 
        ~ & NSGA-II & RIGAA & <0.001 & -0.319, S & ~ & NSGA-II & RIGAA & 0.277 & 0.164, S \\ 
        ~ & NSGA-II & SEMOA & <0.001 & -0.298, S & ~ & NSGA-II & SEMOA & <0.001 & 0.973, L \\ 
        ~ & NSGA-II & SRIGAA & <0.001 & -0.813, L & ~ & NSGA-II & SRIGAA & <0.001 & 0.98, L \\ 
        ~ & RIGAA & SEMOA & 0.001 & 0.091, N & ~ & RIGAA & SEMOA & <0.001 & 0.953, L \\ 
        ~ & RIGAA & SRIGAA & <0.001 & -0.521, L & ~ & RIGAA & SRIGAA & <0.001 & 0.964, L \\ 
        ~ & SEMOA & SRIGAA & <0.001
        & -0.719, L & ~ & SEMOA & SRIGAA & 0.641 & 0.071, N \\ \hline
    \end{tabular}}
\end{table}

%% file: tables/rq3_best_value_mean.tex
\begin{table}[!ht]
    \centering
    \caption{Best value found over iterations}
    \begin{tabular}{ccccccc}
    \hline
        Problem & Metric & Random & NSGA-II & RIGAA & SEMOA & SRIGAA \\ \hline
        robot & Mean best value found & 134.384 & 205.297 & 243.458 & 207.677 & 248.384 \\ 
        vehicle & Mean best value found & 8.718 & 9.925 & 11.179 & 9.403 & 11.215 \\ \hline
    \end{tabular}
    \label{tab:rq3_best_fit_mean}
\end{table}

%% file: tables/rq3_best_values_p_val.tex
\begin{table}[!ht]
    \centering
    \caption{Non-parametric test results and effect sizes for best values found over generations}
    \scalebox{0.85}{
    \begin{tabular}{cccccccccc}
    \hline
        Problem & A & B & p-value & Effect size & Problem & A & B & p-value & Effect size \\ \hline
        robot & Random & NSGA-II &<0.001& -1.0, L & vehicle & Random & NSGA-II &<0.001& -0.68, L \\ 
        ~ & Random & RIGAA &<0.001& -1.0, L & ~ & Random & RIGAA &<0.001& -0.984, L \\ 
        ~ & Random & SEMOA &<0.001& -1.0, L & ~ & Random & SEMOA & 0.002 & -0.46, M \\ 
        ~ & Random & SRIGAA &<0.001& -1.0, L & ~ & Random & SRIGAA &<0.001& -0.993, L \\ 
        ~ & NSGA-II & RIGAA &<0.001& -0.976, L & ~ & NSGA-II & RIGAA &<0.001& -0.607, L \\ 
        ~ & NSGA-II & SEMOA & 0.566 & -0.1, N & ~ & NSGA-II & SEMOA & 0.088 & 0.258, S \\ 
        ~ & NSGA-II & SRIGAA &<0.001& -0.996, L & ~ & NSGA-II & SRIGAA &<0.001& -0.642, L \\ 
        ~ & RIGAA & SEMOA &<0.001& 0.982, L & ~ & RIGAA & SEMOA &<0.001& 0.82, L \\ 
        ~ & RIGAA & SRIGAA & 0.308 & -0.182, S & ~ & RIGAA & SRIGAA & 0.877 & -0.024, N \\ 
        ~ & SEMOA & SRIGAA &<0.001& -0.994, L & ~ & SEMOA & SRIGAA &<0.001& -0.869, L \\ \hline
        \end{tabular}}
    \label{tab:rq3_best_values_p_val}
\end{table}

%% file: tables/rq4_mean.tex
\begin{table}[h!]
    \centering
    \caption{The average fitness and diversity of the test suites produced by different algorithms}
    \begin{tabular}{ccccc}
    \hline
        Problem & Metric & Random & NSGA2 & RIGAA \\ \hline
        robot & $F_{av}$ & 3.232 & 6.95 & 8.185 \\ 
        robot & $D_{av}$ & 0.982 & 0.924 & 0.919 \\ 
        vehicle & $F_{av}$ & 0.31 & 0.807 & 0.906 \\ 
        vehicle & $D_{av}$ & 0.953 & 0.806 & 0.749 \\ \hline
    \end{tabular}
    \label{rq4_mean}
\end{table}

%% file: tables/rq4_rob_eff.tex
\begin{table}[!ht]
    \centering
    \caption{Non-parametric test results and effect sizes for an autonomous ant robot test suite fitness and diversity}
    \scalebox{0.8}{
    \begin{tabular}{cccccccccc}
    \hline
        Metric & A & B & p-value & Effect size & Metric & A & B & p-value & Effect size \\ \hline
        Fitness, & Random & NSGA2 &<0.001& -0.703, L & Diversity, & Random & NSGA2 &<0.001& 1.0, L \\ 
        $F_{av}$ & Random & RIGAA &<0.001& -0.747, L & $D_{av}$ & Random & RIGAA &<0.001& 1.0, L \\ 
        ~ & NSGA2 & RIGAA & 0.004 & -0.164, S & ~ & NSGA2 & RIGAA & 0.212 & 0.34, M \\ \hline
    \end{tabular}}
    \label{rq4_rob_eff}
\end{table}

%% file: tables/rq4_veh_eff.tex
 \begin{table}[h!]
 \centering
\caption{Non-parametric test results and effect sizes for an autonomous vehicle lane keeping system test suite fitness and diversity}
\label{tab:rq4_veh_eff}
\scalebox{0.8}{
    \begin{tabular}{llllllllll}
    \hline
        Metric & A & B & p-value & Effect size & Metric & A & B & p-value & Effect size \\ \hline
        Fitness, & Random & NSGA-II & <0.001 & -0.818, L & Diversity, & Random & NSGA-II & <0.001 & 1.0, L \\ 
        $F_{av}$ & Random & RIGAA & <0.001 & -0.901, L & $D_{av}$ & Random & RIGAA & <0.001 & 1.0, L \\ 
        ~ & NSGA-II & RIGAA & <0.001 & -0.438, M & ~ & NSGA-II & RIGAA & 0.149 & 0.382, M \\ \hline
    \end{tabular}}
\end{table}

%% file: tables/rq4_best_val_full_mean.tex
\begin{table}[!ht]
    \centering
    \caption{Best fitness value of the test scenario found over iterations for the full models of the systems}
    \begin{tabular}{ccccc}
    \hline
        Problem & Metric & Random & NSGA2 & RIGAA \\ \hline
        robot & Mean best value found & 12.325 & 13.2 & 16.125 \\ 
        vehicle & Mean best value found & 0.918 & 0.974 & 0.978 \\ \hline
    \end{tabular}
    \label{tab:rq4_best_val_full_mean}
\end{table}

%% file: tables/rq4_best_val_full_p.tex
\begin{table}[!ht]
    \centering
    \caption{Non-parametric test results and effect sizes for best values found over generations for the full models of the systems under test}
    \scalebox{0.85}{
    \begin{tabular}{cccccccccc}
    \hline
        Problem & A & B & p-value & Effect size & Problem & A & B & p-value & Effect size \\ \hline
        robot & Random & NSGA2 & 0.252 & -0.31, S & vehicle & Random & NSGA-II & 0.002 & -0.879, L \\ 
        ~ & Random & RIGAA & <0.001 & -0.96, L & ~ & Random & RIGAA & 0.007 & -0.8, L \\ 
        ~ & NSGA2 & RIGAA & 0.001 & -0.85, L & ~ & NSGA-II & RIGAA & 0.13 & -0.4, M \\ \hline
    \end{tabular}}
    \label{tab:rq4_best_val_full_p}
\end{table}

%% file: tables/rq5_lkas_mean.tex
\begin{table}[h]
    \centering
    \caption{Average number of failures detected by the tools and their sparseness}
    \begin{tabular}{cccc}
    \hline
        Metric & RIGAA & AmbieGen & Frenetic \\ \hline
        Number of failures & 25.722 & 20.636 & 14.034 \\ 
        Failure sparseness & 26.228 & 25.569 & 27.084 \\ \hline
    \end{tabular}
    \label{rq4_mean_lkas}
\end{table}

%% file: tables/rq4_lkas_p.tex
\begin{table}[h]
    \centering
    \caption{Statistical testing results and effect sizes}
    \scalebox{0.8}{
    \begin{tabular}{cccccccccc}
    \hline
        Metric & A & B & p-value & Effect size & Metric & A & B & p-value & Effect size \\ \hline
        Failures & RIGAA & AmbieGen & 0.022 & 0.515, L & Sparseness & RIGAA & AmbieGen & 0.486 & 0.161, S \\ 
        ~ & RIGAA & Frenetic & <0.001 & 0.931, L & ~ & RIGAA & Frenetic & 0.437 & 0.137, L \\ 
        ~ & AmbieGen & Frenetic & <0.001 & 0.783, L & ~ & AmbieGen & Frenetic & 0.115 & 0.329, S \\ \hline
    \end{tabular}}
    \label{rq4_lkas_p}
\end{table}

%% file: main.bbl

\begin{thebibliography}{69}


\ifx \showCODEN    \undefined \def \showCODEN     #1{\unskip}     \fi
\ifx \showDOI      \undefined \def \showDOI       #1{#1}\fi
\ifx \showISBNx    \undefined \def \showISBNx     #1{\unskip}     \fi
\ifx \showISBNxiii \undefined \def \showISBNxiii  #1{\unskip}     \fi
\ifx \showISSN     \undefined \def \showISSN      #1{\unskip}     \fi
\ifx \showLCCN     \undefined \def \showLCCN      #1{\unskip}     \fi
\ifx \shownote     \undefined \def \shownote      #1{#1}          \fi
\ifx \showarticletitle \undefined \def \showarticletitle #1{#1}   \fi
\ifx \showURL      \undefined \def \showURL       {\relax}        \fi
\providecommand\bibfield[2]{#2}
\providecommand\bibinfo[2]{#2}
\providecommand\natexlab[1]{#1}
\providecommand\showeprint[2][]{arXiv:#2}

\bibitem[Abdessalem et~al\mbox{.}(2018a)]%
        {abdessalem2018testing}
\bibfield{author}{\bibinfo{person}{Raja~Ben Abdessalem}, \bibinfo{person}{Shiva
  Nejati}, \bibinfo{person}{Lionel~C Briand}, {and} \bibinfo{person}{Thomas
  Stifter}.} \bibinfo{year}{2018}\natexlab{a}.
\newblock \showarticletitle{Testing vision-based control systems using
  learnable evolutionary algorithms}. In \bibinfo{booktitle}{\emph{2018
  IEEE/ACM 40th International Conference on Software Engineering (ICSE)}}.
  IEEE, \bibinfo{pages}{1016--1026}.
\newblock


\bibitem[Abdessalem et~al\mbox{.}(2018b)]%
        {abdessalem2018testing2}
\bibfield{author}{\bibinfo{person}{Raja~Ben Abdessalem},
  \bibinfo{person}{Annibale Panichella}, \bibinfo{person}{Shiva Nejati},
  \bibinfo{person}{Lionel~C Briand}, {and} \bibinfo{person}{Thomas Stifter}.}
  \bibinfo{year}{2018}\natexlab{b}.
\newblock \showarticletitle{Testing autonomous cars for feature interaction
  failures using many-objective search}. In
  \bibinfo{booktitle}{\emph{Proceedings of the 33rd ACM/IEEE International
  Conference on Automated Software Engineering}}. \bibinfo{pages}{143--154}.
\newblock


\bibitem[Ammann and Offutt(2016)]%
        {ammann_offutt_2016}
\bibfield{author}{\bibinfo{person}{Paul Ammann} {and} \bibinfo{person}{Jeff
  Offutt}.} \bibinfo{year}{2016}\natexlab{}.
\newblock \bibinfo{booktitle}{\emph{Introduction to software testing}}.
\newblock \bibinfo{publisher}{Cambridge University Press}.
\newblock


\bibitem[Antoniol et~al\mbox{.}(2005)]%
        {antoniol2005search}
\bibfield{author}{\bibinfo{person}{Giuliano Antoniol},
  \bibinfo{person}{Massimiliano Di~Penta}, {and} \bibinfo{person}{Mark
  Harman}.} \bibinfo{year}{2005}\natexlab{}.
\newblock \showarticletitle{Search-based techniques applied to optimization of
  project planning for a massive maintenance project}. In
  \bibinfo{booktitle}{\emph{21st IEEE International Conference on Software
  Maintenance (ICSM'05)}}. IEEE, \bibinfo{pages}{240--249}.
\newblock


\bibitem[Arcuri and Briand(2014)]%
        {arcuri2014hitchhiker}
\bibfield{author}{\bibinfo{person}{Andrea Arcuri} {and} \bibinfo{person}{Lionel
  Briand}.} \bibinfo{year}{2014}\natexlab{}.
\newblock \showarticletitle{A hitchhiker's guide to statistical tests for
  assessing randomized algorithms in software engineering}.
\newblock \bibinfo{journal}{\emph{Software Testing, Verification and
  Reliability}} \bibinfo{volume}{24}, \bibinfo{number}{3}
  (\bibinfo{year}{2014}), \bibinfo{pages}{219--250}.
\newblock


\bibitem[Arnold and Alexander(2013)]%
        {arnold2013testing}
\bibfield{author}{\bibinfo{person}{James Arnold} {and} \bibinfo{person}{Rob
  Alexander}.} \bibinfo{year}{2013}\natexlab{}.
\newblock \showarticletitle{Testing autonomous robot control software using
  procedural content generation}. In \bibinfo{booktitle}{\emph{Computer Safety,
  Reliability, and Security: 32nd International Conference, SAFECOMP 2013,
  Toulouse, France, September 24-27, 2013. Proceedings 32}}. Springer,
  \bibinfo{pages}{33--44}.
\newblock


\bibitem[B{\"a}ck et~al\mbox{.}(1997)]%
        {back1997handbook}
\bibfield{author}{\bibinfo{person}{Thomas B{\"a}ck}, \bibinfo{person}{David~B
  Fogel}, {and} \bibinfo{person}{Zbigniew Michalewicz}.}
  \bibinfo{year}{1997}\natexlab{}.
\newblock \showarticletitle{Handbook of evolutionary computation}.
\newblock \bibinfo{journal}{\emph{Release}} \bibinfo{volume}{97},
  \bibinfo{number}{1} (\bibinfo{year}{1997}), \bibinfo{pages}{B1}.
\newblock


\bibitem[BeamNG.tech(2021)]%
        {beamng}
\bibfield{author}{\bibinfo{person}{BeamNG.tech}.}
  \bibinfo{year}{2021}\natexlab{}.
\newblock \bibinfo{title}{BeamNG GmbH.}
\newblock
\newblock
\urldef\tempurl%
\url{https://www.beamng.gmbh/research}
\showURL{%
\tempurl}


\bibitem[Berner et~al\mbox{.}(2019)]%
        {berner2019dota}
\bibfield{author}{\bibinfo{person}{Christopher Berner}, \bibinfo{person}{Greg
  Brockman}, \bibinfo{person}{Brooke Chan}, \bibinfo{person}{Vicki Cheung},
  \bibinfo{person}{Przemys{\l}aw D{\k{e}}biak}, \bibinfo{person}{Christy
  Dennison}, \bibinfo{person}{David Farhi}, \bibinfo{person}{Quirin Fischer},
  \bibinfo{person}{Shariq Hashme}, \bibinfo{person}{Chris Hesse},
  {et~al\mbox{.}}} \bibinfo{year}{2019}\natexlab{}.
\newblock \showarticletitle{Dota 2 with large scale deep reinforcement
  learning}.
\newblock \bibinfo{journal}{\emph{arXiv preprint arXiv:1912.06680}}
  (\bibinfo{year}{2019}).
\newblock


\bibitem[Beume et~al\mbox{.}(2007)]%
        {beume2007sms}
\bibfield{author}{\bibinfo{person}{Nicola Beume}, \bibinfo{person}{Boris
  Naujoks}, {and} \bibinfo{person}{Michael Emmerich}.}
  \bibinfo{year}{2007}\natexlab{}.
\newblock \showarticletitle{SMS-EMOA: Multiobjective selection based on
  dominated hypervolume}.
\newblock \bibinfo{journal}{\emph{European Journal of Operational Research}}
  \bibinfo{volume}{181}, \bibinfo{number}{3} (\bibinfo{year}{2007}),
  \bibinfo{pages}{1653--1669}.
\newblock


\bibitem[Birchler et~al\mbox{.}(2022)]%
        {birchler2022cost}
\bibfield{author}{\bibinfo{person}{Christian Birchler},
  \bibinfo{person}{Nicolas Ganz}, \bibinfo{person}{Sajad Khatiri},
  \bibinfo{person}{Alessio Gambi}, {and} \bibinfo{person}{Sebastiano
  Panichella}.} \bibinfo{year}{2022}\natexlab{}.
\newblock \showarticletitle{Cost-effective simulation-based test selection in
  self-driving cars software with SDC-Scissor}. In
  \bibinfo{booktitle}{\emph{2022 IEEE International Conference on Software
  Analysis, Evolution and Reengineering (SANER)}}. IEEE,
  \bibinfo{pages}{164--168}.
\newblock


\bibitem[{Blank} and {Deb}(2020)]%
        {pymoo}
\bibfield{author}{\bibinfo{person}{J. {Blank}} {and} \bibinfo{person}{K.
  {Deb}}.} \bibinfo{year}{2020}\natexlab{}.
\newblock \showarticletitle{pymoo: Multi-Objective Optimization in Python}.
\newblock \bibinfo{journal}{\emph{IEEE Access}}  \bibinfo{volume}{8}
  (\bibinfo{year}{2020}), \bibinfo{pages}{89497--89509}.
\newblock


\bibitem[Briffoteaux(2022)]%
        {briffoteaux2022parallel}
\bibfield{author}{\bibinfo{person}{Guillaume Briffoteaux}.}
  \bibinfo{year}{2022}\natexlab{}.
\newblock \emph{\bibinfo{title}{Parallel surrogate-based algorithms for solving
  expensive optimization problems}}.
\newblock \bibinfo{thesistype}{Ph.\,D. Dissertation}.
  \bibinfo{school}{Universit{\'e} de Lille; Universit{\'e} de Mons (UMONS)}.
\newblock


\bibitem[Burda et~al\mbox{.}(2018)]%
        {burda2018exploration}
\bibfield{author}{\bibinfo{person}{Yuri Burda}, \bibinfo{person}{Harrison
  Edwards}, \bibinfo{person}{Amos Storkey}, {and} \bibinfo{person}{Oleg
  Klimov}.} \bibinfo{year}{2018}\natexlab{}.
\newblock \showarticletitle{Exploration by random network distillation}.
\newblock \bibinfo{journal}{\emph{arXiv preprint arXiv:1810.12894}}
  (\bibinfo{year}{2018}).
\newblock


\bibitem[Campion et~al\mbox{.}(1996)]%
        {campion1996structural}
\bibfield{author}{\bibinfo{person}{Guy Campion}, \bibinfo{person}{Georges
  Bastin}, {and} \bibinfo{person}{Brigitte Dandrea-Novel}.}
  \bibinfo{year}{1996}\natexlab{}.
\newblock \showarticletitle{Structural properties and classification of
  kinematic and dynamic models of wheeled mobile robots}.
\newblock \bibinfo{journal}{\emph{IEEE transactions on robotics and
  automation}} \bibinfo{volume}{12}, \bibinfo{number}{1}
  (\bibinfo{year}{1996}), \bibinfo{pages}{47--62}.
\newblock


\bibitem[Castellano et~al\mbox{.}(2021)]%
        {castellano2021frenetic}
\bibfield{author}{\bibinfo{person}{Ezequiel Castellano}, \bibinfo{person}{Ahmet
  Cetinkaya}, \bibinfo{person}{C{\'e}dric~Ho Thanh}, \bibinfo{person}{Stefan
  Klikovits}, \bibinfo{person}{Xiaoyi Zhang}, {and} \bibinfo{person}{Paolo
  Arcaini}.} \bibinfo{year}{2021}\natexlab{}.
\newblock \showarticletitle{Frenetic at the SBST 2021 tool competition}. In
  \bibinfo{booktitle}{\emph{2021 IEEE/ACM 14th International Workshop on
  Search-Based Software Testing (SBST)}}. IEEE, \bibinfo{pages}{36--37}.
\newblock


\bibitem[Chen et~al\mbox{.}(2021)]%
        {chen2021adversarial}
\bibfield{author}{\bibinfo{person}{Baiming Chen}, \bibinfo{person}{Xiang Chen},
  \bibinfo{person}{Qiong Wu}, {and} \bibinfo{person}{Liang Li}.}
  \bibinfo{year}{2021}\natexlab{}.
\newblock \showarticletitle{Adversarial evaluation of autonomous vehicles in
  lane-change scenarios}.
\newblock \bibinfo{journal}{\emph{IEEE Transactions on Intelligent
  Transportation Systems}} \bibinfo{volume}{23}, \bibinfo{number}{8}
  (\bibinfo{year}{2021}), \bibinfo{pages}{10333--10342}.
\newblock


\bibitem[Chen et~al\mbox{.}(2020)]%
        {chen2020reinforcement}
\bibfield{author}{\bibinfo{person}{Qiong Chen}, \bibinfo{person}{Mengxing
  Huang}, \bibinfo{person}{Qiannan Xu}, \bibinfo{person}{Hao Wang}, {and}
  \bibinfo{person}{Jinghui Wang}.} \bibinfo{year}{2020}\natexlab{}.
\newblock \showarticletitle{Reinforcement learning-based genetic algorithm in
  optimizing multidimensional data discretization scheme}.
\newblock \bibinfo{journal}{\emph{Mathematical Problems in Engineering}}
  \bibinfo{volume}{2020} (\bibinfo{year}{2020}).
\newblock


\bibitem[Chung et~al\mbox{.}(2019)]%
        {chung2019jaccard}
\bibfield{author}{\bibinfo{person}{Neo~Christopher Chung},
  \bibinfo{person}{B{\l}a{\.Z}ej Miasojedow}, \bibinfo{person}{Micha{\l}
  Startek}, {and} \bibinfo{person}{Anna Gambin}.}
  \bibinfo{year}{2019}\natexlab{}.
\newblock \showarticletitle{Jaccard/Tanimoto similarity test and estimation
  methods for biological presence-absence data}.
\newblock \bibinfo{journal}{\emph{BMC bioinformatics}} \bibinfo{volume}{20},
  \bibinfo{number}{15} (\bibinfo{year}{2019}), \bibinfo{pages}{1--11}.
\newblock


\bibitem[Coello et~al\mbox{.}(2007)]%
        {coello2007evolutionary}
\bibfield{author}{\bibinfo{person}{Carlos A~Coello Coello},
  \bibinfo{person}{Gary~B Lamont}, \bibinfo{person}{David~A Van~Veldhuizen},
  {et~al\mbox{.}}} \bibinfo{year}{2007}\natexlab{}.
\newblock \bibinfo{booktitle}{\emph{Evolutionary algorithms for solving
  multi-objective problems}}. Vol.~\bibinfo{volume}{5}.
\newblock \bibinfo{publisher}{Springer}.
\newblock


\bibitem[Coulter(1992)]%
        {coulter1992implementation}
\bibfield{author}{\bibinfo{person}{R~Craig Coulter}.}
  \bibinfo{year}{1992}\natexlab{}.
\newblock \bibinfo{booktitle}{\emph{Implementation of the pure pursuit path
  tracking algorithm}}.
\newblock \bibinfo{type}{{T}echnical {R}eport}.
  \bibinfo{institution}{Carnegie-Mellon UNIV Pittsburgh PA Robotics INST}.
\newblock


\bibitem[Deb and Jain(2013)]%
        {deb2013evolutionary}
\bibfield{author}{\bibinfo{person}{Kalyanmoy Deb} {and}
  \bibinfo{person}{Himanshu Jain}.} \bibinfo{year}{2013}\natexlab{}.
\newblock \showarticletitle{An evolutionary many-objective optimization
  algorithm using reference-point-based nondominated sorting approach, part I:
  solving problems with box constraints}.
\newblock \bibinfo{journal}{\emph{IEEE transactions on evolutionary
  computation}} \bibinfo{volume}{18}, \bibinfo{number}{4}
  (\bibinfo{year}{2013}), \bibinfo{pages}{577--601}.
\newblock


\bibitem[Deb et~al\mbox{.}(2002)]%
        {deb2002fast}
\bibfield{author}{\bibinfo{person}{Kalyanmoy Deb}, \bibinfo{person}{Amrit
  Pratap}, \bibinfo{person}{Sameer Agarwal}, {and} \bibinfo{person}{TAMT
  Meyarivan}.} \bibinfo{year}{2002}\natexlab{}.
\newblock \showarticletitle{A fast and elitist multiobjective genetic
  algorithm: NSGA-II}.
\newblock \bibinfo{journal}{\emph{IEEE transactions on evolutionary
  computation}} \bibinfo{volume}{6}, \bibinfo{number}{2}
  (\bibinfo{year}{2002}), \bibinfo{pages}{182--197}.
\newblock


\bibitem[Eiben and Smith(2015)]%
        {eiben2015introduction}
\bibfield{author}{\bibinfo{person}{Agoston~E Eiben} {and}
  \bibinfo{person}{James~E Smith}.} \bibinfo{year}{2015}\natexlab{}.
\newblock \bibinfo{booktitle}{\emph{Introduction to evolutionary computing}}.
\newblock \bibinfo{publisher}{Springer}.
\newblock


\bibitem[foundation(2020)]%
        {ant-gym-env}
\bibfield{author}{\bibinfo{person}{Farma foundation}.}
  \bibinfo{year}{2020}\natexlab{}.
\newblock \bibinfo{title}{Ant gym}.
\newblock
\newblock
\urldef\tempurl%
\url{https://gymnasium.farama.org/environments/mujoco/ant/}
\showURL{%
\tempurl}


\bibitem[Fu et~al\mbox{.}(2020)]%
        {fu2020d4rl}
\bibfield{author}{\bibinfo{person}{Justin Fu}, \bibinfo{person}{Aviral Kumar},
  \bibinfo{person}{Ofir Nachum}, \bibinfo{person}{George Tucker}, {and}
  \bibinfo{person}{Sergey Levine}.} \bibinfo{year}{2020}\natexlab{}.
\newblock \bibinfo{title}{D4RL: Datasets for Deep Data-Driven Reinforcement
  Learning}.
\newblock
\newblock
\showeprint[arxiv]{2004.07219}~[cs.LG]


\bibitem[Gambi et~al\mbox{.}(2022a)]%
        {gambi2022sbst}
\bibfield{author}{\bibinfo{person}{Alessio Gambi}, \bibinfo{person}{Gunel
  Jahangirova}, \bibinfo{person}{Vincenzo Riccio}, {and}
  \bibinfo{person}{Fiorella Zampetti}.} \bibinfo{year}{2022}\natexlab{a}.
\newblock \showarticletitle{SBST tool competition 2022}. In
  \bibinfo{booktitle}{\emph{2022 IEEE/ACM 15th International Workshop on
  Search-Based Software Testing (SBST)}}. IEEE, \bibinfo{pages}{25--32}.
\newblock


\bibitem[Gambi et~al\mbox{.}(2019)]%
        {gambi2019automatically}
\bibfield{author}{\bibinfo{person}{Alessio Gambi}, \bibinfo{person}{Marc
  Mueller}, {and} \bibinfo{person}{Gordon Fraser}.}
  \bibinfo{year}{2019}\natexlab{}.
\newblock \showarticletitle{Automatically testing self-driving cars with
  search-based procedural content generation}. In
  \bibinfo{booktitle}{\emph{Proceedings of the 28th ACM SIGSOFT International
  Symposium on Software Testing and Analysis}}. IEEE,
  \bibinfo{pages}{318--328}.
\newblock


\bibitem[Gambi et~al\mbox{.}(2022b)]%
        {gambi2022generating}
\bibfield{author}{\bibinfo{person}{Alessio Gambi}, \bibinfo{person}{Vuong
  Nguyen}, \bibinfo{person}{Jasim Ahmed}, {and} \bibinfo{person}{Gordon
  Fraser}.} \bibinfo{year}{2022}\natexlab{b}.
\newblock \showarticletitle{Generating Critical Driving Scenarios from Accident
  Sketches}. In \bibinfo{booktitle}{\emph{2022 IEEE International Conference On
  Artificial Intelligence Testing (AITest)}}. IEEE, \bibinfo{pages}{95--102}.
\newblock


\bibitem[Goldberg and Deb(1991)]%
        {goldberg1991comparative}
\bibfield{author}{\bibinfo{person}{David~E Goldberg} {and}
  \bibinfo{person}{Kalyanmoy Deb}.} \bibinfo{year}{1991}\natexlab{}.
\newblock \showarticletitle{A comparative analysis of selection schemes used in
  genetic algorithms}.
\newblock In \bibinfo{booktitle}{\emph{Foundations of genetic algorithms}}.
  Vol.~\bibinfo{volume}{1}. \bibinfo{publisher}{Elsevier},
  \bibinfo{pages}{69--93}.
\newblock


\bibitem[Haq et~al\mbox{.}(2022a)]%
        {haq2022efficient}
\bibfield{author}{\bibinfo{person}{Fitash~Ul Haq}, \bibinfo{person}{Donghwan
  Shin}, {and} \bibinfo{person}{Lionel Briand}.}
  \bibinfo{year}{2022}\natexlab{a}.
\newblock \showarticletitle{Efficient online testing for DNN-enabled systems
  using surrogate-assisted and many-objective optimization}. In
  \bibinfo{booktitle}{\emph{Proceedings of the 44th International Conference on
  Software Engineering}}. \bibinfo{pages}{811--822}.
\newblock


\bibitem[Haq et~al\mbox{.}(2022b)]%
        {haq2022many}
\bibfield{author}{\bibinfo{person}{Fitash~Ul Haq}, \bibinfo{person}{Donghwan
  Shin}, {and} \bibinfo{person}{Lionel Briand}.}
  \bibinfo{year}{2022}\natexlab{b}.
\newblock \showarticletitle{Many-Objective Reinforcement Learning for Online
  Testing of DNN-Enabled Systems}.
\newblock \bibinfo{journal}{\emph{arXiv preprint arXiv:2210.15432}}
  (\bibinfo{year}{2022}).
\newblock


\bibitem[Haq et~al\mbox{.}(2022c)]%
        {haq2022manyobjective}
\bibfield{author}{\bibinfo{person}{Fitash~Ul Haq}, \bibinfo{person}{Donghwan
  Shin}, {and} \bibinfo{person}{Lionel Briand}.}
  \bibinfo{year}{2022}\natexlab{c}.
\newblock \bibinfo{title}{Many-Objective Reinforcement Learning for Online
  Testing of DNN-Enabled Systems}.
\newblock
\newblock
\showeprint[arxiv]{2210.15432}~[cs.LG]


\bibitem[Harman and Jones(2001)]%
        {harman2001search}
\bibfield{author}{\bibinfo{person}{Mark Harman} {and} \bibinfo{person}{Bryan~F
  Jones}.} \bibinfo{year}{2001}\natexlab{}.
\newblock \showarticletitle{Search-based software engineering}.
\newblock \bibinfo{journal}{\emph{Information and software Technology}}
  \bibinfo{volume}{43}, \bibinfo{number}{14} (\bibinfo{year}{2001}),
  \bibinfo{pages}{833--839}.
\newblock


\bibitem[Harman et~al\mbox{.}(2008)]%
        {harman2008search}
\bibfield{author}{\bibinfo{person}{Mark Harman}, \bibinfo{person}{Phil McMinn},
  \bibinfo{person}{Jerffeson~Teixeira De~Souza}, {and} \bibinfo{person}{Shin
  Yoo}.} \bibinfo{year}{2008}\natexlab{}.
\newblock \showarticletitle{Search based software engineering: Techniques,
  taxonomy, tutorial}.
\newblock In \bibinfo{booktitle}{\emph{LASER Summer School on Software
  Engineering}}. \bibinfo{publisher}{Springer}, \bibinfo{pages}{1--59}.
\newblock


\bibitem[Heess et~al\mbox{.}(2017)]%
        {heess2017emergence}
\bibfield{author}{\bibinfo{person}{Nicolas Heess}, \bibinfo{person}{Dhruva Tb},
  \bibinfo{person}{Srinivasan Sriram}, \bibinfo{person}{Jay Lemmon},
  \bibinfo{person}{Josh Merel}, \bibinfo{person}{Greg Wayne},
  \bibinfo{person}{Yuval Tassa}, \bibinfo{person}{Tom Erez},
  \bibinfo{person}{Ziyu Wang}, \bibinfo{person}{SM Eslami}, {et~al\mbox{.}}}
  \bibinfo{year}{2017}\natexlab{}.
\newblock \showarticletitle{Emergence of locomotion behaviours in rich
  environments}.
\newblock \bibinfo{journal}{\emph{arXiv preprint arXiv:1707.02286}}
  (\bibinfo{year}{2017}).
\newblock


\bibitem[Hill et~al\mbox{.}(2018)]%
        {stable-baselines}
\bibfield{author}{\bibinfo{person}{Ashley Hill}, \bibinfo{person}{Antonin
  Raffin}, \bibinfo{person}{Maximilian Ernestus}, \bibinfo{person}{Adam
  Gleave}, \bibinfo{person}{Anssi Kanervisto}, \bibinfo{person}{Rene Traore},
  \bibinfo{person}{Prafulla Dhariwal}, \bibinfo{person}{Christopher Hesse},
  \bibinfo{person}{Oleg Klimov}, \bibinfo{person}{Alex Nichol},
  \bibinfo{person}{Matthias Plappert}, \bibinfo{person}{Alec Radford},
  \bibinfo{person}{John Schulman}, \bibinfo{person}{Szymon Sidor}, {and}
  \bibinfo{person}{Yuhuai Wu}.} \bibinfo{year}{2018}\natexlab{}.
\newblock \bibinfo{title}{Stable Baselines}.
\newblock
  \bibinfo{howpublished}{\url{https://github.com/hill-a/stable-baselines}}.
\newblock


\bibitem[Hoxha et~al\mbox{.}(2014)]%
        {hoxha2014towards}
\bibfield{author}{\bibinfo{person}{Bardh Hoxha}, \bibinfo{person}{Hoang Bach},
  \bibinfo{person}{Houssam Abbas}, \bibinfo{person}{Adel Dokhanchi},
  \bibinfo{person}{Yoshihiro Kobayashi}, {and} \bibinfo{person}{Georgios
  Fainekos}.} \bibinfo{year}{2014}\natexlab{}.
\newblock \showarticletitle{Towards formal specification visualization for
  testing and monitoring of cyber-physical systems}. In
  \bibinfo{booktitle}{\emph{Int. Workshop on Design and Implementation of
  Formal Tools and Systems}}. sn.
\newblock


\bibitem[Humeniuk et~al\mbox{.}(2022)]%
        {humeniuk2022search}
\bibfield{author}{\bibinfo{person}{Dmytro Humeniuk}, \bibinfo{person}{Foutse
  Khomh}, {and} \bibinfo{person}{Giuliano Antoniol}.}
  \bibinfo{year}{2022}\natexlab{}.
\newblock \showarticletitle{A search-based framework for automatic generation
  of testing environments for cyber-physical systems}.
\newblock \bibinfo{journal}{\emph{Information and Software Technology}}
  (\bibinfo{year}{2022}), \bibinfo{pages}{106936}.
\newblock


\bibitem[Humeniuk et~al\mbox{.}(2023)]%
        {dmytro_humeniuk_2023_8242223}
\bibfield{author}{\bibinfo{person}{Dmytro Humeniuk}, \bibinfo{person}{Foutse
  Khomh}, {and} \bibinfo{person}{Giuliano Antoniol}.}
  \bibinfo{year}{2023}\natexlab{}.
\newblock \bibinfo{booktitle}{\emph{{Reinforcement learning informed
  evolutionary search for autonomous system testing}}}.
\newblock
\urldef\tempurl%
\url{https://doi.org/10.5281/zenodo.8242223}
\showDOI{\tempurl}


\bibitem[Huttenlocher et~al\mbox{.}(1993)]%
        {huttenlocher1993comparing}
\bibfield{author}{\bibinfo{person}{Daniel~P Huttenlocher},
  \bibinfo{person}{Gregory~A. Klanderman}, {and} \bibinfo{person}{William~J
  Rucklidge}.} \bibinfo{year}{1993}\natexlab{}.
\newblock \showarticletitle{Comparing images using the Hausdorff distance}.
\newblock \bibinfo{journal}{\emph{IEEE Transactions on pattern analysis and
  machine intelligence}} \bibinfo{volume}{15}, \bibinfo{number}{9}
  (\bibinfo{year}{1993}), \bibinfo{pages}{850--863}.
\newblock


\bibitem[Konda and Tsitsiklis(1999)]%
        {konda1999actor}
\bibfield{author}{\bibinfo{person}{Vijay Konda} {and} \bibinfo{person}{John
  Tsitsiklis}.} \bibinfo{year}{1999}\natexlab{}.
\newblock \showarticletitle{Actor-critic algorithms}.
\newblock \bibinfo{journal}{\emph{Advances in neural information processing
  systems}}  \bibinfo{volume}{12} (\bibinfo{year}{1999}).
\newblock


\bibitem[Koza et~al\mbox{.}(1994)]%
        {koza1994genetic}
\bibfield{author}{\bibinfo{person}{John~R Koza} {et~al\mbox{.}}}
  \bibinfo{year}{1994}\natexlab{}.
\newblock \bibinfo{booktitle}{\emph{Genetic programming II}}.
  Vol.~\bibinfo{volume}{17}.
\newblock \bibinfo{publisher}{MIT press Cambridge}.
\newblock


\bibitem[Lee et~al\mbox{.}(2020)]%
        {lee2020adaptive}
\bibfield{author}{\bibinfo{person}{Ritchie Lee}, \bibinfo{person}{Ole~J
  Mengshoel}, \bibinfo{person}{Anshu Saksena}, \bibinfo{person}{Ryan~W
  Gardner}, \bibinfo{person}{Daniel Genin}, \bibinfo{person}{Joshua
  Silbermann}, \bibinfo{person}{Michael Owen}, {and} \bibinfo{person}{Mykel~J
  Kochenderfer}.} \bibinfo{year}{2020}\natexlab{}.
\newblock \showarticletitle{Adaptive stress testing: Finding likely failure
  events with reinforcement learning}.
\newblock \bibinfo{journal}{\emph{Journal of Artificial Intelligence Research}}
   \bibinfo{volume}{69} (\bibinfo{year}{2020}), \bibinfo{pages}{1165--1201}.
\newblock


\bibitem[Lehman and Stanley(2011)]%
        {lehman2011evolving}
\bibfield{author}{\bibinfo{person}{Joel Lehman} {and}
  \bibinfo{person}{Kenneth~O Stanley}.} \bibinfo{year}{2011}\natexlab{}.
\newblock \showarticletitle{Evolving a diversity of virtual creatures through
  novelty search and local competition}. In
  \bibinfo{booktitle}{\emph{Proceedings of the 13th annual conference on
  Genetic and evolutionary computation}}. \bibinfo{pages}{211--218}.
\newblock


\bibitem[Macbeth et~al\mbox{.}(2011)]%
        {macbeth2011cliff}
\bibfield{author}{\bibinfo{person}{Guillermo Macbeth}, \bibinfo{person}{Eugenia
  Razumiejczyk}, {and} \bibinfo{person}{Rub{\'e}n~Daniel Ledesma}.}
  \bibinfo{year}{2011}\natexlab{}.
\newblock \showarticletitle{Cliff's Delta Calculator: A non-parametric effect
  size program for two groups of observations}.
\newblock \bibinfo{journal}{\emph{Universitas Psychologica}}
  \bibinfo{volume}{10}, \bibinfo{number}{2} (\bibinfo{year}{2011}),
  \bibinfo{pages}{545--555}.
\newblock


\bibitem[Mann and Whitney(1947)]%
        {mann1947test}
\bibfield{author}{\bibinfo{person}{Henry~B Mann} {and}
  \bibinfo{person}{Donald~R Whitney}.} \bibinfo{year}{1947}\natexlab{}.
\newblock \showarticletitle{On a test of whether one of two random variables is
  stochastically larger than the other}.
\newblock \bibinfo{journal}{\emph{The annals of mathematical statistics}}
  (\bibinfo{year}{1947}), \bibinfo{pages}{50--60}.
\newblock


\bibitem[McKay et~al\mbox{.}(2000)]%
        {mckay2000comparison}
\bibfield{author}{\bibinfo{person}{Michael~D McKay}, \bibinfo{person}{Richard~J
  Beckman}, {and} \bibinfo{person}{William~J Conover}.}
  \bibinfo{year}{2000}\natexlab{}.
\newblock \showarticletitle{A comparison of three methods for selecting values
  of input variables in the analysis of output from a computer code}.
\newblock \bibinfo{journal}{\emph{Technometrics}} \bibinfo{volume}{42},
  \bibinfo{number}{1} (\bibinfo{year}{2000}), \bibinfo{pages}{55--61}.
\newblock


\bibitem[Mnih et~al\mbox{.}(2016)]%
        {mnih2016asynchronous}
\bibfield{author}{\bibinfo{person}{Volodymyr Mnih},
  \bibinfo{person}{Adria~Puigdomenech Badia}, \bibinfo{person}{Mehdi Mirza},
  \bibinfo{person}{Alex Graves}, \bibinfo{person}{Timothy Lillicrap},
  \bibinfo{person}{Tim Harley}, \bibinfo{person}{David Silver}, {and}
  \bibinfo{person}{Koray Kavukcuoglu}.} \bibinfo{year}{2016}\natexlab{}.
\newblock \showarticletitle{Asynchronous methods for deep reinforcement
  learning}. In \bibinfo{booktitle}{\emph{International conference on machine
  learning}}. PMLR, \bibinfo{pages}{1928--1937}.
\newblock


\bibitem[Mnih et~al\mbox{.}(2013)]%
        {mnih2013playing}
\bibfield{author}{\bibinfo{person}{Volodymyr Mnih}, \bibinfo{person}{Koray
  Kavukcuoglu}, \bibinfo{person}{David Silver}, \bibinfo{person}{Alex Graves},
  \bibinfo{person}{Ioannis Antonoglou}, \bibinfo{person}{Daan Wierstra}, {and}
  \bibinfo{person}{Martin Riedmiller}.} \bibinfo{year}{2013}\natexlab{}.
\newblock \showarticletitle{Playing atari with deep reinforcement learning}.
\newblock \bibinfo{journal}{\emph{arXiv preprint arXiv:1312.5602}}
  (\bibinfo{year}{2013}).
\newblock


\bibitem[Moghadam et~al\mbox{.}(2022)]%
        {moghadam2022machine}
\bibfield{author}{\bibinfo{person}{Mahshid~Helali Moghadam},
  \bibinfo{person}{Markus Borg}, \bibinfo{person}{Mehrdad Saadatmand},
  \bibinfo{person}{Seyed~Jalaleddin Mousavirad}, \bibinfo{person}{Markus
  Bohlin}, {and} \bibinfo{person}{Bj{\"o}rn Lisper}.}
  \bibinfo{year}{2022}\natexlab{}.
\newblock \showarticletitle{Machine Learning Testing in an ADAS Case Study
  Using Simulation-Integrated Bio-Inspired Search-Based Testing}.
\newblock \bibinfo{journal}{\emph{arXiv preprint arXiv:2203.12026}}
  (\bibinfo{year}{2022}).
\newblock


\bibitem[Panichella et~al\mbox{.}(2021)]%
        {panichella2021sbst}
\bibfield{author}{\bibinfo{person}{Sebastiano Panichella},
  \bibinfo{person}{Alessio Gambi}, \bibinfo{person}{Fiorella Zampetti}, {and}
  \bibinfo{person}{Vincenzo Riccio}.} \bibinfo{year}{2021}\natexlab{}.
\newblock \showarticletitle{Sbst tool competition 2021}. In
  \bibinfo{booktitle}{\emph{2021 IEEE/ACM 14th International Workshop on
  Search-Based Software Testing (SBST)}}. IEEE, \bibinfo{pages}{20--27}.
\newblock


\bibitem[Pettinger and Everson(2002)]%
        {pettinger2002controlling}
\bibfield{author}{\bibinfo{person}{James~E Pettinger} {and}
  \bibinfo{person}{Richard~M Everson}.} \bibinfo{year}{2002}\natexlab{}.
\newblock \showarticletitle{Controlling genetic algorithms with reinforcement
  learning}. In \bibinfo{booktitle}{\emph{Proceedings of the 4th annual
  conference on genetic and evolutionary computation}}.
  \bibinfo{pages}{692--692}.
\newblock


\bibitem[Plummer(2006)]%
        {plummer2006model}
\bibfield{author}{\bibinfo{person}{Andrew~R Plummer}.}
  \bibinfo{year}{2006}\natexlab{}.
\newblock \showarticletitle{Model-in-the-loop testing}.
\newblock \bibinfo{journal}{\emph{Proceedings of the Institution of Mechanical
  Engineers, Part I: Journal of Systems and Control Engineering}}
  \bibinfo{volume}{220}, \bibinfo{number}{3} (\bibinfo{year}{2006}),
  \bibinfo{pages}{183--199}.
\newblock


\bibitem[Quevedo et~al\mbox{.}(2021)]%
        {quevedo2021using}
\bibfield{author}{\bibinfo{person}{Jose Quevedo}, \bibinfo{person}{Marwan
  Abdelatti}, \bibinfo{person}{Farhad Imani}, {and} \bibinfo{person}{Manbir
  Sodhi}.} \bibinfo{year}{2021}\natexlab{}.
\newblock \showarticletitle{Using reinforcement learning for tuning genetic
  algorithms}. In \bibinfo{booktitle}{\emph{Proceedings of the Genetic and
  Evolutionary Computation Conference Companion}}. \bibinfo{pages}{1503--1507}.
\newblock


\bibitem[Raffin et~al\mbox{.}(2021)]%
        {stable-baselines3}
\bibfield{author}{\bibinfo{person}{Antonin Raffin}, \bibinfo{person}{Ashley
  Hill}, \bibinfo{person}{Adam Gleave}, \bibinfo{person}{Anssi Kanervisto},
  \bibinfo{person}{Maximilian Ernestus}, {and} \bibinfo{person}{Noah Dormann}.}
  \bibinfo{year}{2021}\natexlab{}.
\newblock \showarticletitle{Stable-Baselines3: Reliable Reinforcement Learning
  Implementations}.
\newblock \bibinfo{journal}{\emph{Journal of Machine Learning Research}}
  \bibinfo{volume}{22}, \bibinfo{number}{268} (\bibinfo{year}{2021}),
  \bibinfo{pages}{1--8}.
\newblock
\urldef\tempurl%
\url{http://jmlr.org/papers/v22/20-1364.html}
\showURL{%
\tempurl}


\bibitem[Rajamani(2011)]%
        {rajamani2011vehicle}
\bibfield{author}{\bibinfo{person}{Rajesh Rajamani}.}
  \bibinfo{year}{2011}\natexlab{}.
\newblock \bibinfo{booktitle}{\emph{Vehicle dynamics and control}}.
\newblock \bibinfo{publisher}{Springer Science \& Business Media}.
\newblock


\bibitem[Riccio and Tonella(2020a)]%
        {RiccioTonella_FSE_2020}
\bibfield{author}{\bibinfo{person}{Vincenzo Riccio} {and}
  \bibinfo{person}{Paolo Tonella}.} \bibinfo{year}{2020}\natexlab{a}.
\newblock \showarticletitle{Model-based Exploration of the Frontier of
  Behaviours for Deep Learning System Testing}. In
  \bibinfo{booktitle}{\emph{Proceedings of the ACM Joint European Software
  Engineering Conference and Symposium on the Foundations of Software
  Engineering}} \emph{(\bibinfo{series}{ESEC/FSE '20})}.
  \bibinfo{publisher}{Association for Computing Machinery}, \bibinfo{pages}{13
  pages}.
\newblock
\urldef\tempurl%
\url{https://doi.org/10.1145/3368089.3409730}
\showDOI{\tempurl}


\bibitem[Riccio and Tonella(2020b)]%
        {riccio2020model}
\bibfield{author}{\bibinfo{person}{Vincenzo Riccio} {and}
  \bibinfo{person}{Paolo Tonella}.} \bibinfo{year}{2020}\natexlab{b}.
\newblock \showarticletitle{Model-based exploration of the frontier of
  behaviours for deep learning system testing}. In
  \bibinfo{booktitle}{\emph{Proceedings of the 28th ACM Joint Meeting on
  European Software Engineering Conference and Symposium on the Foundations of
  Software Engineering}}. \bibinfo{pages}{876--888}.
\newblock


\bibitem[Rosales-P{\'e}rez et~al\mbox{.}(2013)]%
        {rosales2013hybrid}
\bibfield{author}{\bibinfo{person}{Alejandro Rosales-P{\'e}rez},
  \bibinfo{person}{Carlos A~Coello Coello}, \bibinfo{person}{Jesus~A Gonzalez},
  \bibinfo{person}{Carlos~A Reyes-Garcia}, {and} \bibinfo{person}{Hugo~Jair
  Escalante}.} \bibinfo{year}{2013}\natexlab{}.
\newblock \showarticletitle{A hybrid surrogate-based approach for evolutionary
  multi-objective optimization}. In \bibinfo{booktitle}{\emph{2013 IEEE
  Congress on Evolutionary Computation}}. IEEE, \bibinfo{pages}{2548--2555}.
\newblock


\bibitem[Sakurai et~al\mbox{.}(2010)]%
        {sakurai2010method}
\bibfield{author}{\bibinfo{person}{Yoshitaka Sakurai}, \bibinfo{person}{Kouhei
  Takada}, \bibinfo{person}{Takashi Kawabe}, {and} \bibinfo{person}{Setsuo
  Tsuruta}.} \bibinfo{year}{2010}\natexlab{}.
\newblock \showarticletitle{A method to control parameters of evolutionary
  algorithms by using reinforcement learning}. In
  \bibinfo{booktitle}{\emph{2010 sixth international conference on signal-image
  technology and internet based systems}}. IEEE, \bibinfo{pages}{74--79}.
\newblock


\bibitem[Schulman et~al\mbox{.}(2015)]%
        {schulman2015high}
\bibfield{author}{\bibinfo{person}{John Schulman}, \bibinfo{person}{Philipp
  Moritz}, \bibinfo{person}{Sergey Levine}, \bibinfo{person}{Michael Jordan},
  {and} \bibinfo{person}{Pieter Abbeel}.} \bibinfo{year}{2015}\natexlab{}.
\newblock \showarticletitle{High-dimensional continuous control using
  generalized advantage estimation}.
\newblock \bibinfo{journal}{\emph{arXiv preprint arXiv:1506.02438}}
  (\bibinfo{year}{2015}).
\newblock


\bibitem[Schulman et~al\mbox{.}(2017)]%
        {schulman2017proximal}
\bibfield{author}{\bibinfo{person}{John Schulman}, \bibinfo{person}{Filip
  Wolski}, \bibinfo{person}{Prafulla Dhariwal}, \bibinfo{person}{Alec Radford},
  {and} \bibinfo{person}{Oleg Klimov}.} \bibinfo{year}{2017}\natexlab{}.
\newblock \showarticletitle{Proximal policy optimization algorithms}.
\newblock \bibinfo{journal}{\emph{arXiv preprint arXiv:1707.06347}}
  (\bibinfo{year}{2017}).
\newblock


\bibitem[Snoek et~al\mbox{.}(2012)]%
        {snoek2012practical}
\bibfield{author}{\bibinfo{person}{Jasper Snoek}, \bibinfo{person}{Hugo
  Larochelle}, {and} \bibinfo{person}{Ryan~P Adams}.}
  \bibinfo{year}{2012}\natexlab{}.
\newblock \showarticletitle{Practical bayesian optimization of machine learning
  algorithms}.
\newblock \bibinfo{journal}{\emph{Advances in neural information processing
  systems}}  \bibinfo{volume}{25} (\bibinfo{year}{2012}).
\newblock


\bibitem[Sotiropoulos et~al\mbox{.}(2016)]%
        {sotiropoulos2016virtual}
\bibfield{author}{\bibinfo{person}{Thierry Sotiropoulos},
  \bibinfo{person}{Guiochet Guiochet}, \bibinfo{person}{Ingrand Ingrand}, {and}
  \bibinfo{person}{Weaselynck Waeselynck}.} \bibinfo{year}{2016}\natexlab{}.
\newblock \showarticletitle{Virtual worlds for testing robot navigation: a
  study on the difficulty level}. In \bibinfo{booktitle}{\emph{2016 12th
  European Dependable Computing Conference (EDCC)}}. IEEE,
  \bibinfo{pages}{153--160}.
\newblock


\bibitem[Sutton and Barto(2018)]%
        {sutton2018reinforcement}
\bibfield{author}{\bibinfo{person}{Richard~S Sutton} {and}
  \bibinfo{person}{Andrew~G Barto}.} \bibinfo{year}{2018}\natexlab{}.
\newblock \bibinfo{booktitle}{\emph{Reinforcement learning: An introduction}}.
\newblock \bibinfo{publisher}{MIT press}.
\newblock


\bibitem[Todorov et~al\mbox{.}(2012)]%
        {todorov2012mujoco}
\bibfield{author}{\bibinfo{person}{Emanuel Todorov}, \bibinfo{person}{Tom
  Erez}, {and} \bibinfo{person}{Yuval Tassa}.} \bibinfo{year}{2012}\natexlab{}.
\newblock \showarticletitle{Mujoco: A physics engine for model-based control}.
  In \bibinfo{booktitle}{\emph{2012 IEEE/RSJ international conference on
  intelligent robots and systems}}. IEEE, \bibinfo{pages}{5026--5033}.
\newblock


\bibitem[Tuncali et~al\mbox{.}(2018)]%
        {tuncali2018simulation}
\bibfield{author}{\bibinfo{person}{Cumhur~Erkan Tuncali},
  \bibinfo{person}{Georgios Fainekos}, \bibinfo{person}{Hisahiro Ito}, {and}
  \bibinfo{person}{James Kapinski}.} \bibinfo{year}{2018}\natexlab{}.
\newblock \showarticletitle{Simulation-based adversarial test generation for
  autonomous vehicles with machine learning components}. In
  \bibinfo{booktitle}{\emph{2018 IEEE Intelligent Vehicles Symposium (IV)}}.
  IEEE, \bibinfo{pages}{1555--1562}.
\newblock


\bibitem[Zohdinasab et~al\mbox{.}(2021)]%
        {zohdinasab2021deephyperion}
\bibfield{author}{\bibinfo{person}{Tahereh Zohdinasab},
  \bibinfo{person}{Vincenzo Riccio}, \bibinfo{person}{Alessio Gambi}, {and}
  \bibinfo{person}{Paolo Tonella}.} \bibinfo{year}{2021}\natexlab{}.
\newblock \showarticletitle{Deephyperion: exploring the feature space of deep
  learning-based systems through illumination search}. In
  \bibinfo{booktitle}{\emph{Proceedings of the 30th ACM SIGSOFT International
  Symposium on Software Testing and Analysis}}. \bibinfo{pages}{79--90}.
\newblock


\end{thebibliography}
